\pdfoutput=1
\documentclass[twoside]{article}

\usepackage[preprint]{timetofakeit}

\newif\ifarxivheader
\arxivheadertrue
\fancypagestyle{arxivheader}{%
  \fancyhf{}%
  \fancyhead[C]{\small\bfseries Time to Fake It: Continual Machine Learning Intensifies Unfairness in Deepfake Detection}%
  \renewcommand{\headrulewidth}{0.5pt}}
\ifarxivheader\else\fi

\newcommand{\coderepo}{{\small\url{https://github.com/OWL-ML/time-to-fake-it}}}
\usepackage{hyperref}
\hypersetup{
    colorlinks=true,
    linkcolor=red,
    filecolor=magenta,
    urlcolor=blue,
    citecolor=purple,
    pdftitle={Time to Fake It: Continual Machine Learning Intensifies Unfairness in Deepfake Detection},
    pdfpagemode=FullScreen,
    }

\usepackage{times}
\usepackage{soul}
\usepackage{url}
\usepackage{xurl}
\usepackage[utf8]{inputenc}
\usepackage[small]{caption}
\usepackage{graphicx}
\usepackage{amsmath}
\usepackage{amsthm}
\usepackage{booktabs}
\usepackage{algorithm}
\usepackage{algorithmic}
\usepackage[switch]{lineno}
\usepackage{multirow}
\usepackage{array}
\usepackage{makecell}
\usepackage{subcaption}
\usepackage{amssymb}
\usepackage{wrapfig}
\usepackage{diagbox}
\usepackage{xcolor}
\usepackage{comment}
\usepackage{placeins}
\usepackage{adjustbox}
\usepackage{microtype}
\usepackage{graphicx}
\usepackage{tikz}
\usetikzlibrary{arrows.meta,calc}
\usepackage{helvet}   %

\definecolor{cDet}{HTML}{0072B2}   %
\definecolor{cGan}{HTML}{E69F00}   %
\definecolor{cDiff}{HTML}{CC79A7}  %

\newcommand{\subw}{0.63cm}   %

\newcommand{\safeimg}[1]{%
  \IfFileExists{#1}{\includegraphics[width=\subw]{#1}}{%
    \tikz[baseline=0pt]{
      \fill[black!8] (0,0) rectangle (\subw,\subw);
      \draw[black!25, line width=0.3pt] (0,0) rectangle (\subw,\subw);
      \node[font=\tiny\ttfamily, text=black!45, rotate=90,
            inner sep=0pt] at (0.5*\subw,0.5*\subw) {missing};}}}

\usepackage[round]{natbib}

\begin{document}

\twocolumn[

\papertitle{Time to Fake It: \\ Continual Machine Learning Intensifies Unfairness in Deepfake Detection}

\paperauthor{ Kajaree Das, Oliver Böse, Subarnaduti Paul, Gabriel Storch, Martin Mundt }
\paperaddress{Faculty of Mathematics and Computer Science, University of Bremen, Bremen, Germany \\ \texttt{\{dask, mundtm\}@uni-bremen.de}}
\vspace{0.03in} ]
\thispagestyle{empty}

\begin{abstract}
    Deepfake detectors safeguard against misinformation, rendering their failures as much a societal as a technical concern. As generators constantly evolve, detectors trained on static datasets miss newly arising threats, while progressive training interferes catastrophically. Yet, mitigating forgetfulness in terms of accuracy, the prevalent focus of continual learning (CL), is insufficient to reflect real-world impact. We therefore scrutinize the effects of CL methods with respect to evolving fairness. To this end, we analyze four dimensions of fairness across diverse demographics for detectors trained progressively on a chronological sequence of 11 real-world deepfake generation methods. Tracking fairness metrics over time, we expose non-trivial interplays between performance and fairness that depend on both the choice of CL algorithm and demographic intersections. We find accuracy degradation to typically exacerbate unfairness and, conversely, fairness improvements to remain scarce over time. Dissecting the CL method families (rehearsal, prompts, regularization), we observe that inequitable forgetting is not explained by underrepresentation alone, but likely entangled with representation learning. We thus issue a call to action for CL methods to take social outcomes into account when balancing the stability-plasticity tradeoff.
\end{abstract}

\section{Introduction}
Modern digital platforms experience an increasing influx of AI-generated and manipulated media. The latter range from playful image filters to potentially malicious content, such as misinformation or fabrications created for social engagement. Among these threats, \emph{Deepfakes} have raised particular concern \citep{deepfakeUnesco}, as recent generation methods have surpassed the ``Uncanny valley'' of visual realism \citep{nightingale2022}. They now produce synthetic images, videos, and audio that are often indistinguishable from authentic content. The societal consequences are severe: a substantial fraction of online deepfake imagery constitutes Non-Consensual Intimate Imagery, primarily targeting women and children \citep{eubrief}. Prior studies \citep{groh2024humandetectionpoliticalspeech,tarchi2023eeg,diel_human_2024} demonstrate that human observers are becoming progressively less reliable at identifying such manipulated media through visual inspection alone; increasingly placing the burden of detection on automated systems.
\begin{figure*}[t]
    \centering
    \includegraphics{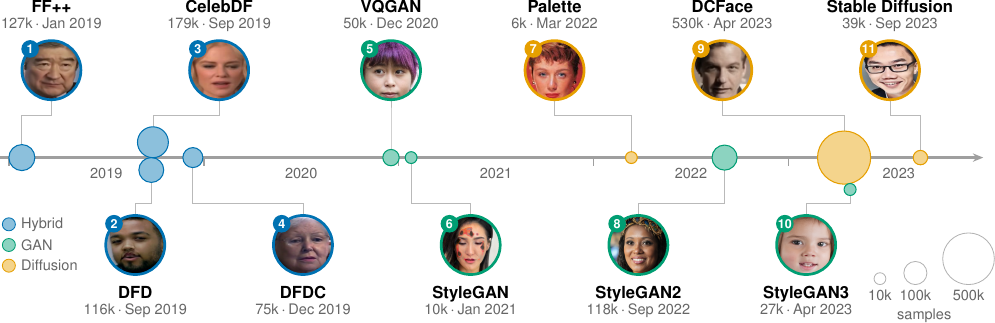}
    \caption{Our chronological sequence of 11 Deepfake datasets, on which a single detector is trained continually. Each task shows its index, the generator, size ($k$ = thousands), date and an example image. Beads mark publication dates; bead area is proportional to dataset size and color denotes the generator family. Fairness is evaluated on demographic test annotations after every training step, but training only ever uses real/fake labels.}
    \label{fig:timeline}
\end{figure*}

However, deepfake generation is far from static. New synthesis techniques, improved architectures, and novel training regimes emerge frequently. Successive distribution shifts can therefore arise in abrupt ways, often induced by core advances in machine generation methodology.
Respectively, empirical evidence \citep{pirogov2025evaluatingdeepfakedetectorswild,deepfake-study} suggests that current deepfake detection systems remain highly brittle under evolving generators.
In fact, a recent systematic study by \cite{le2025soksystematizationbenchmarkingdeepfake} evaluated $51$ deepfake detectors introduced between $2019-2023$. They conclude that no single detector consistently performs well across generators and evaluation protocols. For instance, the detectors optimized for a specific deepfake generation paradigm (e.g.~face swapping) frequently collapse when confronted with newer synthesis techniques, while cross-dataset evaluations reveal large performance disparities.

Intuitively, continual learning (CL) is poised to resolve these discrepancies. It naturally frames Deepfake detection as a temporally evolving adversarial task, where it becomes unsurprising that progressively fine-tuned detectors on novel manipulation techniques are subject to well-known catastrophic interference \citep{MCCLOSKEY1989109}. In consequence, prior works have suggested CL as a solution to maintain accuracy in continual Deepfake detection \citep{fontana2025, li2022continual}. Alas their focus on accuracy transfer discounts the application's high-stake implications in terms of societal outcomes. Regularizing past knowledge without storing data may ameliorate privacy and computational burdens, but uneven performance degradation across demographic groups may also exacerbate existing disparities or make underrepresented communities prone to fraud.

In this work, we present a systematic study of different CL methods for Deepfake detection with a particular focus on how continual adaptation affects demographic fairness over time. In contrast to contemporary works that posit CL to inherently improve fairness on aggregated metrics \citep{Churamani_2023,mansilla2025bmclbiasmitigationlens,ceccon2025fairnessevolutioncontinuallearning}, we emphasize the complex socio-technical nature of fairness \citep{selbst_fairness_2019} and account for demographic attributes to reside on an intersectional spectrum \citep{Castelnovo2022FairenssNuances}. To this end, we leverage the demographically annotated AI-Face \citep{lin2025aiface} repository and extract a temporally chronological sequence of 11 datasets, each corresponding to a unique deepfake face generation method as shown in figure \ref{fig:timeline}. In addition to accuracy over time, we analyze the evolution of four prominent fairness metrics \citep{Narayanan2018Fairness21Def}. Through our extensive evaluation, we expose that CL strategies provide only selective fairness improvements over time and instead mostly intensify unfairness. Interestingly, further analysis shows that under-representation in the data is not the sole cause. Although replay buffers faithfully reproduce the demographic mix, the disparities observed in other CL method families cannot be attributed simply to data prevalence.

\section{Continual Learning and Evolving Fairness}
A recent perspective by \cite{verwimp2024continuallearningapplicationsroad} describes continual learning as focusing on ``%
incrementally training models on a stream of data with the aim of accumulating knowledge over time''. Indeed, large bodies of research are being pointed to on the road to solve this challenge. Far fewer works have focused on the task of continual Deepfake detection, with even less studies transcending accuracy analysis and investigating the interplay of continual learning and fairness. The latter is the focus of this paper.

\textbf{Background: Formalizing Continual Deepfake Detection.} Deepfake detection can be formulated as a sequence of two-class classification tasks that aim to map input data $x_i \in \mathbb{R}^d$ to corresponding labels $y_i \in \{0,1\}$, where $0$ indicates a real and $1$ a fake item. In continual learning, the fake detector is exposed to a sequence of $T$ such tasks $\{\mathcal{T}_1, \mathcal{T}_2, \ldots, \mathcal{T}_T\}$, each associated with a respective dataset $\mathcal{D}_t = \{(x_i^{(t)}, y_i^{(t)})\}_{i=1}^{N_t}$ with $N_t$ data points at time $t$.
During training, the model learns exclusively on the current task $\mathcal{T}_t$ of the sequence, typically by optimizing Cross-Entropy, while CL methods simultaneous aim to retain knowledge from previously observed tasks $\{\mathcal{T}_1, \mathcal{T}_2, \ldots, \mathcal{T}_{t-1}\}$. The performance retention of the detector at each time can then be measured for each generation algorithm. Importantly, in our continual fake face detection scenario CL methods solely have access to true and fake labels. Whereas we later evaluate the impact from several fairness perspectives, the loss itself is never tuned explicitly on sensitive attributes like gender or skin color.

\textbf{The State of Continual Fake Face Detection.}
Given the recent proliferation of Deepfakes, their continual detection is a largely novel field. In the context of fake human faces, existing works explore only select algorithms \citep{Tassone_2024}, conduct limited analysis on smaller corpora (usually $\leq2k$ samples) \citep{10034569}, or investigate accuracy on un-intuitive task orders. For instance, the Continual Deepfake Detection Benchmark \citep{li2022continual} arranges deepfake datasets according to perceived complexity of the generators, rather than their real-world temporal onset. To analyze existing class-incremental CL methods, the setting is further relaxed for each generator to be considered a separate class. In contrast, \cite{fontana2025} addresses the transient nature of deepfakes by ordering datasets according to year of release, but the focus is placed only on detection accuracy. Although valuable, these works thus emphasize the distinct visual artifacts of generators without studying their potential to propagate demographic biases. Face detection algorithms in particular are prone to such biases, skewing final predictions towards certain demographic groups based on e.g.~gender, age or skin tone \citep{pmlr-v81-buolamwini18a,Waelen2022-WAETSF,kotwal2025reviewdemographicfairnessface}. Aggregated accuracy is unable to reflect the nuances in fair outcomes and thus the evolution of fairness over time has to be considered. We extend aforementioned studies with such analysis of how fairness interacts with CL methods across different generators.
\begin{figure}[t]
    \centering
    \includegraphics[width=0.975\linewidth]{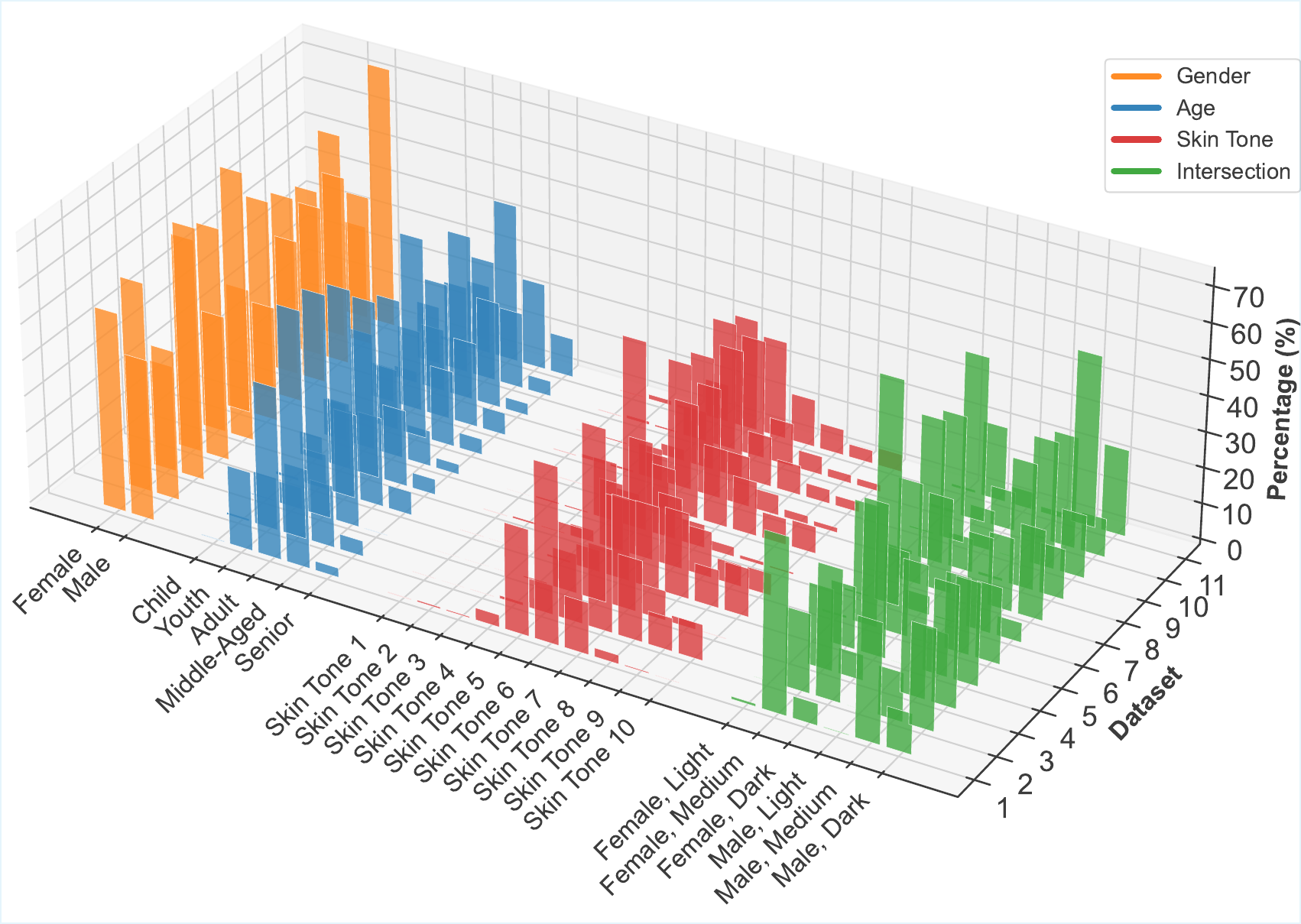}
    \caption{Demographic distribution across the 11 test sets. Colors denote disjoint attributes, each summing to 100\%; green shows binary gender intersected with three skin-tone groups (of ten). Some groups are underrepresented throughout (e.g.~very light skin tones), while others fluctuate heavily over time.}
    \label{fig:subgroups_percentage}
\end{figure}

\textbf{Analysis of Fairness in Continual Learning.}
One way fairness in CL has previously been examined is through the lens of class imbalance \citep{Lian2025MetricAgnosticCL,fatemi-etal-2023-improving}. Here, the notion of fairness corresponds to treating class labels in representational parity, e.g.~in continual semantic segmentation  \citep{truong2023fairnesscontinuallearningapproach} or image classification \citep{zhao2019maintainingdiscriminationfairnessclass}. However, fairness in light of its socio-technical challenges \citep{grote_fairness_2024} remains substantially less studied. In particular, its dimensions beyond balancing prediction rates across classes remains underexplored \citep{fairness_survey_acm2025}. Select existing works have only begun to untangle its complexity as a continually evolving property across shifting environments. On the one hand, \cite{ceccon2025fairnessevolutioncontinuallearning} equates fairness with equal opportunity and shows a single-metric is improvable through continual optimization. \cite{mansilla2025bmclbiasmitigationlens} narrows fairness down to discrepancies in accuracy across groups and conclude similarly. On the other hand, the claim that fairness can be improved through CL methods seems to mostly originate from set-ups that isolate sensitive attributes temporally \citep{Churamani_2023}.
However, the real world does not (or should not) explicitly introduce data to a system in disjoint sets of demographic, highly sensitive and likely protected, attribute over time. Our work substantially expands on these initial investigations by moving towards natural sequences with assessment spanning several fairness metrics. Contrary to mentioned prior works, we show that CL algorithms seem to mostly exacerbate unfairness in complex interrelation of CL algorithm choice and demographic dimension.

\section{A Chronology of Deepfake Datasets}
\label{sec:cffd}
In order to assemble a temporally coherent sequence of deepfake generators that allows nuanced investigation of fairness, we leverage the AI‑Face‑FairnessBench \citep{lin2025aiface}. The original benchmark is composed of a variety of sources and aggregates several generation methods, bringing together a total of $400{,}885$ real and $1{,}245{,}660$ fake face images. In addition to real or fake labels, each image has been annotated with demographic attributes such as an age group (child, youth, adult, middle-aged, senior), binary gender, and skin tone. The latter has been measured using a Monk scale \citep{monk_skin_tone}, mapping skin tone to ten categories to avoid ambiguous colors and racialized population descriptions. Intersections between skin tones and genders have further been constructed.
For intersections, Monk tones are grouped into Light (ST~[1--3]), Medium (ST~[4--7]) and Dark (ST~[8--10]).

For our purposes of training continually and analyzing the respective evolution of the detector's fairness, we have filtered AI-Face-FairnessBench and sorted data chronologically. First, we have separated subsets that correspond to unique generation methodology. We separate according to obtain a wide spectrum, spanning Face2Face, FaceSwap, DeepFakes, and NeuralTextures in FaceForensics++ \citep{rössler2019faceforensicslearningdetectmanipulated} and constituent dataset like DFD \citep{dfd_google_jigsaw}, CelebDF \citep{celeb_df}, DFDC \citep{dolhansky2020deepfakedetectionchallengedfdc}, but also data generated using methods ranging from VQGAN \citep{esser2020taming}, StyleGAN, and its variations \citep{giudice2021fighting,david_beniaguev_2022_SFHQ,stylegan3}, to more recent diffusion based generators like Palette \citep{artifact}, DCFace \citep{DCFace}, or Stable Diffusion \citep{Song2023RobustnessAG}. Subsets that only provide very small fractions of fake data based on nuanced generator differences have been discarded. As AI-Face-FairnessBench does not include real data for some generators, we have additionally queried external sources (i.e. CelebA \citep{liu2015faceattributes}, Metfaces \citep{10.5555/3495724.3496739}, Casia Webfaces \citep{Yi2014LearningFR}), following the original training specifications of each generator.
Subsequently, we have assigned each subset a date according to the corresponding dataset's public appearance.
Our construction thus reflects the time-frames in which a specific set of fake imagery became publicly available as a dataset and a corresponding fake face detector could have been trained. This allows us to mimic a real continual learning application and analyze its evolution in terms of accuracy in tandem with fairness. Figure \ref{fig:timeline} depicts the timeline with data subset sizes and generation methods.
Overall, our deepfake generator sequence consists of 11 data subsets, respectively referred to as ``tasks'' to stay consistent with continual learning terminology. Although training ever only has access to fake and true labels, figure \ref{fig:subgroups_percentage} highlights each task's test distribution in terms of demographic representation.

Here, two critical observations emerge. First, select demographics are represented to different degrees in each subset. In many cases, intermittent skin tones compose the majority of the data. Similarly, children and seniors are less overall represented. Second, the demographic distribution changes substantially across time, mimicking likely occurring real world sampling and pipeline biases based on developer considerations. We posit that this changing demographic distribution, that is inaccessible to the learner, will lead to complex interplays in continual training, which we investigate in our later experimental section across four fairness metrics. Additional details about all datasets, download sources, and where relevant, pre-processing steps, are all contained in the Appendix \ref{app:dataset_stats}.

\subsection{Evaluating the Evolution of Fairness}
\label{subsec:training_setup_eval_metrics}
For a deepfake detector to be effective it needs to become and remain accurate over time. For the detector to serve in the real world it also needs to remain fair. Specifically, a detector becomes problematic when trade-offs in accuracy over time come at the expense of select demographics.

To study this evolution, we report both end-of-training averages and performance on previously seen subsets at every step of the training sequence, capturing overall quantities as well as stability and emerging disparities over time. We measure detection performance by accuracy and fairness by four well established metrics, framed as unfairness to highlight discrepancies, where a normalized value of $0$ means perfect parity, and values closer to $1$ indicate unfairness. The following describes each metric intuitively and motivates why differing notions of ``objectivity'' need to be captured. Full mathematical definitions are in Appendix \ref{app:fairness_metrics}.

\textbf{Demographic Parity ($F_{DP}$)} quantifies whether a face being detected as ``fake'' is equally likely across different groups irrespective of ground truth labels \citep{dwork2011fairnessawareness}. It is particularly useful when the dataset is skewed, allowing us to look for  manipulation artifacts instead of spurious correlations like ``female means fake''. For example, if a model detects $10\%$ of the inputs as fake, around $10\%$ should be flagged as fake for every demographic. However, since equal fake prediction rate does not guarantee the correctness of predictions, using just $F_{DP}$ in isolation could be highly misleading in practice.

\textbf{Equal Opportunity ($F_{EOP}$)} quantifies whether true positive rates (TPR) are comparable across demographics, allowing false positive rates (FPR) to differ \citep{hardt2016eo}. It mandates that every fake face is equally likely to be correctly classified irrespective of demographic identity. It is especially useful when the priority is providing protection against threats, rather than having a focus on false alarms. However, $F_{EOP}$ overlooks the fact that equal protection for groups may come at the cost of higher rate of false accusations for minorities.

\textbf{Equalized Odds ($F_{EO}$)} quantifies a stricter version of $F_{EOP}$. It compares TPR, i.e.~rate of correctly detecting fakes, with FPR, i.e.~ rate of incorrectly flagging real as fake \citep{hardt2016eo}. It is highly practical, since protecting each group against threats and avoiding false accusations are both central to deepfake detection. However, $F_{EO}$ falters if TPR and FPR vary substantially across groups. In general, balancing these rates requires accepting a trade-off at the expense of lower performance.

\textbf{Overall Accuracy Equality ($F_{OAE}$)} quantifies if total accuracy of predictions remains invariant across groups and thereby captures if any disparity in overall performance exists \citep{berk2017fairnesscriminaljusticerisk}. Real faces from every demographic should experience comparable authentication success rate and reliable detection rate of fraudulent attempts. However, $F_{OAE}$ does not reveal which precise source of error is responsible for the accuracy gap.

\begin{figure*}[t]
\begin{minipage}[t]{0.485\textwidth}
    \centering
    \includegraphics[height=6cm, trim=10 13 10 13, clip]{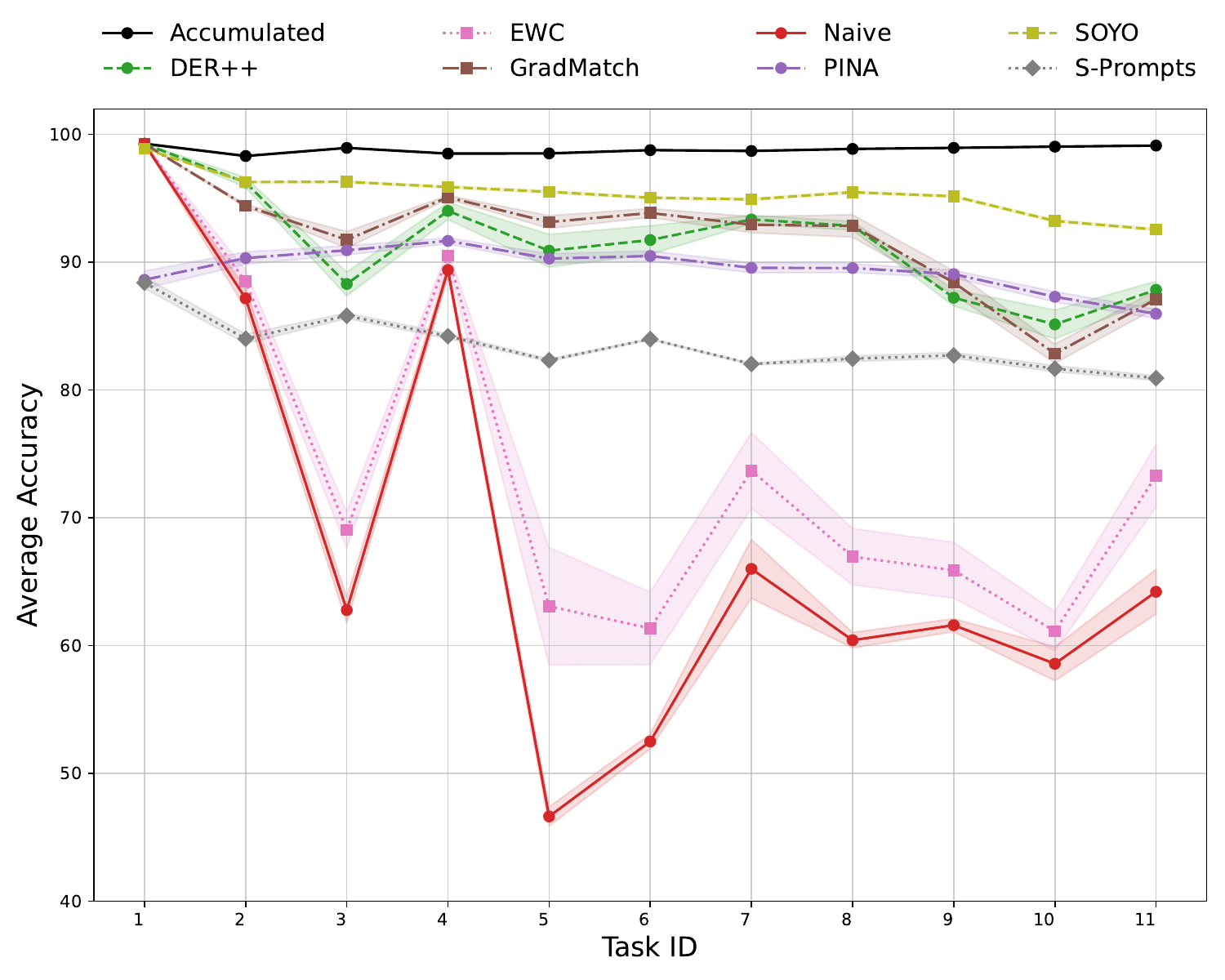}
    \caption{Average accuracy (over five seeds) on all seen tasks (y-axis) after training on each generator (x-axis). Shading shows the standard deviation. Naive fine-tuning and EWC degrade substantially, while SOYO, GradMatch and DER++ approach the accumulated upper-bound. PINA and S‑Prompts are lower but more stable; GradMatch and DER++ lose stability in later tasks.}
    \label{fig:avg-accuracy-trajectory}
    \end{minipage}
    \hfill
    \begin{minipage}[t]{0.485\textwidth}
    \centering
    \includegraphics[height=6cm, trim=19 22 60 84, clip]{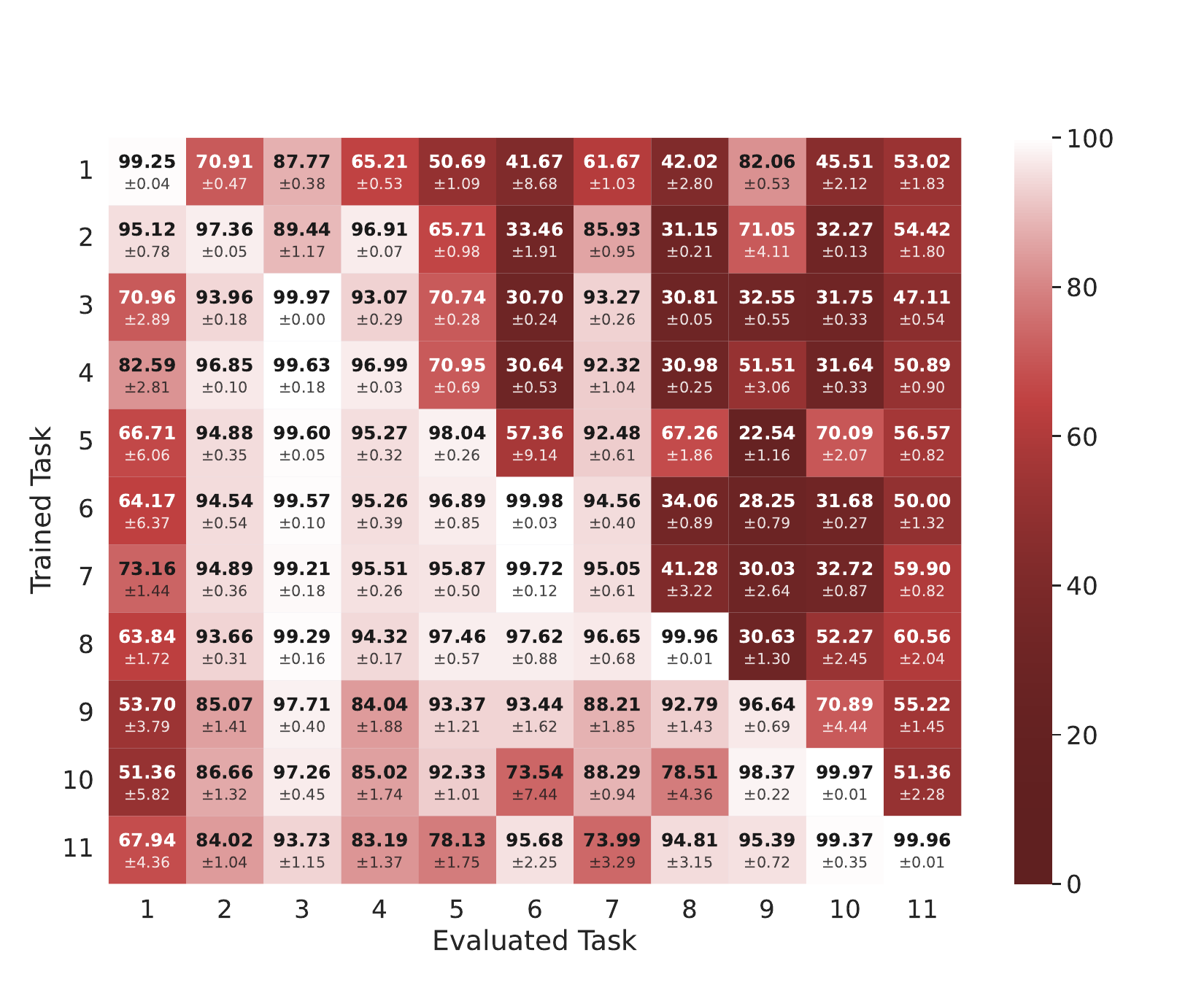}
    \caption{Inter-task accuracy of DER++ by training step (y-axis) and evaluated task (x-axis), as mean $\pm$ std over five seeds. Each task is solvable alone (diagonal), but unseen generators (top-right) show inconsistent improvement, as no method targets forward generalization. CL mitigates forgetting of most tasks (bottom-left). Appendix \ref{app:acc-heatmaps} shows all other CL methods.}
     \label{fig:derpp_heatmap}
\end{minipage}
\end{figure*}
\section{Interplay between Accuracy and Fairness Over Time}
To investigate retention of fake detection accuracy in interplay with fairness across the 11 generators over time, we evaluate two common continual learning baselines and six prominent algorithms. As baseline references we consider the best-case and worst-case scenario, where respectively the dataset is simply progressively grown over time (referred to as ``accumulated'') or alternatively the learner is only ever exposed to the current dataset at a time without any technique to retain prior knowledge (referred to as ``naive sequential'' training or sometimes called naive ``fine-tuning''). From a perspective of accuracy they allow to contextualize all other methods, representing the highest achievable accuracy and conversely largest possible amount of incurred forgetting. As CL methods, we consider EWC \citep{Kirkpatrick2017EWC}, Dark Experience Replay (DER++) \citep{buzzega2020darkexperiencegeneralcontinual}, GradMatch \citep{killamsetty2021gradmatchgradientmatchingbased}, SOYO \citep{wang2025boosting}, PINA \citep{wang2024non} and S-Prompt \citep{wang2022sprompt} in a balance of regularization-based, replay-based, and prompt-based strategies for continual learning.

For all models, we employ a Vision Transformer backbone (ViT-B/16) \citep{dosovitskiy2021imageworth16x16words} pretrained on ImageNet. All results are averaged over five random seeds and reported with corresponding standard deviations. A more detailed description of the methods and implementation choices are provided in Appendix \ref{app:method_overview} and \ref{app:hyperparams} respectively. Our code and data sequences are publicly available at: \\ \coderepo \\
We structure our ensuing analysis and discussing of results according to the following four aspects:%
\begin{enumerate}
    \item We highlight that select CL methods are capable of progressively maintaining high detection accuracy.
    \item We dissect final fairness discrepancies between CL methods after training on the entire sequence is done.
    \item We showcase how the evolution of (un)fairness is spread across individual dataset slices over time, exposing how CL algorithms can result in unfair treatment of vulnerable demographics.
    \item We probe to what degree the statistical under-representation of training data for certain demographics over time directly induces the unfair treatment of each CL method family.
\end{enumerate}

% Side-by-side fairness tables spanning the full \textwidth: the outer image
% edges sit on the text margins, caption width == image width, gap adjustable.
%
% table* (not figure*) so \caption numbers them as "Table N" and they enter the
% List of Tables. The graphic is typeset once into a savebox, so \wd gives its
% final width; \captionbox (caption package) then sets the caption to exactly
% that width.
%
% Preamble ------------------------------------------------------------------
\newsavebox{\tabboxA}\newsavebox{\tabboxB}
\newlength{\tabwA}\newlength{\tabwB}
\newlength{\tabgap}

% The gap between the two images. Each image is (\textwidth - \tabgap)/2 wide,
% so the pair always fills \textwidth. \columnsep matches the gap between the
% text columns, making each image exactly \columnwidth; larger values open the
% gap and narrow both images by half the difference.
\setlength{\tabgap}{\columnsep}

% Body ----------------------------------------------------------------------
\savebox{\tabboxA}{%
  \includegraphics[width=\dimexpr(\textwidth-\tabgap)/2\relax]{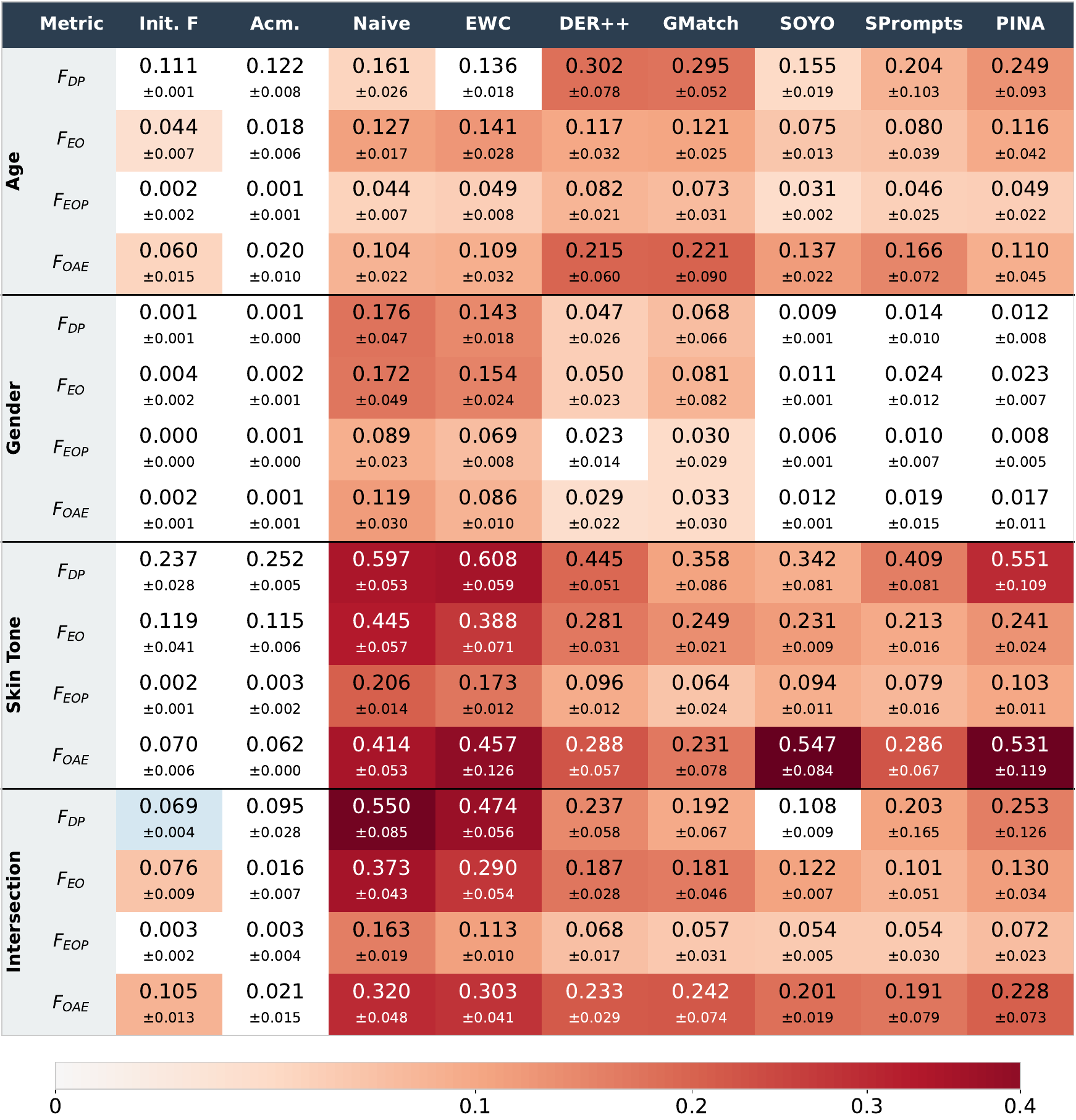}}
\savebox{\tabboxB}{%
  \includegraphics[width=\dimexpr(\textwidth-\tabgap)/2\relax]{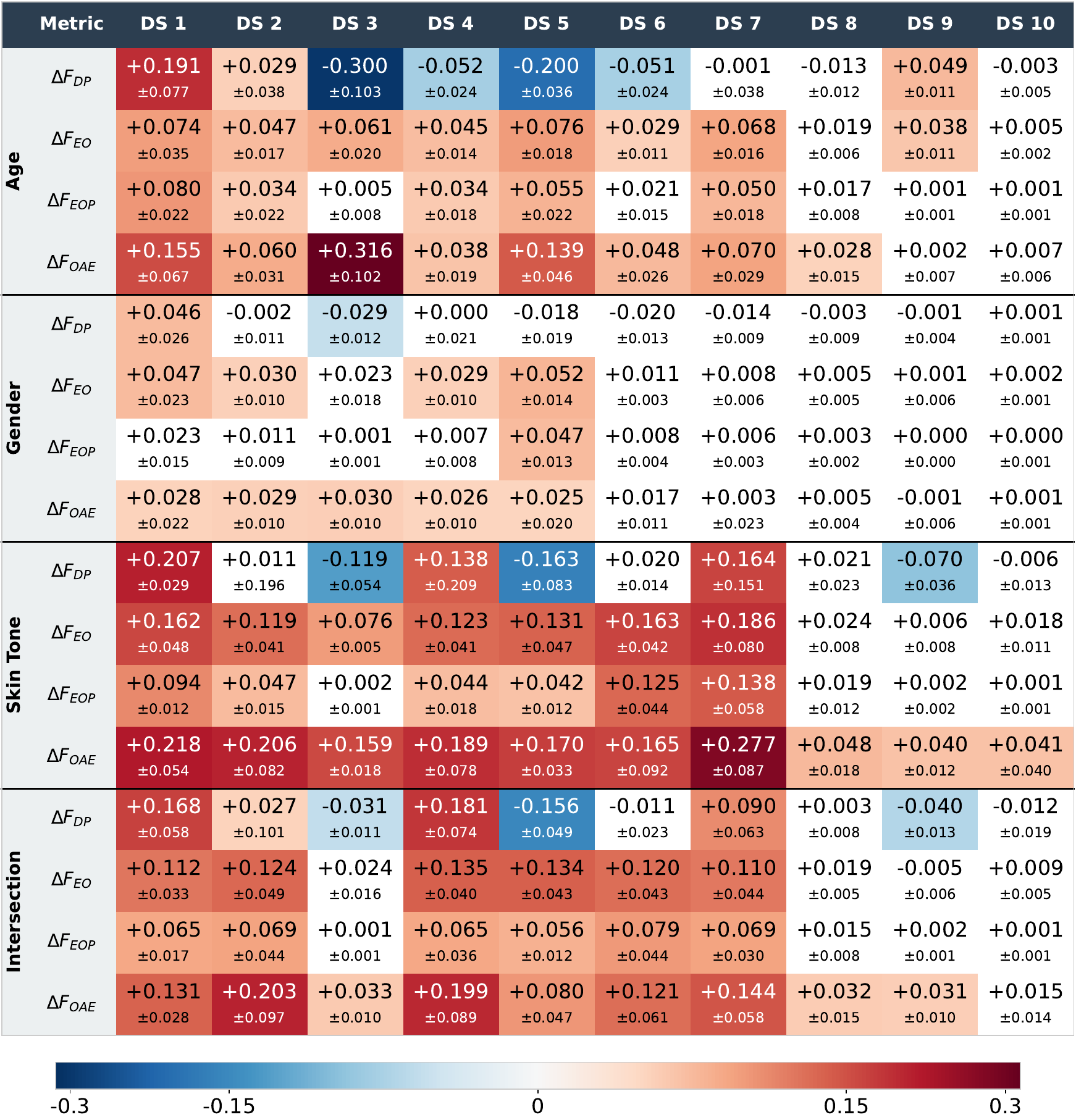}}
\setlength{\tabwA}{\wd\tabboxA}
\setlength{\tabwB}{\wd\tabboxB}

\begin{table*}[t]
  \centering
  \captionbox{(Un)fairness of CL methods on the first generator after training on all 11 generators, per demographic dimension and intersection (four normalized metrics; 0 = parity, 1 = most unfair), as mean $\pm$ std over five seeds.
  ``Init. F'': fairness after training on the first generator only, before any continual update. Blue marks improvement and red marks deterioration in fairness relative to ``Init. F''. The accumulated upper bound (``Acm.'') largely retains fairness, whereas others worsen disproportionately to accuracy.\label{tab:ds_1_gen_11}}[\tabwA]{\usebox{\tabboxA}}%
  \hfill
  \captionbox{Change in (un)fairness of DER++ on each earlier task's test set after training on all 11 generators, relative to its fairness right after training on that task (e.g.~test set 2 after task 2 vs.~after task 11), as mean $\pm$ std over five seeds. Blue marks improvement and red marks deterioration from the reference. Most fairness dimensions deteriorate (red) and only a few improve (blue). Cells with an absolute change below 2.5\% are left white for legibility.\label{tab:derpp_per_ds}}[\tabwB]{\usebox{\tabboxB}}
\end{table*}

\subsection{CL Methods Preserve Detection Accuracy}
\label{subsec:accuracy_eval}
We start by comparing how overall detection accuracy evolves as new deepfake generation techniques are introduced in figure \ref{fig:avg-accuracy-trajectory}. This aggregated view of overall performance highlights the expected severe forgetting when only training naively on the new generator, dropping to 46.6\% average accuracy at task 5 and recovering only modestly afterwards. In contrast, the accumulated baseline is rightfully treated as a continual learning upper-bound to accuracy, remaining close to $99\%$ throughout. More modern CL approaches fall in between, with EWC falling short, SOYO, GradMatch and DER++ forming fairly flat curves that retain most performance while learning to detect new generator content, ending at an average accuracy of about 88\% (DER++, GradMatch) and 93\% (SOYO).

To provide a more granular view, figure \ref{fig:derpp_heatmap} shows an exemplary inter-task accuracy matrix. Here, the diagonal elements highlight performance on the current generator at the point of its training, whereas the off-diagonal triangles respectively quantify forgetting (bottom-left) and zero-shot generalization (top-right). In general, each fake generator can be learned to be recognized. However, it can also be seen that deepfake detectors are highly specific, as they do not generalize well to unseen fake data. This motivates our continual learning approach, where the column values below the diagonal highlight increased forgetting through darker shadings. Although the particular matrix only shows DER++ (remaining matrices in the Appendix \ref{app:acc-heatmaps}), the general trend can easily be summarized. As is already evident from figure \ref{fig:avg-accuracy-trajectory}, the naive baseline collapses in terms of stability. In contrast, the accumulated baseline is light throughout the entire bottom-left triangle, whereas the top-right ``forward generalization'' remains limited for all as no method addresses zero-shot learning. As CL methods, SOYO, GradMatch and DER++ produce a uniformly lighter matrix than naive and EWC, indicating that they retain earlier knowledge almost as well as they acquire new knowledge. Although forgetting is largely mitigated, especially older tasks cannot be retained fully. As such, a model’s loss of earlier learned knowledge can become an exploitable vulnerability eventually, especially if the observed degradation in accuracy manifests asymmetrically across demographics.

\subsection{CL Methods Are Heterogeneously Unfair}
\label{subsec:contnual_fairness_eval}
Detection accuracy tells us how many faces are labeled correctly, but it masks how the detector performs across protected groups. That is, does the model consistently misclassify a particular group while maintaining an acceptable overall performance over time?

To analyze this question, table \ref{tab:ds_1_gen_11} compares unfairness scores for different CL methods over age, gender, skin tone and intersection (gender $\times$ skin tone) for the first generator's test set at the end of training on the eleven-fold sequence. ``Init.\ F.'' column (see table \ref{tab:ds_1_gen_11}) shows the fairness metrics obtained after training on the first generator, i.e., before any forgetting prevention measures are applied and thus serves as the reference. The prompt based methods do not share this starting point, but are referenced against it throughout for consistency.

Again, the accumulated baseline provides a upper bound for fairness stability with values close to ``Init.\ F.'' while other CL methods widen demographic disparities in comparison. Conversely, the naive baseline illustrates the expected impact of unmitigated forgetting, which also results in fairness disparities. Yet, in several dimensions the CL methods end up more unfair still. Particularly, $F_{OAE}$ is worse than naive for age across all six methods and for skin tone under EWC, SOYO and PINA. Exceeding the lower bound is thus specific to particular metrics and methods at limited dimensions rather than a general trend and past knowledge retention does not automatically guarantee fairness parity.
For Gender, the prompt based methods stay close to the accumulated upper bound while EWC and naive get worse in $F_{DP}$ and $F_{EO}$ relative to ``Init.\ F.''. For skin tone, SOYO and PINA end with the highest $F_{OAE}$ (0.547 and 0.531, against 0.414 for naive), while EWC's skin-tone $F_{DP}$ is merely on par with naive (0.608 vs.\ 0.597).
Interestingly, for intersection between gender and skin tone groups, prompt based methods recover in $F_{DP}$ and $F_{OAE}$ but naive and EWC remain among the most unfair.

These observations hold despite the fact that GradMatch, DER++ and SOYO feature less accuracy degradation, highlighting that knowledge retention is attained at the expense of select demographics.
The drift in fairness is also heterogeneous and no consistent pattern emerges for methods across attributes and metrics. Thus, unlike accuracy, which is largely retained, CL approaches thus seem to largely amplify demographic disparities and disproportionately exacerbate unfairness in deepfake detection over time. In Appendix \ref{app:fairness_disparities}, we extend this analysis by additionally tracing the evolution of these fairness metrics at a more granular level across all demographic subgroups.

\subsection{Not Only Older Subsets Across The Sequence Experience Fairness Degradation}
\label{subsec:fairness_vs_forgetting}
To nuance our findings and find whether degradation holds equally for all subsets, we analyze fairness with respect to each test set at the end of continual learning. To emphasize the continual aspect, we now focus further on absolute differences in fairness over time. Specifically, we compute the discrepancy for a given fairness score on a particular demographic group between the model trained on dataset $t$ and respectively evaluated at that time, and the same test set of generator $t$ evaluated at the end of training on the full eleven generator sequence. In other words, we highlight the relative fairness change experienced due to continued training of the detector on further fake face generator data.

We show the respective improvement or degradation in table \ref{tab:derpp_per_ds} for DER++. In short, we find fairness degradation to not directly correlate
with overall forgetting. For instance, test dataset (DS) 7 exhibits the worst $F_{EO}$ and $F_{OAE}$ for skin tone at the second highest point of forgetting (recall figure \ref{fig:derpp_heatmap}). At the same time, DS 6 and 4 also show substantial fairness degradation, despite experiencing significantly lower forgetting (~$4-13\%$, respectively). Conversely, fairness degradation is more pronounced for DS 1 despite having final accuracy comparable to DS 7. The inconsistency becomes even more perceptible at DS 3, which exhibits relatively lower forgetting ($\approx6\%$), yet shows the largest increase in $F_{OAE}$ for age.

Similar tables for all other methods are again deferred to the Appendix \ref{app:fairness_vs_forgetting}.
Fairness gains or stability appear almost only in prompt based methods (see table \ref{tab:app_fairness_per_ds_1}), mainly on later tasks (DS 8–10), with earlier ones (DS 2 and 6) disproportionately affected.
We attempt to further clarify the relationship between forgetting and fairness by interpreting the individual metrics. Across methods and tasks, $F_{DP}$ changes in both direction with marked improvements on some test sets (e.g., DS 3--5 for age under DER++), whereas $F_{EO}$, $F_{EOP}$ and $F_{OAE}$ predominantly worsen.
The improving $F_{DP}$ means demographics are labeled as `fake' at roughly similar rates, yet the worsening $F_{OAE}$ suggests that decision accuracy diverges. The higher increase in $F_{EO}$ relative to $F_{EOP}$ further indicates that certain demographics are now misclassified as `fake' more often and forgetting is unevenly distributed. Although an improved $F_{DP}$ may imply more equitable outcomes, the complementary perspectives provided by $F_{EO}$ and $F_{OAE}$ expose hidden disparate treatment of specific groups.
We therefore hypothesize that the decline in fairness stems from forgetting features specific to protected groups, rather than from aggregate forgetting alone.

\begin{figure*}[t]
    \centering
    \includegraphics[width=0.98\linewidth]{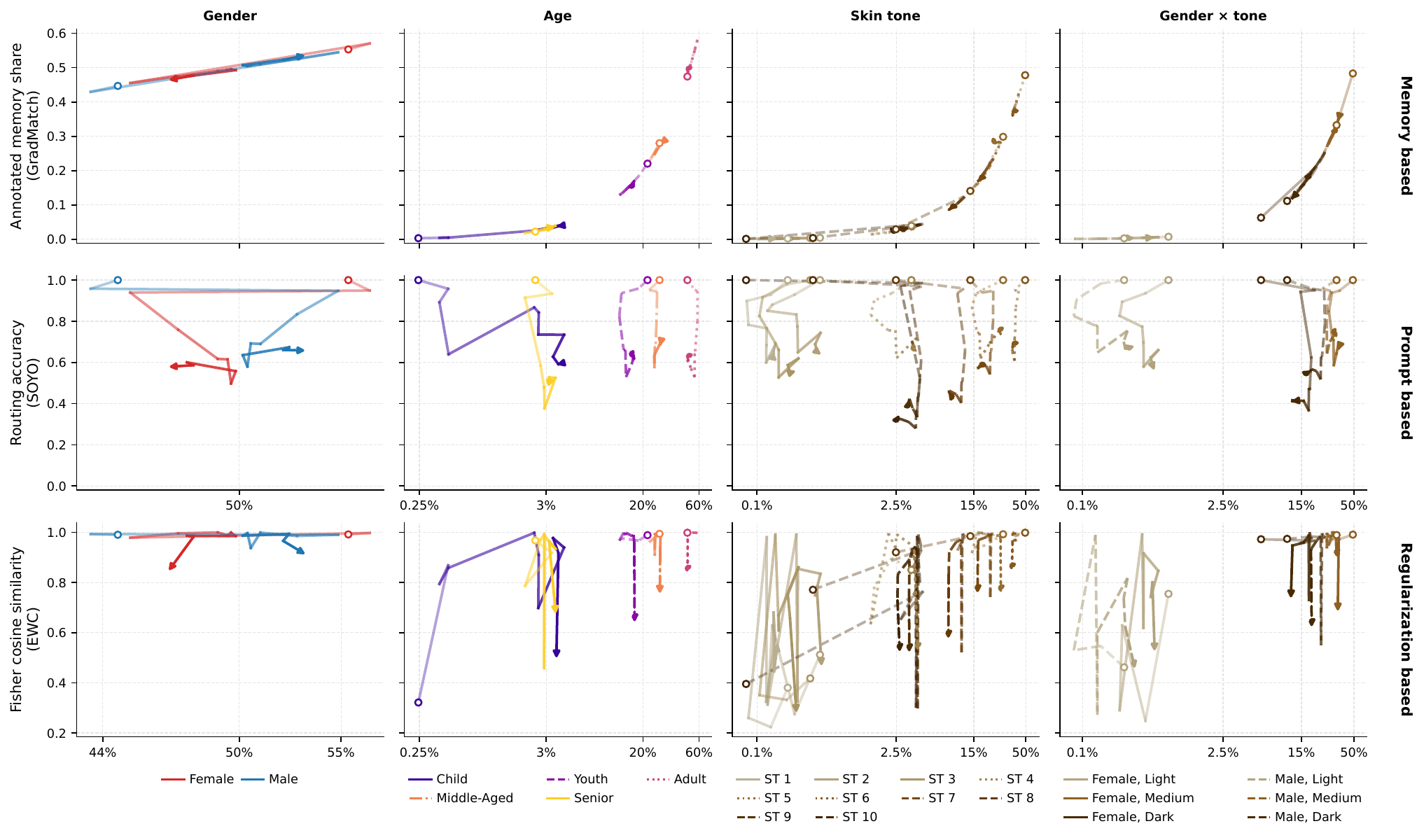}
    \caption{Subgroup ($g$) trajectories over tasks 1$\rightarrow$11 (circle$\rightarrow$arrow) against cumulative share of training data (log scale).
            \emph{Top (GradMatch, memory-based):} \emph{memory share}, the $g$'s share of the replay buffer, i.e.\ how much memory is spent rehearsing it.
            \emph{Middle (SOYO, prompt-based):} \emph{routing accuracy}, the fraction of the $g$'s test samples from all seen tasks routed to their own task's prompt. Lower means more often matched with another task's prompt.
            \emph{Bottom (EWC, regularization-based):} \emph{Fisher cosine similarity} between the diagonal Fisher on the $g$'s samples and on each task's full dataset at the final checkpoint. Lower means the $g$'s important parameters are less protected.
            Scores are averaged over five seeds. The shared log x-axis allows comparing each subgroup across method families.}
    \label{fig:failure_mode_analysis_grid}
\end{figure*}

\subsection{Does Under‑Representation Explain Unfairness?}
To identify which demographics are disproportionately forgotten by CL methods and why, we evaluate how each method family retain and retrieve knowledge for each subgroup.
Since our deepfake detection is deliberately posed as a binary classification problem and protected demographic attributes are never exposed during training, \emph{one could posit that inter‑task knowledge preservation is essentially governed by the dominant patterns in the training data. Unfairness is then hypothesized to be exacerbated in CL due to under-representation of certain demographics in the data stream translating into diminished prompt routing accuracy, lower replay buffer retention rates, or weaker alignment with the parameter importance scores used by regularization methods.}

To evaluate this hypothesis, we trace each demographic subgroup ($g$) across successive tasks and plot its cumulative data share ($s_g(t)$) against the corresponding method-specific metric (figure~\ref{fig:failure_mode_analysis_grid}).
At inference, \emph{prompt based methods} route each sample to the most similar prompt, irrespective of its true task, so a misrouted sample may be misclassified even if its own task's prompt would classify it correctly.
We therefore track the \textbf{routing accuracy}, the fraction of $g$'s test samples routed to the prompt of their own task.
\emph{Replay based methods} keep a buffer of limited size in which samples from current and previous tasks compete for slots. From saved snapshots of the buffer after each task, we compute the \textbf{memory share}, the fraction of samples that belong to $g$ in the buffer.
\emph{Regularization-based methods} penalize changes to parameters deemed important for the data. The \textbf{Fisher similarity} is the cosine similarity between the diagonal empirical Fisher of the final model computed on $g$'s samples of a given task and on all samples of that task. A low similarity means that the parameters important for $g$ are less protected.
If under-representation drives unfairness, subgroups with a smaller cumulative share should score lower on all three measures consistently.

As per our findings, \textbf{memory Share} mirrors the distribution observed in the training data. Each subgroup’s share follows its cumulative data share, as evidenced by the consistently smoother trajectories for each demographic in figure \ref{fig:failure_mode_analysis_grid}. Consequently, under‑represented subgroups receive fewer rehearsals and are therefore more vulnerable to forgetting.
\textbf{Routing accuracy} shows a more nuanced pattern.
Despite an almost balanced gender distribution, `female' are slightly more misrouted as tasks progress. `Child', as scarce as `senior', are routed as well as `adult'. Even with a higher cumulative data share, routing accuracy for darker skin tones (ST 8–10) drop below $50\%$, while lighter tones (ST 1–3) stay at $50-60\%$. Dark skinned intersectional subgroups suffer similar disparities. Since routing is driven by feature similarity, the poorer alignment of darker skinned samples with their task prompts cannot be attributed solely to their data share.
\textbf{Fisher cosine similarity} again places `female' samples at a slight disadvantage. Scores for `senior' and `child' oscillate across tasks, with a sharp decline for `child' at the final task. Unlike routing, where lighter and darker tones diverge, Fisher similarity shows both subgroups to be almost equally misaligned and therefore both receive weaker protection than the dominant medium tones. The lack of representation clearly penalizes the $Female\times light$ and $Male\times light$ subgroups. Yet, $Female\times dark$ ($s_g(T) \approx 12\%$) is better protected than $Female\times Medium$ ($s_g(T) \approx34\%$) and $Male\times Dark$ ($s_g(T) \approx19\%$). Thus, representation can not fully account for which subgroups EWC protects.

Accordingly, these results provide only partial support for our initial hypothesis. While smaller share in training data clearly leads to weaker representation in replay memory, it does not systematically weaken routing or parameter protection.
Consequently, under‑representation alone is insufficient to explain the observed fairness degradation. Disparities also differ across methods, as darker skin tones are the most misrouted ones yet no more under protected than lighter tones under regularization. The disparities also diverge across task sequences. Fisher similarity degrades for nearly all subgroups on the final task, whereas routing accuracy shows no such pattern. This heterogeneity across subgroups, methods and datasets renders a one-size-fits-all correction inadequate. Instead, safeguards must be method specific and rely on intrinsic data signals without the need for protected demographic labels.

\section{Conclusion}
As machine learning methods advance to generate ever novel deepfakes, their respective detection systems must inevitably rely on continual learning (CL) to keep up to date. However, the prevalent focus on technical performance indicators such as accuracy, fails to capture the asymmetric social consequences of continuously deploying such models. Unlike accuracy, fairness is multi-faceted and cannot be collapsed into a single metric.
As our findings show, its notions often conflict. Therefore, any claim of `improved fairness' must specify which notion improves and at whose expense.
Moreover, CL methods often exacerbate unfairness in non-trivial ways, with underrepresented groups being disproportionately affected. These disparities cannot be explained by uneven representations alone. Nor can one simply curate balanced benchmarks, as real world data streams are inherently skewed and hardly ever cleanly labelled. Demographic information in particular is sensitive which makes storing and accumulating such data prohibitive.
Thus, trustworthy and responsible systems must account for imbalance by design, combining temporal coherence with method specific safeguards, ensuring that technological adaptability does not come at the cost of ethical accountability.

\section*{Acknowledgments}
Gabriel Storch was supported by the Research Training Group ``KD$^2$School - Designing Biosignal-Adaptive Systems for Decision-Making Processes``, funded by the Deutsche Forschungsgemeinschaft (DFG, German Research Foundation) as GRK 2739/2, Project Nr. 447089431.
\bibliography{sample}

\clearpage
\appendix
\thispagestyle{empty}

\onecolumn
\papertitle{Supplementary Material for
``Time to Fake It: \\ Continual Machine Learning Intensifies Unfairness in Deepfake Detection''}
\noindent This supplementary document contains all details of the datasets considered, an overview of the evaluated fairness metrics and continual learning (CL) methods, alongside implementation choices and display of the full set of experimental results. These materials, in particular the latter set of results, are intended to provide complete clarity and support reproducibility. \emph{As described in the main body, respective code and data sequences are publicly available at: \\} \coderepo.

\section{Empirical Fairness Definitions}
\label{app:fairness_metrics}
We have provided intuitive definitions for the considered fairness metrics and have described their importance for the considered continual deepfake detection task in the main body. In this supplementary section, we include the respective mathematical definitions. For the following equations, we consider the following notation. Let $Y
\in{0,1}$ denote the ground truth label and $\hat{Y}\in{0,1}$ be the predicted label. ${D}$ is the demographic variable and $\mathcal{J}$ is the set of subgroups with each subgroup ${\mathcal{J}_i \in \mathcal{J}}$. $n$ denotes the number of samples and $\mathbb{I}[\cdot]$ is an indicator function.

\paragraph{\textbf{Demographic Parity} ($F_{DP}$)} quantifies disparity in outcome allocation across demographic subgroups. For each subgroup ${\mathcal{J}_i}$, we compute the proportions of samples predicted as positive and then take the maximum difference in prediction rates across all demographic subgroups.
\begin{equation}
\begin{aligned}
F_{DP} :=\;&
\max_{\mathcal{J}_i \in \mathcal{J}}
\frac{\sum_{j=1}^{n} \mathbb{I}\!\left[\hat{Y}_j = 1,\, D_j = \mathcal{J}_i\right]}
     {\sum_{j=1}^{n} \mathbb{I}\!\left[D_j = \mathcal{J}_i\right]}
-
\min_{\mathcal{J}_{i'} \in \mathcal{J}}
\frac{\sum_{j=1}^{n} \mathbb{I}\!\left[\hat{Y}_j = 1,\, D_j = \mathcal{J}_{i'}\right]}
     {\sum_{j=1}^{n} \mathbb{I}\!\left[D_j = \mathcal{J}_{i'}\right]} .
\end{aligned}
\end{equation}

\paragraph{\textbf{Equal Opportunity} ($F_{EOP}$)} measures subgroup specific true positive rate (TPR) and measures its deviation from overall TPR and then average over all available subgroups. A higher value indicates that some groups are systematically less likely to receive correct positive predictions.
\begin{equation}
\begin{aligned}
F_{EOP} :=\;&
\frac{1}{|\mathcal{J}|}
\sum_{\mathcal{J}_i \in \mathcal{J}}
\Bigg|
\frac{\sum_{j=1}^{n}
\mathbb{I}\!\left[\hat{Y}_j = 1,\, Y_j = 1,\, D_j = \mathcal{J}_i\right]}
     {\sum_{j=1}^{n}
\mathbb{I}\!\left[Y_j = 1,\, D_j = \mathcal{J}_i\right]}
-
\frac{\sum_{j=1}^{n}
\mathbb{I}\!\left[\hat{Y}_j = 1,\, Y_j = 1\right]}
     {\sum_{j=1}^{n}
\mathbb{I}\!\left[Y_j = 1\right]}
\Bigg| .
\end{aligned}
\end{equation}

\paragraph{\textbf{Equalized Odds} ($F_{EO}$)} is stricter form of $F_{EOP}$ and it computes the disparity in true positive rate (TPR) and false positive rate (FPR) between each subgroup ${\mathcal{J}_i}$ and the overall population and and then average over all available subgroups. Larger $F_{EO}$ indicates that groups unevenly suffer from missed detection (low TPR) or false alarms (high FPR).
\begin{equation}
\begin{aligned}
F_{EO} :=\;&
\frac{1}{|\mathcal{J}|}
\sum_{\mathcal{J}_i \in \mathcal{J}}
\sum_{q=0}^{1}
\Bigg|
\frac{\sum_{j=1}^{n}
\mathbb{I}\!\left[\hat{Y}_j = 1,\, Y_j = q,\, D_j = \mathcal{J}_i\right]}
     {\sum_{j=1}^{n}
\mathbb{I}\!\left[Y_j = q,\, D_j = \mathcal{J}_i\right]}
 -
\frac{\sum_{j=1}^{n}
\mathbb{I}\!\left[\hat{Y}_j = 1,\, Y_j = q\right]}
     {\sum_{j=1}^{n}
\mathbb{I}\!\left[Y_j = q\right]}
\Bigg| .
\end{aligned}
\end{equation}

\paragraph{\textbf{Overall Accuracy Equality} ($F_{OAE}$)} compares the total prediction accuracy across each group and measures the maximum accuracy gap. Higher disparity indicates that some demographic groups suffer from less reliable predictions over all, even if TPR or FPR appear balanced.
\begin{equation}
\begin{aligned}
F_{OAE} :=\;&
\max_{\mathcal{J}_i \in \mathcal{J}}
\frac{\sum_{j=1}^{n}
\mathbb{I}\!\left[\hat{Y}_j = Y_j,\, D_j = \mathcal{J}_i\right]}
     {\sum_{j=1}^{n}
\mathbb{I}\!\left[D_j = \mathcal{J}_i\right]}
 -
\min_{\mathcal{J}_{i'} \in \mathcal{J}}
\frac{\sum_{j=1}^{n}
\mathbb{I}\!\left[\hat{Y}_j = Y_j,\, D_j = \mathcal{J}_{i'}\right]}
     {\sum_{j=1}^{n}
\mathbb{I}\!\left[D_j = \mathcal{J}_{i'}\right]} .
\end{aligned}
\end{equation}

\paragraph{\textbf{From group aggregates to individual subgroups}}

Each demographic attribute spans several subgroups (table \ref{tab:app_demographics}): binary gender, five age groups, ten skin tones (ST), and six intersectional subgroups combining gender with the skin tone groups Light (ST~[1--3]), Medium (ST~[4--7]) and Dark (ST~[8--10]). The group-level measures of fairness restrict the view to one number per attribute and may thus mask conflicting granular patterns. The same value can arise from one specific subgroup departing sharply while the rest sit near the average, or from a broadly even spread across all of them. We therefore also report, for each subgroup, its deviation from the overall test set, which is not broken down by demographics at all. These overall rates are the performance figures reported without any demographic breakdown, and they serve as the same reference for every attribute. So deviations are directly comparable across gender, age, skin tone and their intersection.

\begin{equation}
\begin{aligned}
F_{DP}^{(\mathcal{J}_i)}  &= \big\lvert \mathrm{PR}(\mathcal{J}_i)  - \mathrm{PR}  \big\rvert, \\
F_{EOP}^{(\mathcal{J}_i)} &= \big\lvert \mathrm{TPR}(\mathcal{J}_i) - \mathrm{TPR} \big\rvert, \\
F_{EO}^{(\mathcal{J}_i)}  &= \big\lvert \mathrm{TPR}(\mathcal{J}_i) - \mathrm{TPR} \big\rvert
               + \big\lvert \mathrm{FPR}(\mathcal{J}_i) - \mathrm{FPR} \big\rvert, \\
F_{OAE}^{(\mathcal{J}_i)} &= \big\lvert \mathrm{ACC}(\mathcal{J}_i) - \mathrm{ACC} \big\rvert .
\end{aligned}
\end{equation}

where, $ACC$ represents the fraction of samples correctly predicted, positive rate $PR$ represents the fraction of samples predicted fake irrespective of their true label, $TPR$ is the True Positive Rate and $FPR$ is the False Positive Rate.
$F_{EOP}$ and $F_{EO}$ are the means of these per-subgroup values over $\mathcal{J}$, so the subgroup columns decompose the group number exactly. $F_{DP} = \max_i PR(\mathcal{J}_i) - \min_i PR(\mathcal{J}_i)$ and $F_{OAE} = \max_i ACC(\mathcal{J}_i) - \min_i ACC(\mathcal{J}_i)$ are instead ranges between the extreme subgroups, so averaging their subgroup columns does not reproduce them.

\section{Additional Dataset Statistics}
\label{app:dataset_stats}
As described in the main paper, we base our experiments on AI-Face-FairnessBench \citep{lin2025aiface} which is a curated benchmark designed for evaluating algorithmic fairness in deepfake face detection and provides both real and fake faces along with demographic annotations for multiple protected attributes (as described in table \ref{tab:app_demographics}). This benchmark provides a total of $400{,}885$ real and $1{,}245{,}660$ fake face images generated using different GANs, DMs or hybrid methods. There are 22 subsets of fake face images along with two subsets of real images, FFHQ and IMDB-Wiki. Out of these 22 subsets, FaceForensics, DFD, CelebDF and DFDC include real images as well. To construct our chronological dataset setup, we consider 11 out of these 22 sequences. We discard the subsets that contribute very small amount of fake data based on nuanced generator differences. Each of these 11 considered sequences constitute a distinct ``task'' (or respectively discrete time-step) in our continual learning setup. The tasks are then ordered by the dates the dataset sources (as provided in the benchmark) or corresponding publications were made available publicly. An overview of these tasks are provided in table \ref{tab:temporal_slices}.
\begin{table}[h!]
    \centering
    \begin{tabular}{|c|c|}
        \hline
        \makecell{\shortstack{Demographic \\ Attributes}} & Description \\
        \hline
        Gender & 1 - Male, 0 - Female \\ \hline
        \multirow{2}{*}{Age}   & 0 - Child, 1 - Youth, 3 - Adult,  \\ & 4 - Middle-aged, 5-Senior \\
        \hline
        Skin Tone & According to Monk Skin Tone Scale \\ \hline
        \multirow{2}{*}{Intersection}   & 0-(Female,Light), 1-(Female,Medium),  \\ & 2-(Female,Dark), 3-(Male,Light), \\ & 4-(Male,Medium), 5-(Male,Dark)\\
        \hline
    \end{tabular}
    \caption{Summary of demographic attributes provided by AI‑Face‑FairnessBench and used in our fairness analysis.}
\label{tab:app_demographics}
\end{table}
To maintain adequate representations of real faces in training, we fetch the real data for each tasks wherever necessary using the real datasets, as specified in the original papers that introduced the respective deepfake generators. To this end, we consider two internal sources, FFHQ and IMDB-Wiki and three external sources, CelebA \citep{liu2015faceattributes}, MetFaces \citep{10.5555/3495724.3496739} and Casia Webfaces \citep{Yi2014LearningFR}. An overview of these sources are provided in table \ref{tab:external_internal_real_sources}. We consider each task as a two-class (fake vs. real) classification problem and all models are trained and their accuracy of predictions are evaluated without access to any demographic attributes, using only predefined training, validation or test splits respectively.

Demographic attributes are used in our temporal fairness evaluation across subgroups, without inclusion into training to ensure that protected attributes do not influence performance optimization. To balance the test splits, we either consider real images from the datasets included in the AI-Face-FairnessBenchmark if applicable, or construct such subsets (for VQGAN, Palette and DCFace) by accumulating all real images from the test sets of all previously seen tasks. No externally fetched real images have been used in fairness evaluation due to lack of demographic annotations.

\begin{table*}[h!]
    \small
    \centering
    \begin{tabular}{c ccc ccc}
    \toprule
    \multirow{2}{*}{\textbf{Datasets}} &
    \multicolumn{3}{c}{\textbf{Train}} &
    \multicolumn{3}{c}{\textbf{Test}} \\
    \cmidrule(lr){2-4} \cmidrule(lr){5-7}
     & \textbf{Real} & \textbf{Fake} & \textbf{Total}
     & \textbf{Real} & \textbf{Fake} & \textbf{Total} \\
    \midrule

    FaceForensics (FF++) & 17{,}218 & 84{,}346 & 101{,}564 & 4{,}375 & 21{,}233 & 25{,}608 \\
    DFD                 & 37{,}231 & 55{,}099 & 92{,}330  & 9{,}461 & 13{,}823 & 23{,}284 \\
    CelebDF             & 18{,}883 & 124{,}647 & 143{,}530 & 4{,}762 & 30{,}842 & 35{,}604 \\
    DFDC                & 30{,}136 & 30{,}038 & 60{,}174  & 7{,}700 & 7{,}494 & 15{,}194 \\

    \bottomrule
    \end{tabular}
    \caption{FaceForensics, DFD, CelebDF and DFDC are the four datasets which have real data already included and provided as part of AI-Face-FairnessBench. Here we include a detailed overview of the number of samples provided for each of these datasets. $10\%$ of every respective training set is held out as validation set.}
    \label{tab:appendix_real_sets}
\end{table*}

\newcounter{rowno}
\begin{table*}[h!]
\centering
\small
\begin{tabular}{c c c c ccc c}
\toprule
\multirow{3}{*}{\makecell{\textbf{SL.} \\ \textbf{No.}}} &
\multirow{3}{*}{\makecell{\textbf{Dataset} \\ \textbf{Publication} \\ \textbf{Date}}} &
\multirow{3}{*}{\makecell{\textbf{Dataset or} \\ \textbf{Generator} \\ \textbf{Used}}} &
\multirow{3}{*}{\makecell{\textbf{Type of} \\ \textbf{Generation}}} &
\multicolumn{2}{c}{\textbf{Samples}} &
\multirow{3}{*}{\makecell{\textbf{Real Dataset} \\ \textbf{Used} \\ \textbf{in Training}}} \\
&
 &
 &
 &
 & \\
\cmidrule{5-6}
 &
 &
 &
 & \textbf{Train} & \textbf{Test}
 & \\
\toprule

\stepcounter{rowno}\arabic{rowno} & 25.01.19 & FaceForensics (FF++) & Hybrid &  101{,}564 & 25{,}608 & NA\\
\stepcounter{rowno}\arabic{rowno} & 24.09.19 & DFD & Hybrid & 92{,}330 & 23{,}284 & NA \\
\stepcounter{rowno}\arabic{rowno} & 27.09.19 & CelebDF & Hybrid  &  143{,}530 & 35{,}604 & NA \\

\stepcounter{rowno}\arabic{rowno} & 11.12.19 & DFDC & Hybrid  &  60{,}174 & 15{,}194 & NA \\
\stepcounter{rowno}\arabic{rowno} & 17.12.20 & VQGAN & GAN 	& 39{,}943 & 10{,}057  & CelebA, MetFaces \\
\stepcounter{rowno}\arabic{rowno} & 24.01.21 & StyleGAN & GAN  & 7{,}978 & 2{,}021 & FFHQ \\
\stepcounter{rowno}\arabic{rowno} & 11.03.22 & Palette & DM  & 4{,}804 & 1{,}196 & CelebA, MetFaces \\
\stepcounter{rowno}\arabic{rowno} & 03.09.22 & StyleGAN2 & GAN   & 94{,}660 & 23{,}698  & FFHQ \\
\stepcounter{rowno}\arabic{rowno} & 14.04.23 & DCFace & DM  & 424{,}452 & 105{,}529 & Casia Webfaces \\
\stepcounter{rowno}\arabic{rowno} & 25.04.23 & StyleGAN3 & GAN & 21{,}416 & 5{,}354  & FFHQ \\
\stepcounter{rowno}\arabic{rowno} & 05.09.23 & Stable Diffusion & DM  & 31{,}177 & 7923  & wiki \\
\bottomrule
\end{tabular}

\caption{Task definitions and chronological ordering used across all our experiments. $10\%$ of every respective training set is held out as validation set. This shows the corresponding date of publication of the dataset sources as provided in \citep{lin2025aiface}. Dates refer to the public release of the images, which for some generators
(e.g., StyleGAN) may lag the generator's first publication. The table also shows real data sources used for training and/or evaluation stage. A detailed overview of these real datasets are provided in table \ref{tab:external_internal_real_sources}.}
\label{tab:temporal_slices}
\end{table*}

\begin{table*}[h!]
\centering

\small
\begin{tabular}{ccccccc}
\toprule
\textbf{Dataset} & \textbf{Source} & \textbf{\# Images} & \textbf{Generators} & \textbf{Training} & \textbf{Validation} & \textbf{Test}\\
\toprule

\multirow{2}{*}{CelebA} & \multirow{2}{*}{\makecell{https://mmlab.ie.cuhk.edu.hk/\\projects/CelebA.html}} & \multirow{2}{*}{202{,}599} &  VQGAN & 35{,}347 & 3{,}928 & NA\\
    \cmidrule(lr){4-7}
    &  &  & Palette & 3{,}722 & 414 & NA\\
\midrule
\multirow{2}{*}{MetFaces} & \multirow{2}{*}{https://github.com/NVlabs/metfaces-dataset}& \multirow{2}{*}{1{,}336} &  VQGAN & 601 & 67 & NA\\
    \cmidrule(lr){4-7}
    & & & Palette & 601 & 67 & NA\\
\midrule
\makecell{\shortstack{CASIA \\ Webfaces}} & \makecell{https://github.com/deepinsight/insightface}
& 490{,}623 & DCFace & 382{,}007 & 42{,}445 & NA\\
\midrule
\multirow{3}{*}{FFHQ} & \multirow{3}{*}{AI-Face-FairnessBench} & & StyleGAN  & 3{,}244 & 360 & 908\\
    \cmidrule(lr){4-7}
    &  & & StyleGAN2 & 38{,}490 & 4{,}277 & 10{,}641 \\
    \cmidrule(lr){4-7}
    & & & StyleGAN3 & 8{,}708 & 968 & 2{,}404 \\
\midrule

Wiki & AI-Face-FairnessBench & & Stable Diffusion &  23{,}833 & 2{,}648 & 6{,}668  \\
\bottomrule
\end{tabular}

\caption{We fetched real data for generators if needed from external datasets or internal ones in AI-Face-FairnessBench. External data are not considered for testing as they do not have demographic annotations available. Here we include a detailed overview of the number of samples provided for each of these datasets. $10\%$ of every respective training set is held out as validation set unless otherwise mentioned.}
\label{tab:external_internal_real_sources}
\end{table*}

\begin{table}[t]
\centering
\small
\setlength{\tabcolsep}{4pt}
\begin{tabular}{rl rr rr rr}
\toprule
 & & \multicolumn{2}{c}{Train} & \multicolumn{2}{c}{Validation} & \multicolumn{2}{c}{Test} \\
\cmidrule(lr){3-4} \cmidrule(lr){5-6} \cmidrule(lr){7-8}
\# & Dataset & Real & Fake & Real & Fake & Real & Fake \\
\midrule
1  & FaceForensics++  &  15,496 &  75,911 &   1,722 &   8,435 &   4,375 &  21,233 \\
2  & DFD              &  33,508 &  49,589 &   3,723 &   5,510 &   9,461 &  13,823 \\
3  & CelebDF          &  16,995 & 112,182 &   1,888 &  12,465 &   4,762 &  30,842 \\
4  & DFDC             &  27,122 &  27,034 &   3,014 &   3,004 &   7,700 &   7,494 \\
5  & VQGAN            &  35,948 &  35,949 &   3,995 &   3,994 &  26,298 &  10,057 \\
6  & StyleGAN         &   3,244 &   7,180 &     360 &     798 &     908 &   2,021 \\
7  & Palette          &   4,323 &   4,324 &      481 &     480 &  27,206 &   1,196 \\
8  & StyleGAN2        &  38,490 &  85,194 &   4,277 &   9,466 &  10,641 &  23,698 \\
9  & DCFace           & 382,007 & 382,007 &  42,445 &  42,445 &  37,847 & 105,529 \\
10 & StyleGAN3        &   8,708 &  19,274 &     968 &   2,142 &   2,404 &   5,354 \\
11 & Stable Diffusion &  23,833 &  28,059 &   2,648 &   3,118 &   6,668 &   7,923 \\
\bottomrule
\end{tabular}
\caption{Number of real and fake images per task and split. For VQGAN, Palette and DCFace, real training and validation images come from external sources (Table \ref{tab:external_internal_real_sources}), and real test images are re-used from the test sets of earlier tasks.}
\label{tab:split_counts}
\end{table}

\section{Overview of Methods}
\label{app:method_overview}
Here we briefly summarize the continual learning methods evaluated in the main body of this work.

\paragraph{Baselines}
The naive sequential acts as a lower bound as it trains the model on each subsequent task without any explicit continual learning strategy. Whereas the accumulated baseline trains the model on the accumulation of all the data observed up to the current tasks and serves as an upper bound for performance in both detection and fairness stability.
\paragraph{EWC: } We adopt Elastic Weight Consolidation (EWC) \citep{Kirkpatrick2017EWC} as our first continual learning strategy that aims to mitigate catastrophic forgetting by constraining parameter updates according to their importance for previously learned tasks. The core idea is to selectively regularize updates to network parameters, such that parameters that are critical for past tasks are preserved, while less important ones remain plastic. Concretely, EWC introduces a quadratic weight regularization term into the loss function, which penalizes deviations of each parameter from its previously learned value in proportion to its estimated importance, with the regularization strength additionally weighted by an experimentally determined hyper-parameter $\lambda$. The importance is quantified using the Fisher Information Matrix (FIM), which captures the model's sensitivity to changes in each parameter. We subsequently approximate FIM by considering only the diagonal of the hessian matrix.

\paragraph{DER++: }
Dark Experience Replay (DER) \citep{buzzega2020darkexperiencegeneralcontinual} is a rehearsal-based continual learning strategy that improves upon the standard Experience Replay by distilling past knowledge along the same training trajectory. Instead of relying solely on storing raw input instances, DER preserves the model’s past predictions (logits) for buffered samples and enforces consistency between current and stored logits during training. This logit-matching objective has been shown to be more effective than replaying raw examples alone, as it directly constrains the functional behavior of the model and stabilizes optimization across tasks.
However, simple reservoir sampling can limit the effectiveness of logit distillation when buffer coverage is sparse. To address this, DER++ augments the original objective with an additional supervised loss term on the buffered samples, using their ground-truth labels alongside the logit-matching regularization. This combination balances knowledge distillation with explicit supervision, yielding improved stability and performance over the base DER strategy in continual learning settings.
\paragraph{GradMatch: } GradMatch \citep{killamsetty2021gradmatchgradientmatchingbased} is another replay-based continual learning strategy that populates the replay buffer with a carefully selected subset of training samples whose gradients closely approximate those of a reference validation set. The key idea is to preserve samples that are most representative in terms of their contribution to the optimization dynamics, rather than relying on random or heuristic sampling. Formally, GradMatch formulates sample selection as a gradient matching problem, where the objective is to minimize the discrepancy between the gradient induced by a small coreset and the gradient computed over the full dataset. To solve this problem efficiently, GradMatch employs the Orthogonal Matching Pursuit (OMP) algorithm, which greedily selects samples that best reduce the gradient mismatch. This results in a compact yet informative replay buffer that better preserves past knowledge and improves data efficiency during continual training

\paragraph{SPrompts: } As a representative prompt-based continual learning strategy, we adopt S-Prompt \citep{wang2022sprompt} to sequentially train the model across generator domains. S-Prompt builds on the idea of learning domain-specific prompts on top of a pre-trained Vision Transformer, thereby eliminating the need to store exemplars from previous domains. In this framework, the frozen pre-trained backbone is responsible for extracting low-level, domain-agnostic visual features, while the learnable prompts capture high-level, domain-specific knowledge. During continual training, the entire Vision Transformer backbone is kept fixed, and only the prompts corresponding to the current domain, together with the classifier, are optimized. This design substantially reduces training-time computational overhead while simultaneously improving training stability by preventing destructive updates to shared representations. To facilitate domain identification at inference time, K-means clustering is used to store representative centroids for each training domain. At test time, a k-nearest neighbors (kNN) search is performed to associate an input image with its nearest domain centroid, enabling the selection of the appropriate prompt for prediction.
\begin{table*}[t]
    \centering
    \small
    \begin{tabular}{lcccc}
        \toprule
        Methods & Epochs & Learning Rate & Buffer/ Prompt Size & Specific Parameter\\
        \toprule
        Accumulated & 25 & $1\times10^{-5}$ & NA & NA\\
        Naive & 25 & $1\times10^{-5}$ & NA & NA \\
        EWC & 25 & $1\times10^{-5}$ & NA & $\lambda = 5000$\\
        DER++ & 25 & $1\times10^{-5}$ & Buffer Size = 3000 & $\alpha = 0.8, \beta = 0.5$\\
        GradMatch & 25 & $1\times10^{-5}$ & Buffer Size = 3000 & $\beta = 0.5$\\
        SOYO & 25 & $5\times10^{-4}$ & Prompt Length = 10 & NA \\
        PINA & 25 & $1\times10^{-4}$ & Prompt Length = 10 & NA \\
        S-Prompts & 25 & $1\times10^{-4}$ & Prompt Length = 10 & NA \\
        \bottomrule
    \end{tabular}
    \caption{Overview of method specific and shared hyperparameter settings used in all our experiments.}
    \label{tab:app_hyperparam}
\end{table*}
\paragraph{PINA: }
Much like S-Prompt, PINA \citep{wang2024non} adopts a prompt-based paradigm to realize an exemplar-free domain-incremental learning strategy. The method is designed to sequentially train a model under domain shifts by explicitly disentangling generic and domain-specific knowledge. To this end, PINA introduces three key components:
(1) a Unified Classifier (UC) that captures domain-agnostic classification knowledge shared across all domains;
(2) a Domain-Specific Alignment (DSA) module, which is responsible for aligning and integrating classes from newly encountered domains with the semantic space of the base domain; and
(3) a Patch Shuffle Selector (PSS) that serves as a domain label predictor, enabling the selection of appropriate domain-specific parameters at inference time.
During training on the base domain, PINA optimizes only the Unified Classifier while introducing a learnable prompt token that is concatenated with the class token before feature extraction by the Vision Transformer. For subsequent incremental domains, the backbone remains fixed and the DSA module is employed to acquire and encode supplementary domain-specific knowledge. This modular design allows PINA to effectively adapt to new domains while preserving previously learned representations, without relying on exemplar storage.
\paragraph{SOYO: } The final method considered is SOYO \citep{wang2025boosting}, a parameter-isolation strategy for domain-incremental learning. It employs a lightweight, trainable framework to select optimal PIDIL parameters by predicting domain labels from domain-specific features. To address the unavailability of data from previous domains during later training stages, SOYO introduces a Gaussian Mixture Compressor, which summarizes features from previous domains using a compact set of Gaussian components, and a Domain Feature Resampler, which generates pseudo-domain features through random sampling. Additionally, SOYO incorporates a Multilevel Domain Feature Fusion Network that integrates shallow spatial features with deep semantic features, resulting in more discriminative domain representations.

\section{Implementation Details and Hyperparameters}
\label{app:hyperparams}
We employ a Vision Transformer backbone (ViT) pretrained on ImageNet-21k (14M images, 21{,}843 classes) and finetuned on ImageNet(1M images and 1k classes). Every input image is resized to $224\times224$ pixels. In general, training is performed using a batch size of 128 up to 25 epochs, with early stopping triggered by the accuracy on validation set unless otherwise mentioned.  For instance, we chose $\lambda=5000$ based one preliminary tuning as smaller values led to higher forgetting in earlier tasks and too large (like, $10000$) overly constrained the updates. We selected $\alpha=0.8$ and $\beta=0.5$ after initial examination with lower values to prioritize knowledge preservation. For prompt based methods, hyper parameters are adjusted accordingly.

\begin{figure*}[t]
    \centering
    \subfloat[Accumulated Training]{
        \includegraphics[width=0.48\textwidth]{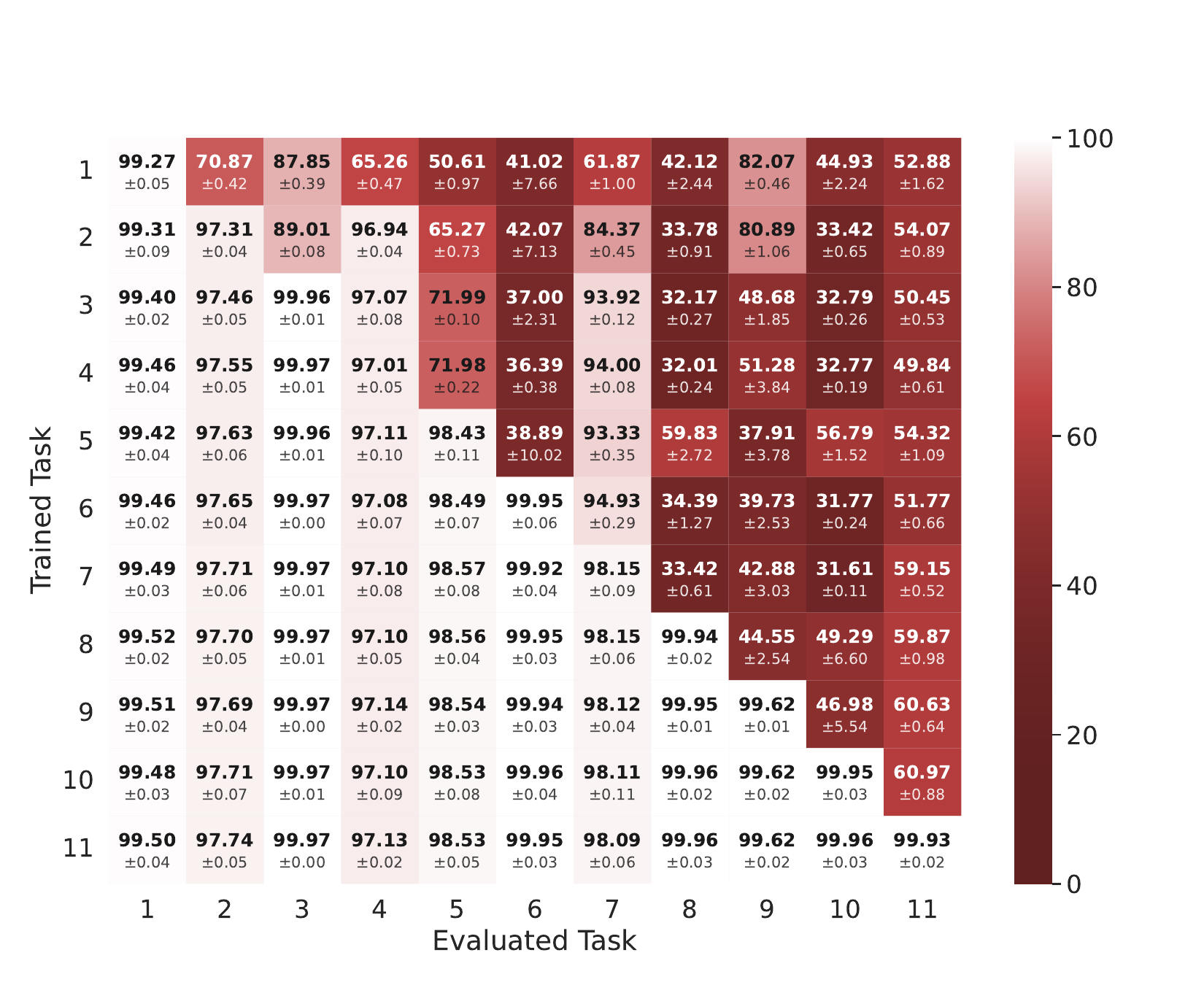}
        \label{fig:accumulated_heatmap}

    }
    \hfill
    \subfloat[Naive Sequential Training]{
        \includegraphics[width=0.48\textwidth]{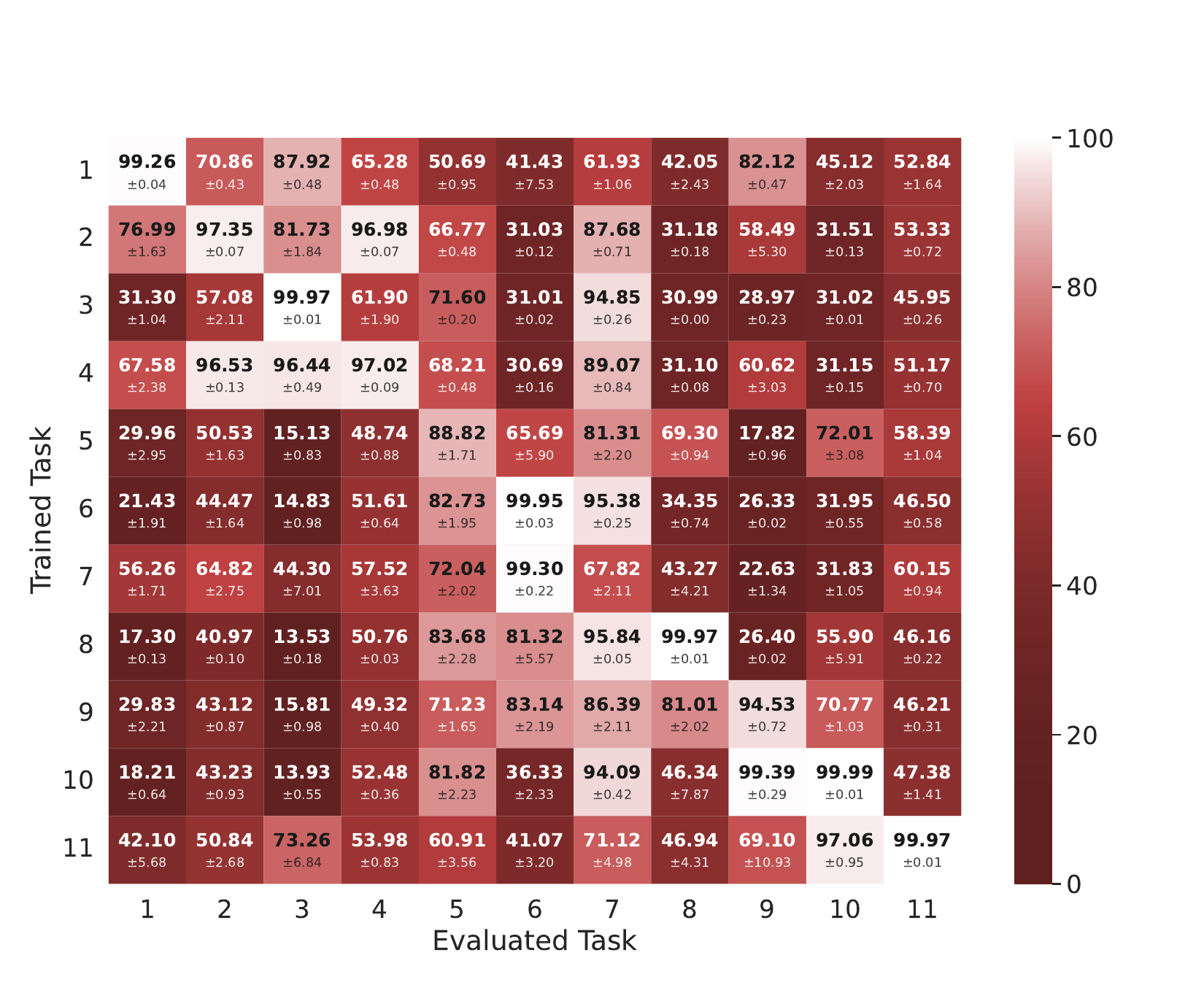}
        \label{fig:naive_heatmap}
    }
    \caption{Inter-task accuracy matrix (mean $\pm$ std over five seeds) for the accumulated upper-bound and naive sequential lower-bound by training step (y-axis) and evaluated task (x-axis). The matrix highlights that each task is solvable in itself (diagonal values), but accuracy on generators not yet seen in training (top-right off-diagonal) remains limited and varies strongly across generators.}
    \label{fig:appendix_heatmaps_baselines}
\end{figure*}
\begin{figure*}[p]
    \centering
    \subfloat[EWC]{
        \includegraphics[width=0.46\textwidth]{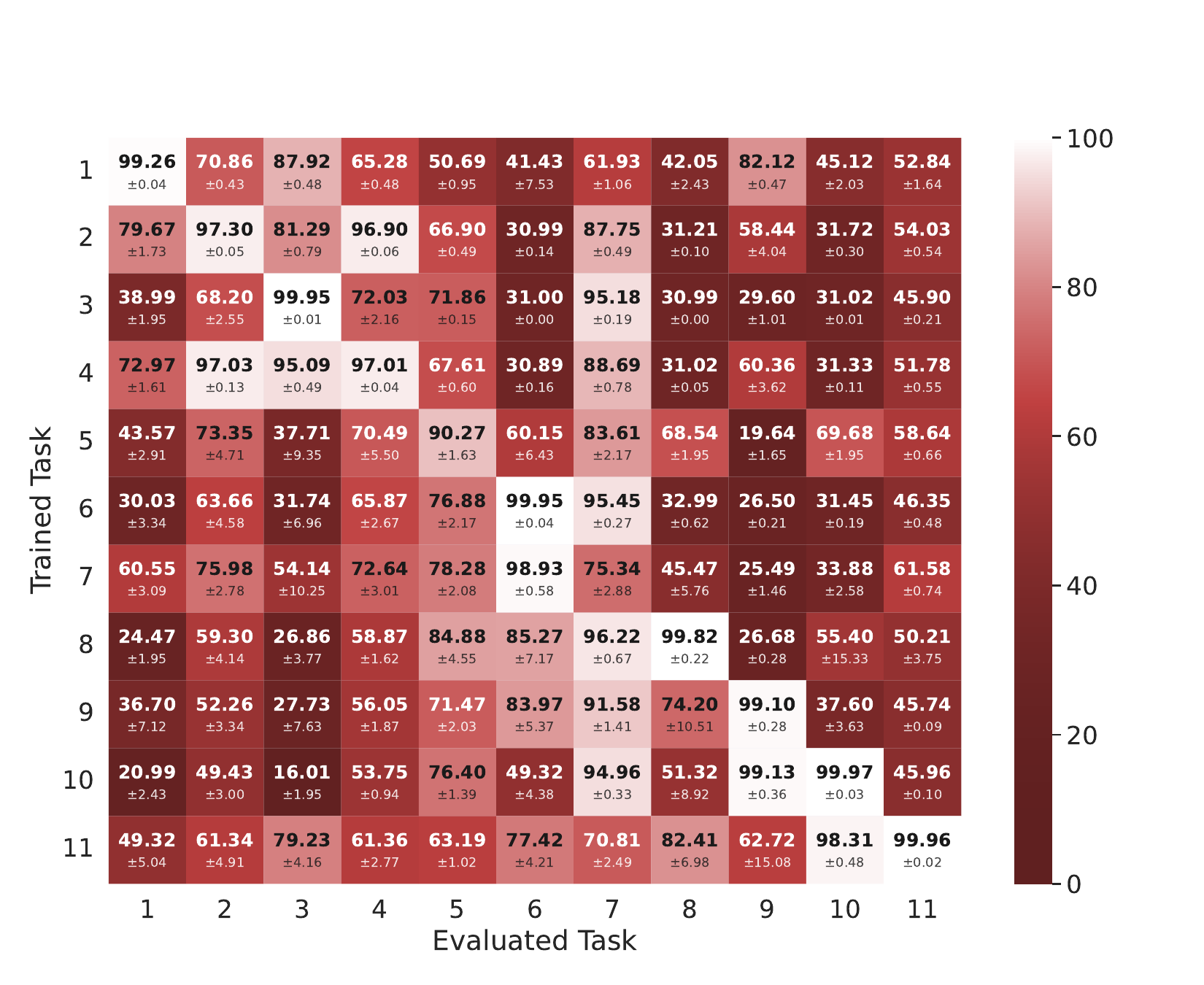}
        \label{fig:ewc_heatmap}
    }
    \hfill
    \subfloat[GradMatch]{
        \includegraphics[width=0.46\textwidth]{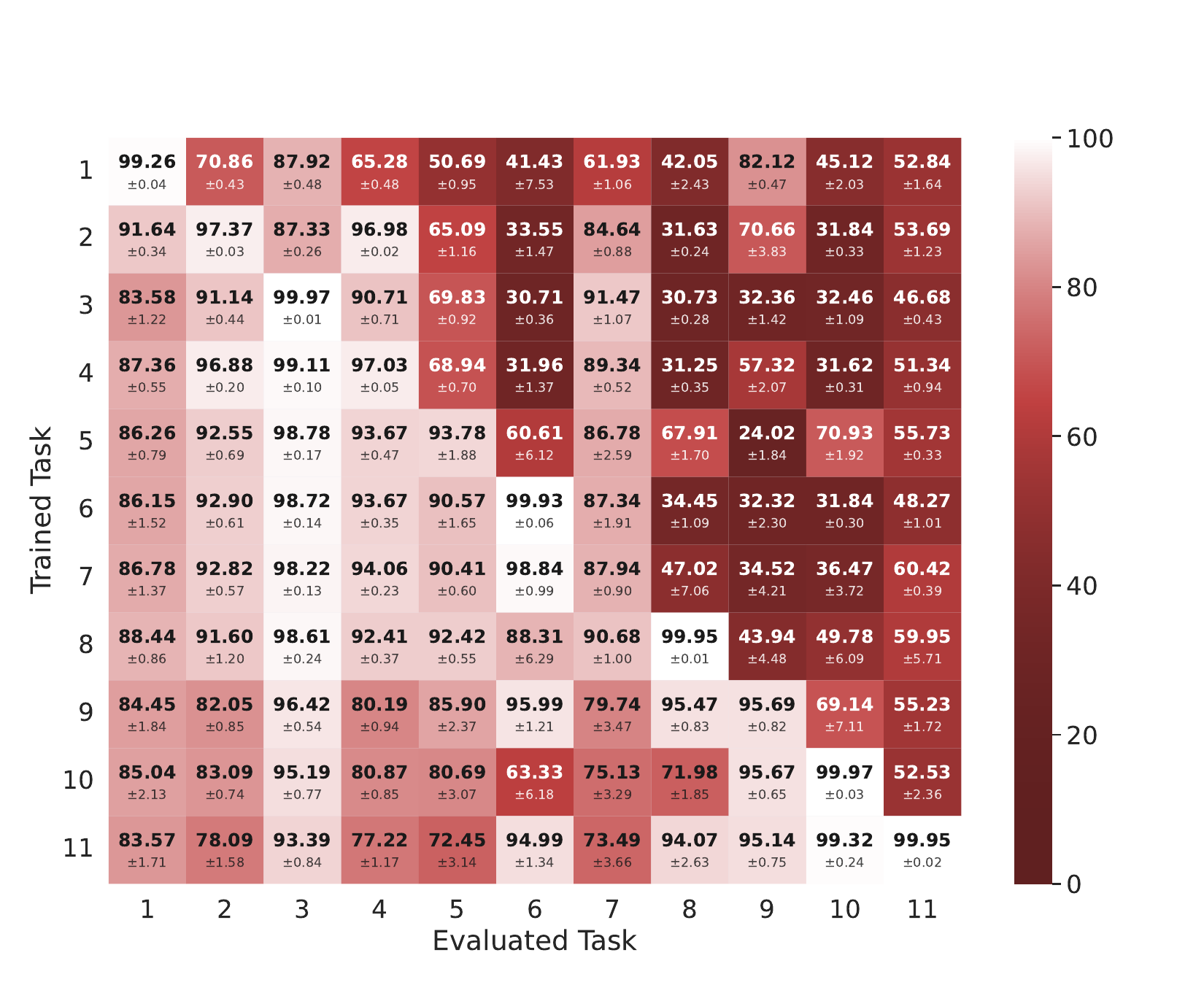}
        \label{fig:gradmatch_heatmap}
    }
    \vfill
    \subfloat[PINA]{
        \includegraphics[width=0.46\textwidth]{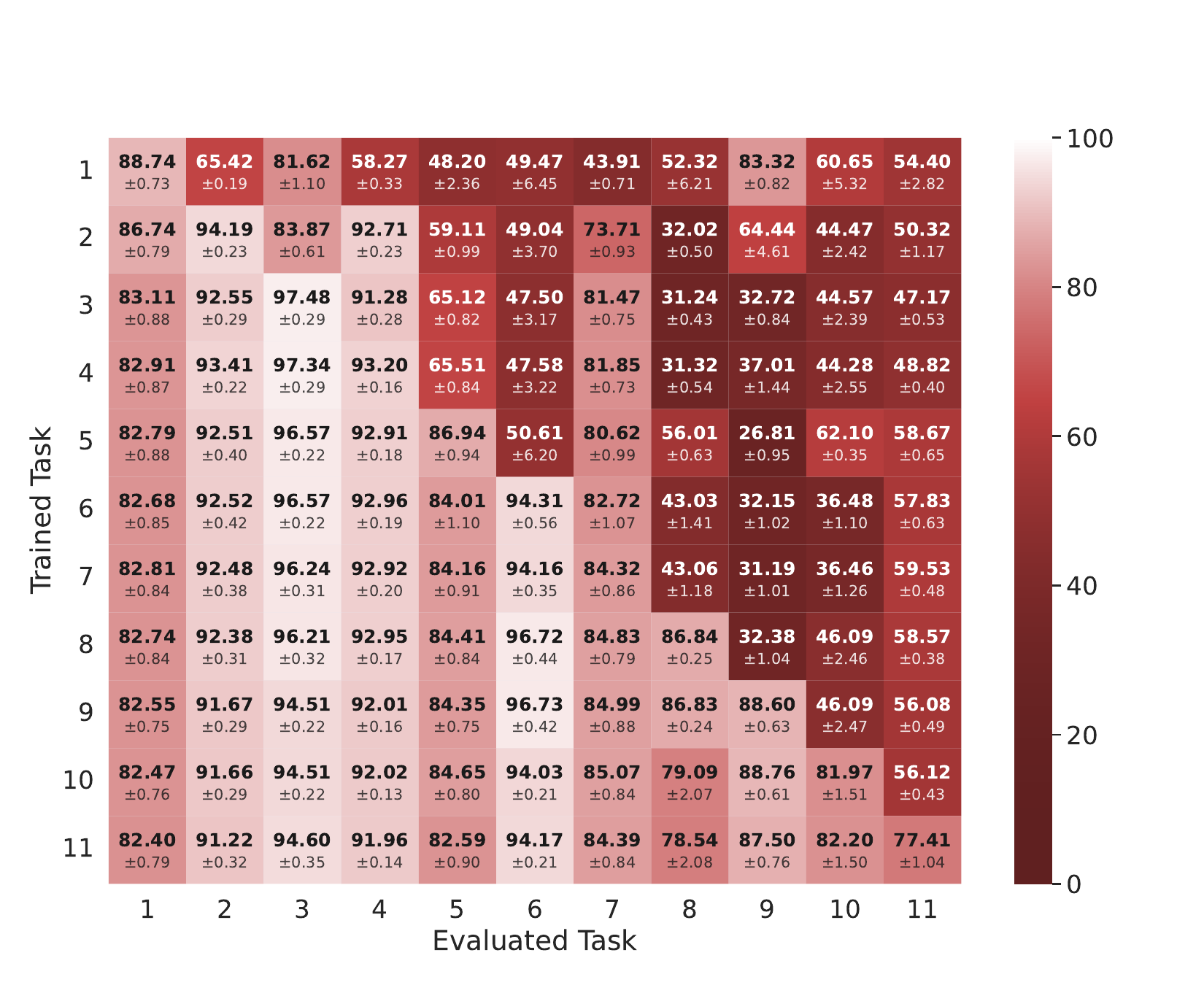}
        \label{fig:pina_heatmap}
    }
    \hfill
    \subfloat[SOYO]{
        \includegraphics[width=0.46\textwidth]{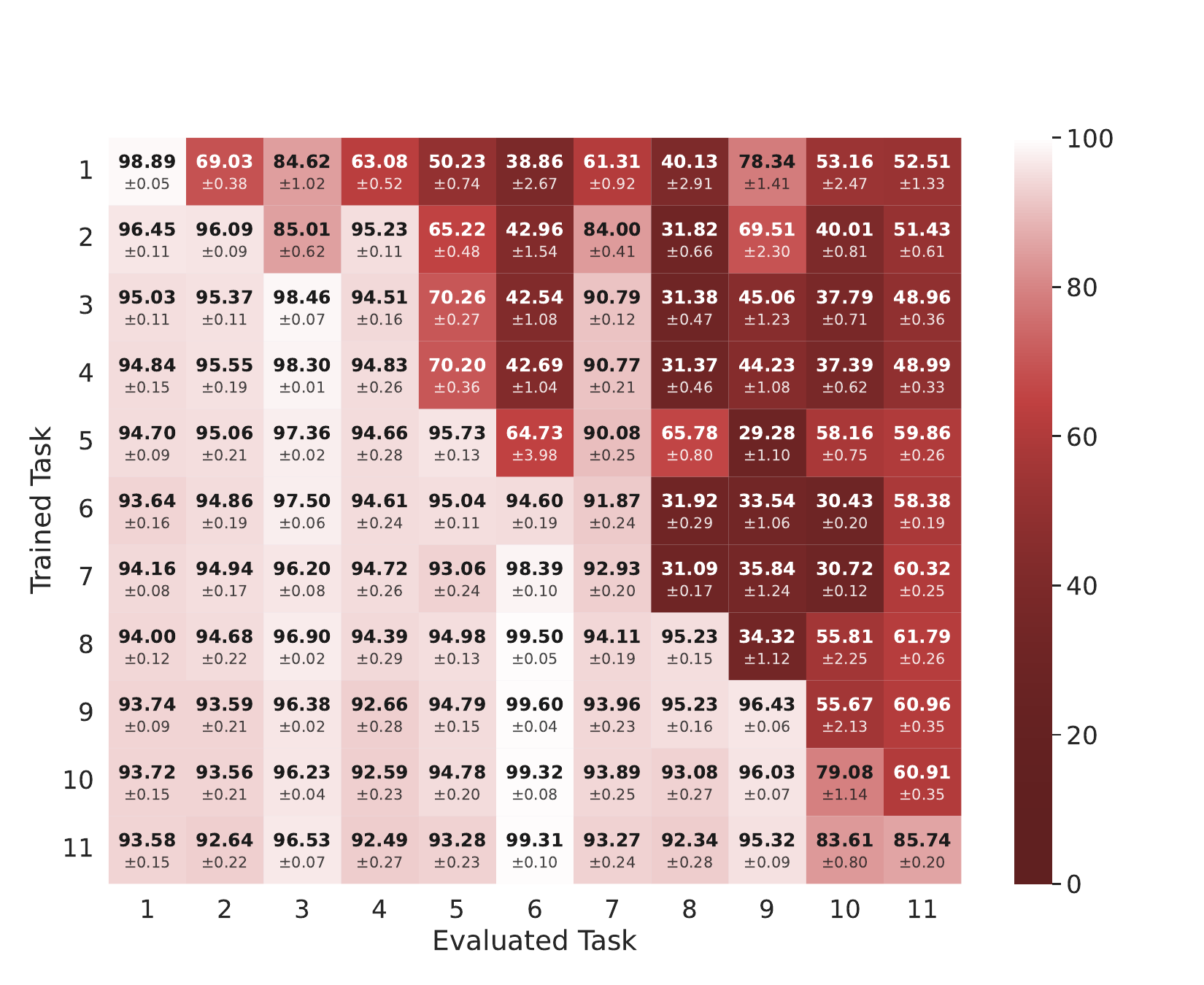}
        \label{fig:soyo_heatmap}
    }
    \vfill
    \subfloat[S-Prompts]{
        \includegraphics[width=0.46\textwidth]{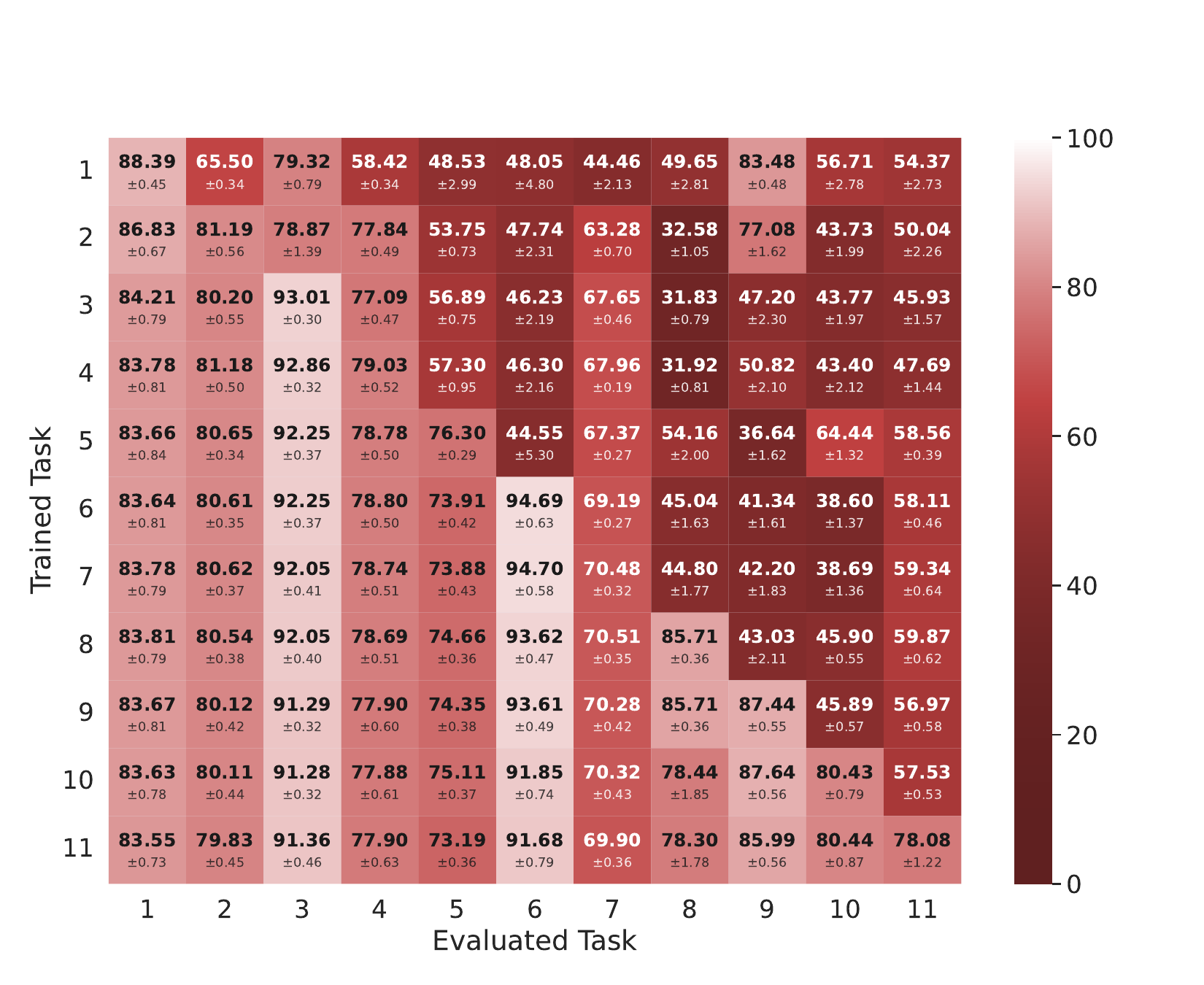}
        \label{fig:sprompts_heatmap}
    }
    \caption{Inter-task accuracy matrix (mean $\pm$ std over five seeds) for EWC, GradMatch, PINA, SOYO and S-Prompts by training step (y-axis) and evaluated task (x-axis). The matrix highlights that each task is solvable in itself (diagonal values), but accuracy on generators not yet seen in training (top-right off-diagonal) remains limited and varies strongly across generators. CL methods mitigate forgetting for most generators (bottom-left off-diagonal).}
    \label{fig:appendix_heatmaps}
\end{figure*}

\section{Analyzing Inter-Task Accuracy Matrices}
\label{app:acc-heatmaps}
This section complements the main body's findings with a brief overview of inter-task accuracies for additional continual learning (CL) methods that we have considered in our evaluations: SOYO, PINA, S-Prompt, EWC, and GradMatch, along with the accumulated and naive sequential baselines.
\begin{table*}[p]
\centering
\begin{subtable}{0.48\textwidth}
\centering
\includegraphics[width=\linewidth]{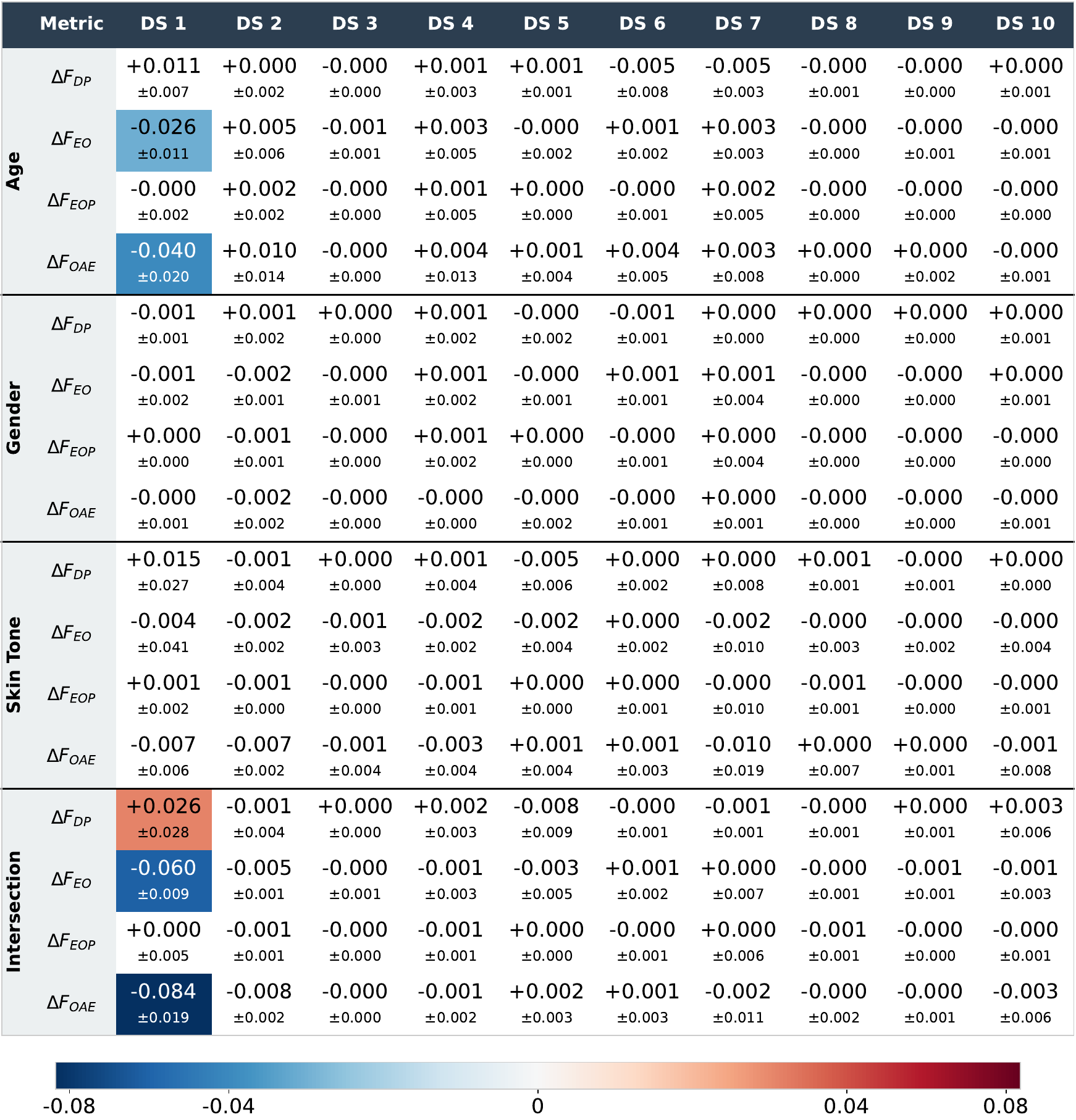}
\caption{Accumulated}
\label{tab:accumulated_per_ds}
\end{subtable}\hfill
\begin{subtable}{0.48\textwidth}
\centering
\includegraphics[width=\linewidth]{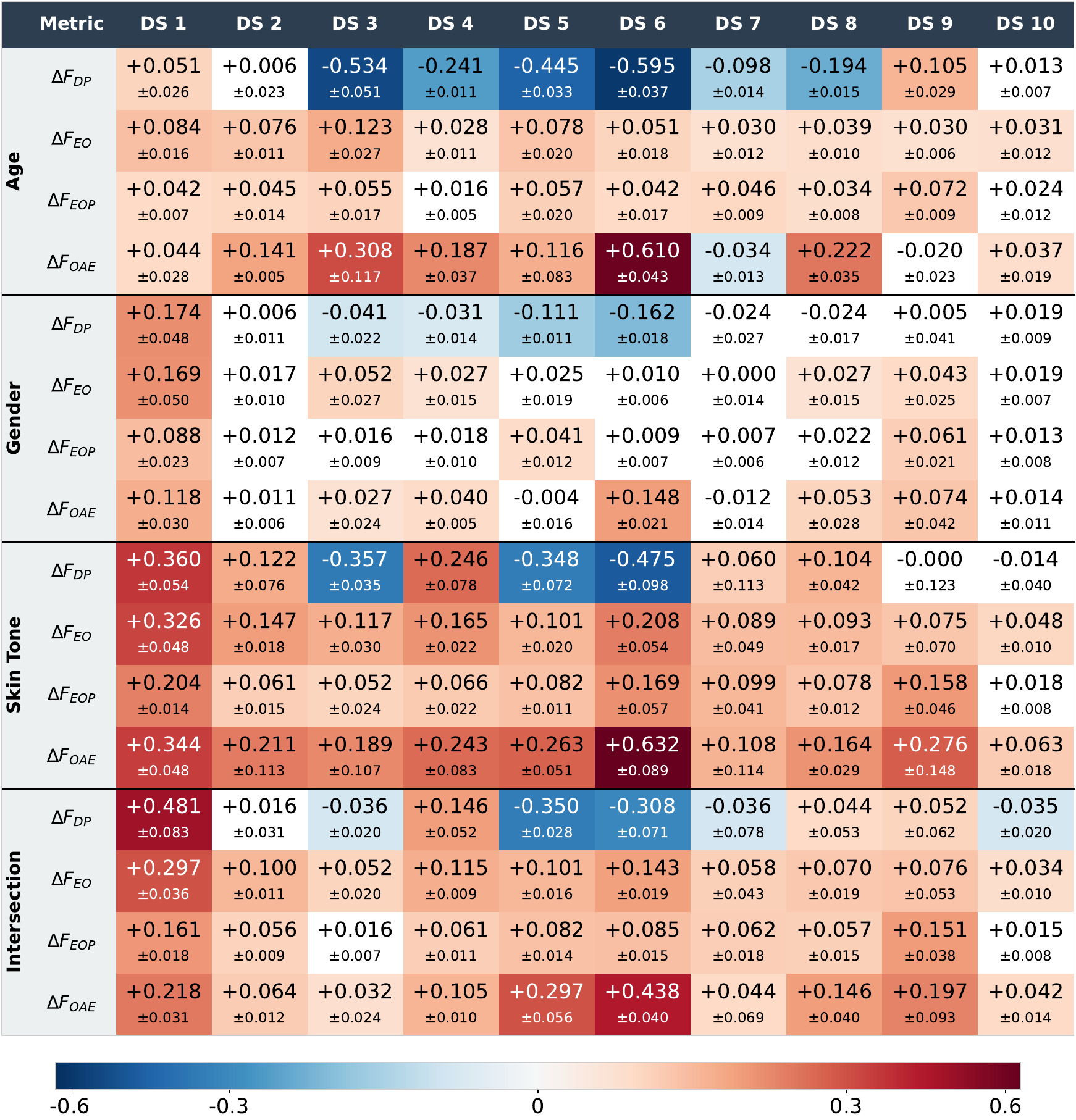}
\caption{Naive}
\label{tab:naive_per_ds}
\end{subtable}
\\[0.5cm]
\begin{subtable}{0.48\textwidth}
\centering
\includegraphics[width=\linewidth]{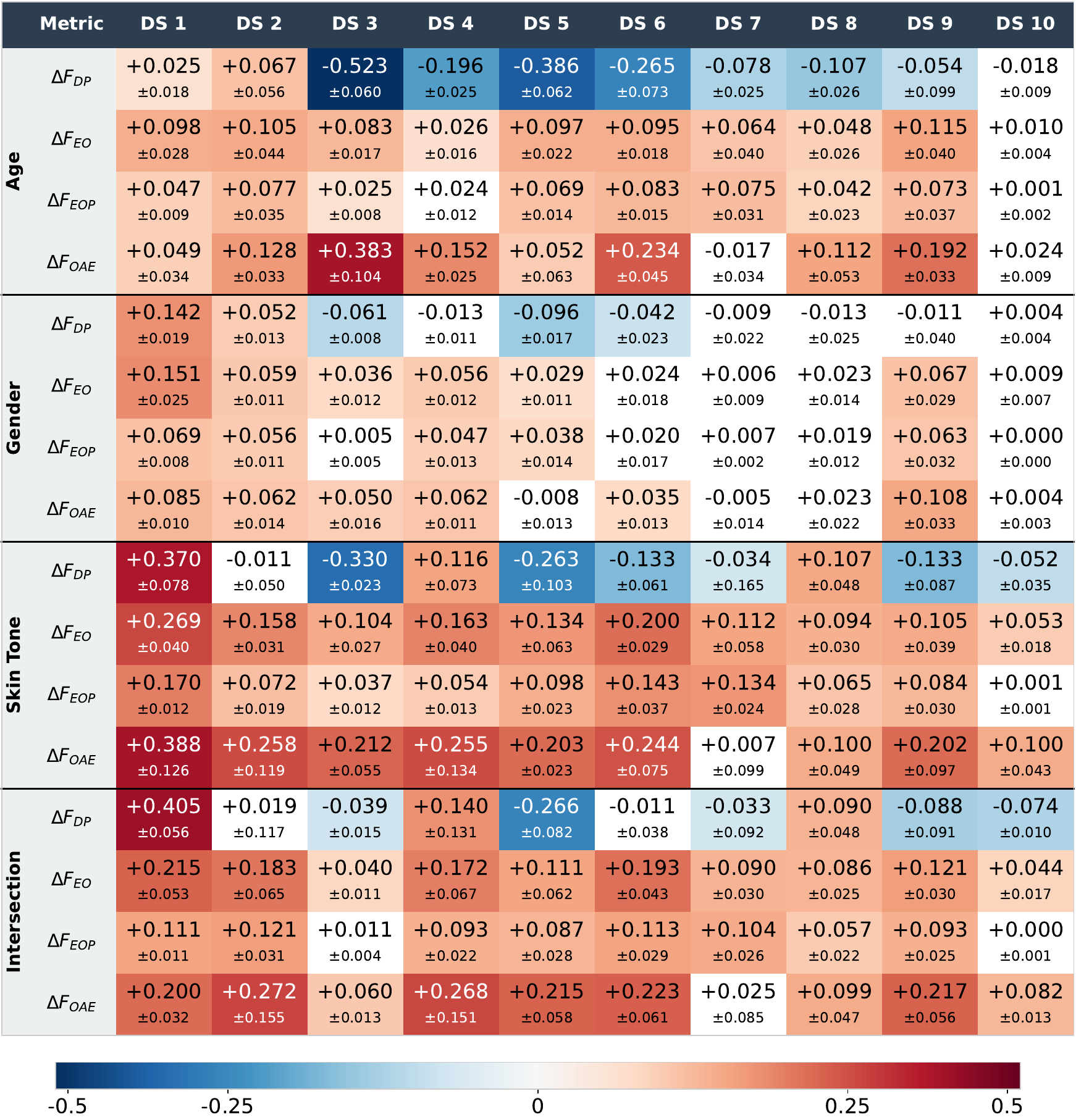}
\caption{EWC}
\label{tab:ewc_per_ds}
\end{subtable}\hfill
\begin{subtable}{0.48\textwidth}
\centering
\includegraphics[width=\linewidth]{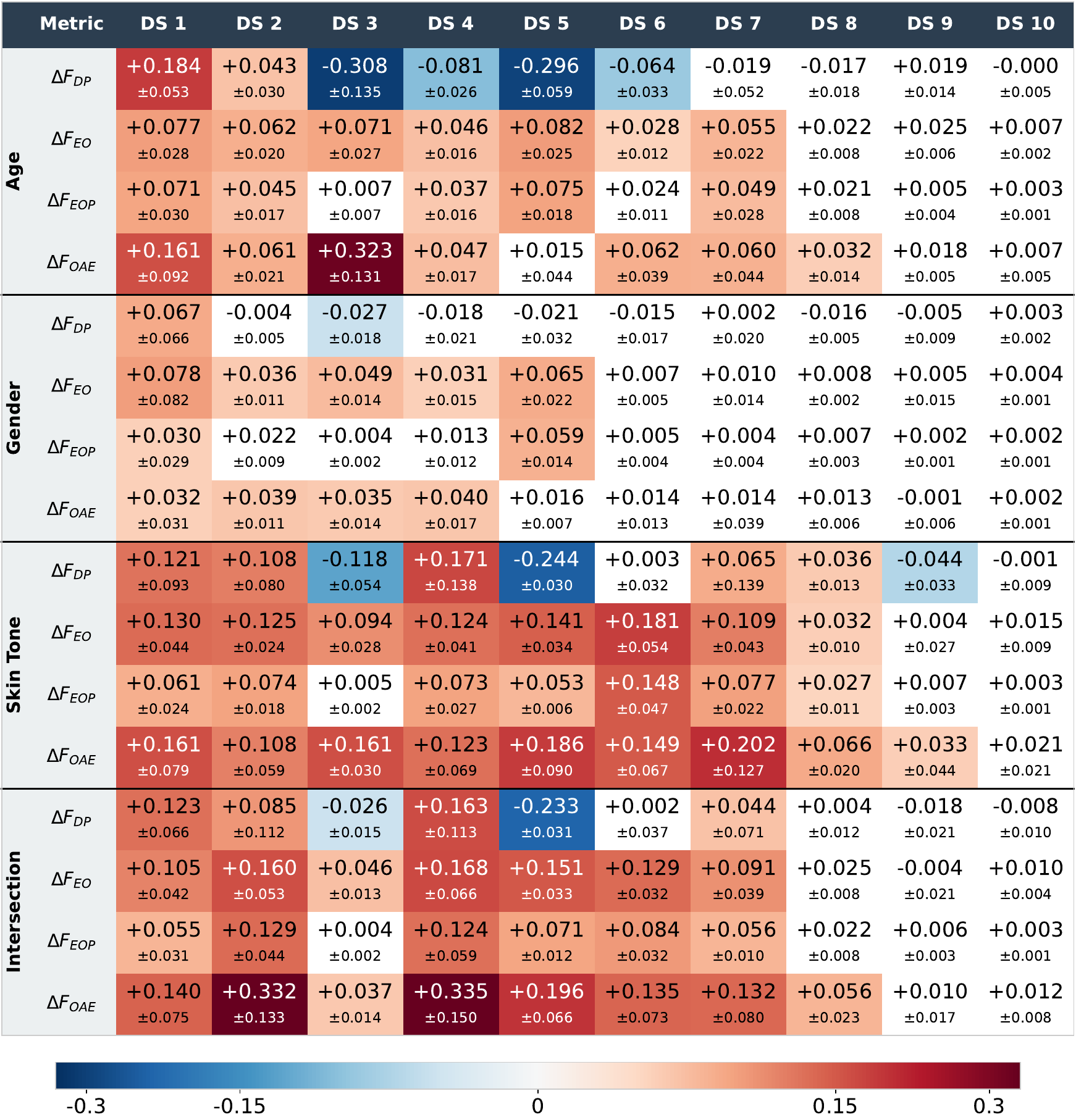}
\caption{GradMatch}
\label{tab:gm_per_ds}
\end{subtable}
\caption{Comparison of (Un)fairness at the end of continually learning across the 11 generator sequence with accumulated and naive sequential baselines along with EWC and GradMatch for test sets of all previously observed tasks, i.e., test set 1 after training on 11 tasks, test set 2 after 11 tasks, and so on. To focus on whether the CL algorithm has overall improved or deteriorated fairness, the shown values (mean $\pm$ std over five seeds) are differences of the final fairness with respect to the fairness obtained right after training on the respective generator, e.g.~discrepancy in fairness of task 2 after training on task 2 vs. task 2 after training on all 11 tasks. As before, red (blue) marks the many deteriorating (few improving) fairness dimensions. Cells remain white if the absolute difference is less than 2.5\% to ensure the figure remains legible.}
\label{tab:app_fairness_per_ds_baselines}
\end{table*}

\begin{table}[p]
\centering
\begin{subtable}{0.48\textwidth}
\centering
\includegraphics[width=\linewidth]{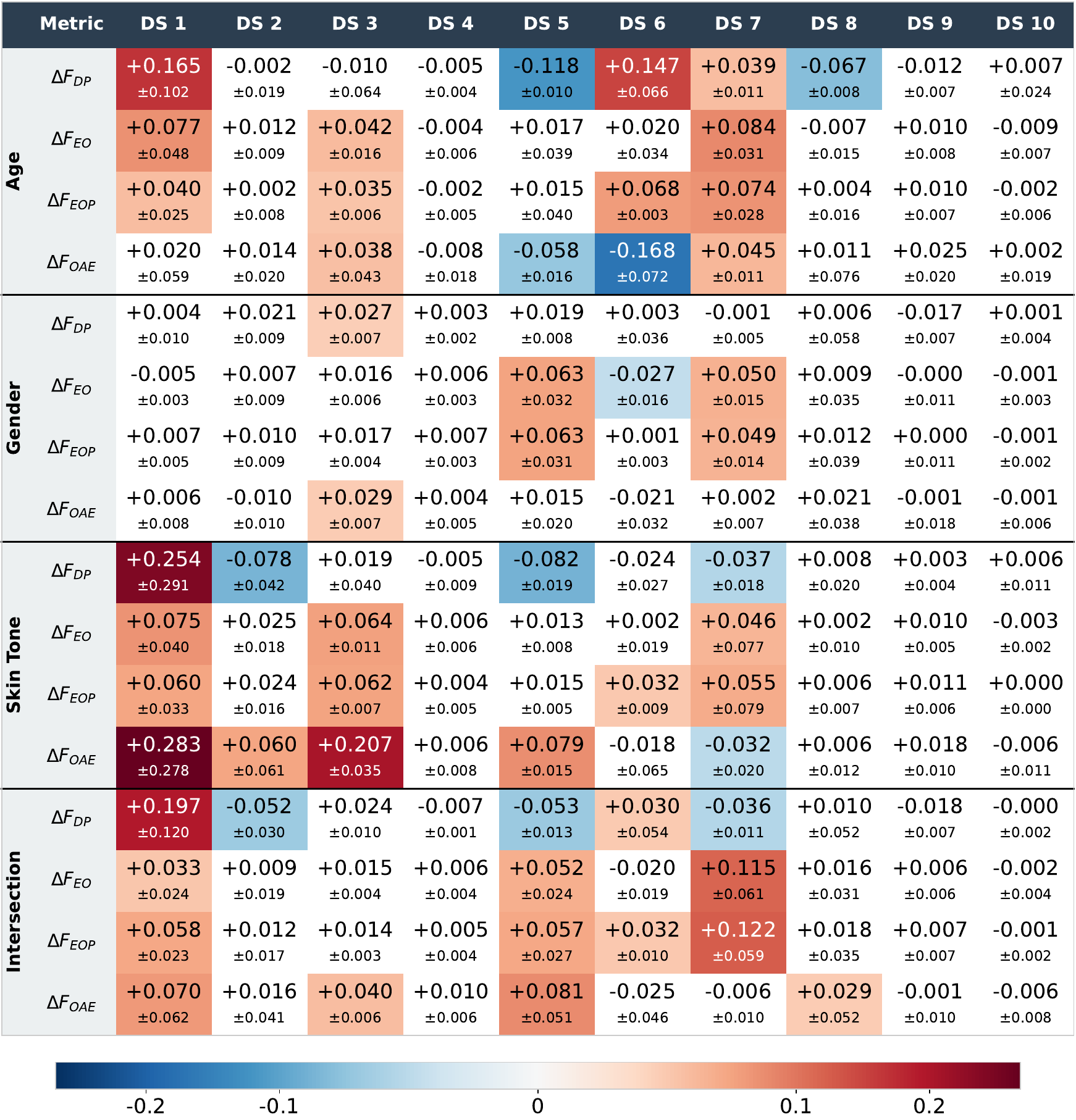}
\caption{PINA}
\label{tab:pina_per_ds}
\end{subtable}\hfill
\begin{subtable}{0.48\textwidth}
\centering
\includegraphics[width=\linewidth]{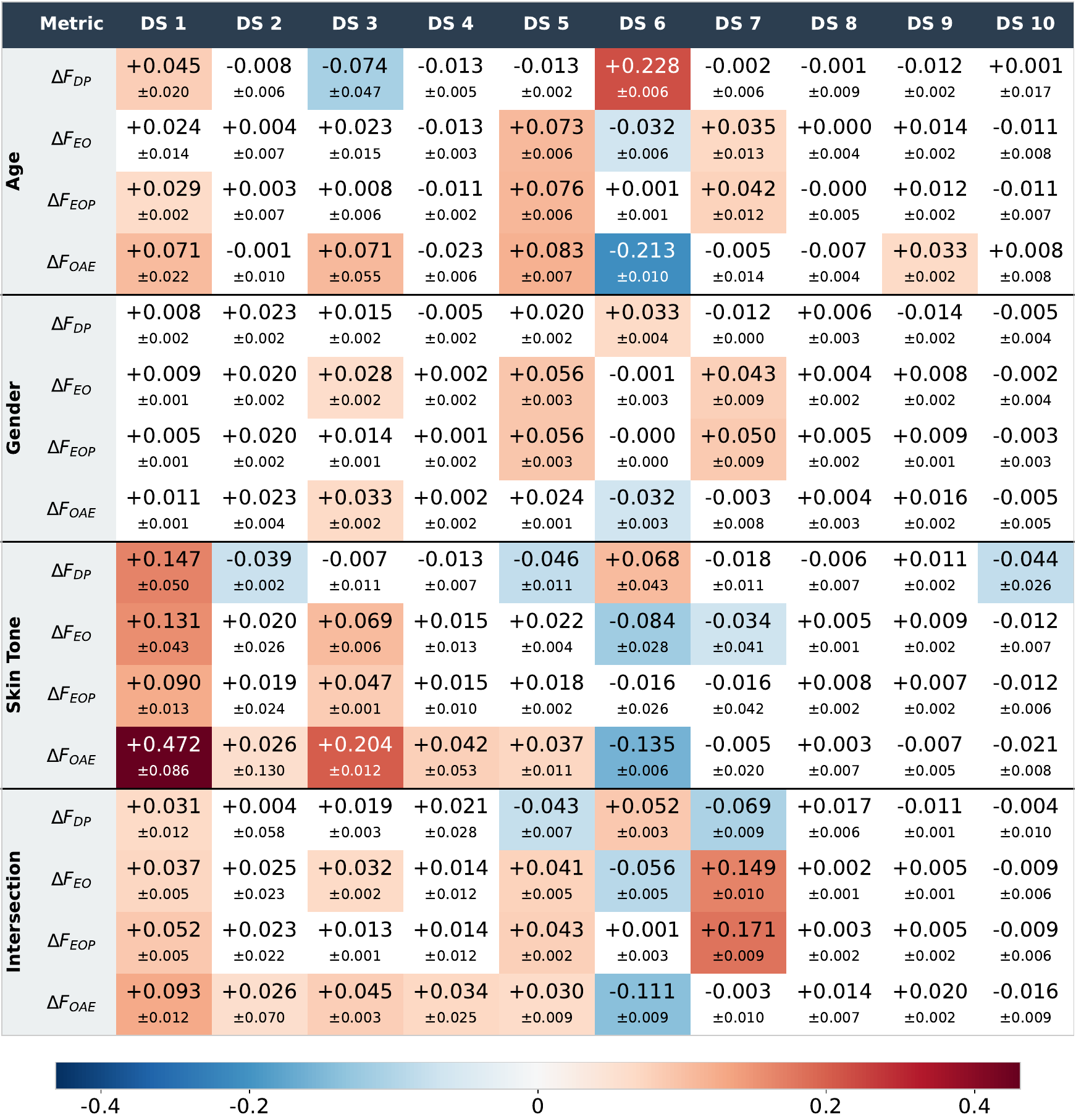}
\caption{SOYO}
\label{tab:soyo_per_ds}
\end{subtable}
\\[0.5cm]
\begin{subtable}{0.48\textwidth}
\centering
\includegraphics[width=\linewidth]{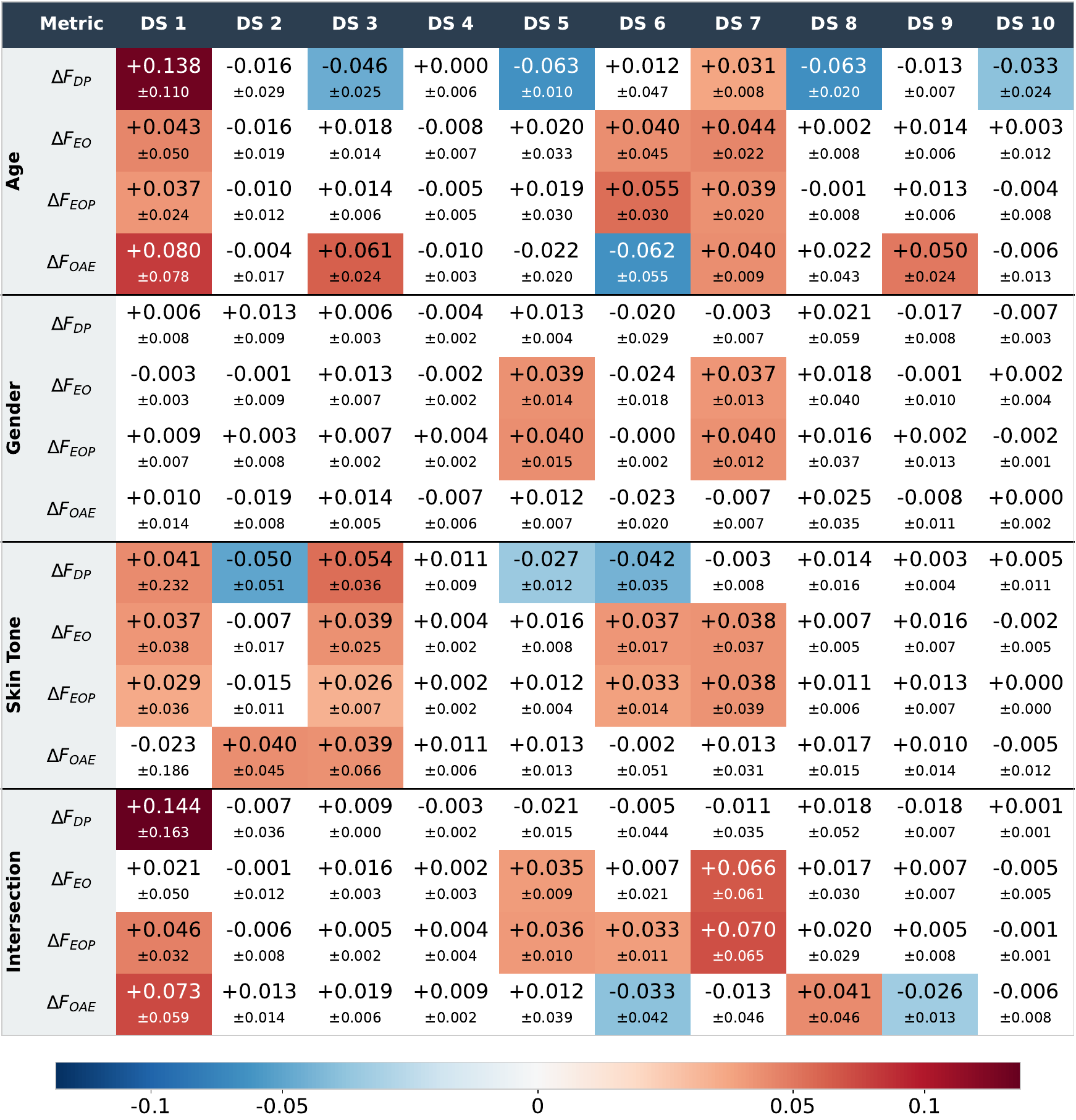}
\caption{S-Prompts}
\label{tab:sprompts_per_ds}
\end{subtable}
\caption{Comparison of (Un)fairness at the end of continually learning across the 11 generator sequence with PINA, SOYO, and S-Prompts for test sets of all previously observed tasks, i.e.~test set 1 after training on 11 tasks, test set 2 after 11 tasks, and so on. To focus on whether the CL algorithm has overall improved or deteriorated fairness, the shown values (mean $\pm$ std over five seeds) are differences of the final fairness with respect to the fairness obtained right after training on the respective generator, e.g.~discrepancy in fairness of task 2 after training on task 2 vs. task 2 after training on all 11 tasks. As before, red (blue) marks the many deteriorating (few improving) fairness dimensions. Cells remain white if the absolute difference is less than 2.5\% to ensure the figure remains legible.}
\label{tab:app_fairness_per_ds_1}
\end{table}

We show inter-task accuracy matrices for all these methods in figure \ref{fig:appendix_heatmaps_baselines} and figure \ref{fig:appendix_heatmaps}. These visualizations follow a similar structure as the exemplary matrix for DER++ as presented in the main paper (figure \ref{fig:derpp_heatmap}) and offer a granular view of task specific dynamics. Here, the diagonal cells highlight performance on the current generator at the point of its training, whereas the off-diagonal triangles respectively quantify forgetting (bottom-left) and zero-shot generalization (top-right).
The naive lower bound (figure \ref{fig:naive_heatmap}) exhibits the worst case of forgetting on earlier tasks as training progresses, depicted by the darkest bottom-left triangle. Off diagonal values drop to $13.98\%$ at test set (DS) 3 after training on task 10 marking a classic manifestation of catastrophic forgetting. In contrast, for accumulated, the matrix (figure \ref{fig:accumulated_heatmap}) remains lighter all over, close to white in the bottom left triangle, depicting the expected upper bound when the model has access to all past data at each training step. However, as pointed out in the main paper, the top-right triangle representing ``forward generalization'' remains limited for any method as no method explicitly addresses zero-shot learning.

EWC (figure \ref{fig:ewc_heatmap}) shows similar forgetting pattern as the naive lower bound and accuracy declines rapidly for earlier tasks, with performance barely improving over naive sequential, as the number of task increases. GradMatch displays better forgetting prevention than DER++ and EWC with more homogeneous knowledge retention over all, resulting in a more even color distribution in the inter-task accuracy matrix, as shown in figure \ref{fig:gradmatch_heatmap}.
Prompt based methods manage to maintain a stable accuracy across tasks and overall, SOYO (figure \ref{fig:soyo_heatmap}) achieves the best performance, outperforming GradMatch and DER++ (presented in the main paper) and approaching the accumulated baselines. The diagonal and bottom left triangle maintain a lighter hue overall showing consistently higher accuracy on previous tasks as well as current task. It even manage to retain a detection accuracy above $80\%$ for all the earlier tasks after being trained on $11$ task sequences, whereas, DER++ (as presented in the main paper) shows accuracy dropping below $80\%$ for some test sets. PINA shows significant knowledge preservation overall with some still darker ``stripes'' in the accuracy matrix (figure \ref{fig:pina_heatmap}). S-Prompts achieves similar performance but with more pronounced forgetting at DS 5 and 7.

As pointed out in the main paper, heterogeneous forgetting of previous knowledge could manifest as unpredictable vulnerabilities in a deployed application and could potentially be exploited and probed by adversaries. For instance, an attacker may deliberately use an older deepfake generation technique that the model has ``forgotten'' about to evade detection while committing an identity fraud. Thus, ensuring mitigated forgetting is the first critical step in a safety critical application.

\section{Asymmetric Fairness Degradation under Continual Updates}
\label{app:fairness_vs_forgetting}
This section complements the main body's findings with a brief overview of the evolution of fairness metrics for additional continual learning (CL) methods that we have considered in our evaluations: SOYO, PINA, S-Prompt, EWC, and GradMatch, along with the accumulated and naive sequential baselines. As described in the main paper, we compute the discrepancy for a given fairness score on a particular demographic group between the model trained on dataset $t$ and respectively evaluated at that time, and the same test set of generator $t$ evaluated at the end of training on the full eleven generator sequence. This discrepancy in fairness values has been presented for DER++ in the main paper (table \ref{tab:derpp_per_ds}) and similar tables for all other methods are presented in tables \ref{tab:app_fairness_per_ds_baselines} and \ref{tab:app_fairness_per_ds_1}.

As noted in the main paper, fairness degradation is in a complex overall interplay with forgetting. For instance, the accumulated baseline (table \ref{tab:accumulated_per_ds}) shows overall stable fairness. Under naive sequential training (table \ref{tab:naive_per_ds}), DS 1 experiences more than $50\%$ forgetting yet exhibits fairness degradation almost comparable to DS 9, where forgetting is only about half as much (as shown in figure \ref{fig:naive_heatmap}). Lowered $F_{DP}$ in combination with worsening $F_{OAE}$ reveals that the detector increasingly flags every group as fake at similar rates, but the share of correct decisions drifts further apart between groups.

Despite relatively better knowledge prevention, EWC shows similar disparities in fairness as the naive lower bound. For instance, fairness score degrades similarly for gender under both EWC and naive sequential. The disparities remain pronounced even in later tasks with extreme degradation in $F_{OAE}$ at DS 1, 2, 4 and 6 for skin tone and intersection and at DS 2 and DS 4, we see worse disparities for age at DS 3. However, accuracies at DS 3 and 6 are between $77\%$ and $79\%$ while performance at DS 2 and 4 remain around $61\%$.
For GradMatch, table \ref{tab:gm_per_ds}, $F_{OAE}$ for age degrade more for DS 3 than DS 5 but losses in accuracy are around $6\%$ and $20\%$ respectively.
Following better performance with respect to detection accuracy, SOYO also performs more fairly than other methods. As shown in table \ref{tab:soyo_per_ds} and signaled in the main paper, overall outcome becomes even more balanced across all demographics for later tasks, as reflected by almost all white cells post DS 7.
Other prompt based methods, S-Prompts maintains better overall fairness than PINA even though the former suffers higher cumulative forgetting, specially in later tasks. In DS 6, both PINA and S-Prompt maintain detection accuracy over $90\%$. Nevertheless, for skin tone, $F_{EO}$ degrades for S-Prompt and for age, both get an improved $F_{OAE}$.

Based on these observations, we conclude that the drift in fairness is heterogeneous and scores oscillate incoherently across methods. These asymmetric pattern reveals that there exists a complex interplay between the learning dynamics and data distribution highlighting complex and non linear relationship between forgetting and fairness across tasks. These findings corroborate the inference drawn in the main paper and hence, we present the main insights in there and defer a more detailed discussion to the appendix due to space constraints.

\section{CL Methods Can Treat Vulnerable Groups Even More Unfairly Than Fine-Tuning}
\label{app:fairness_disparities}

\begin{figure*}[p]
    \centering
    \subfloat{%
        \includegraphics[width=0.42\textwidth]{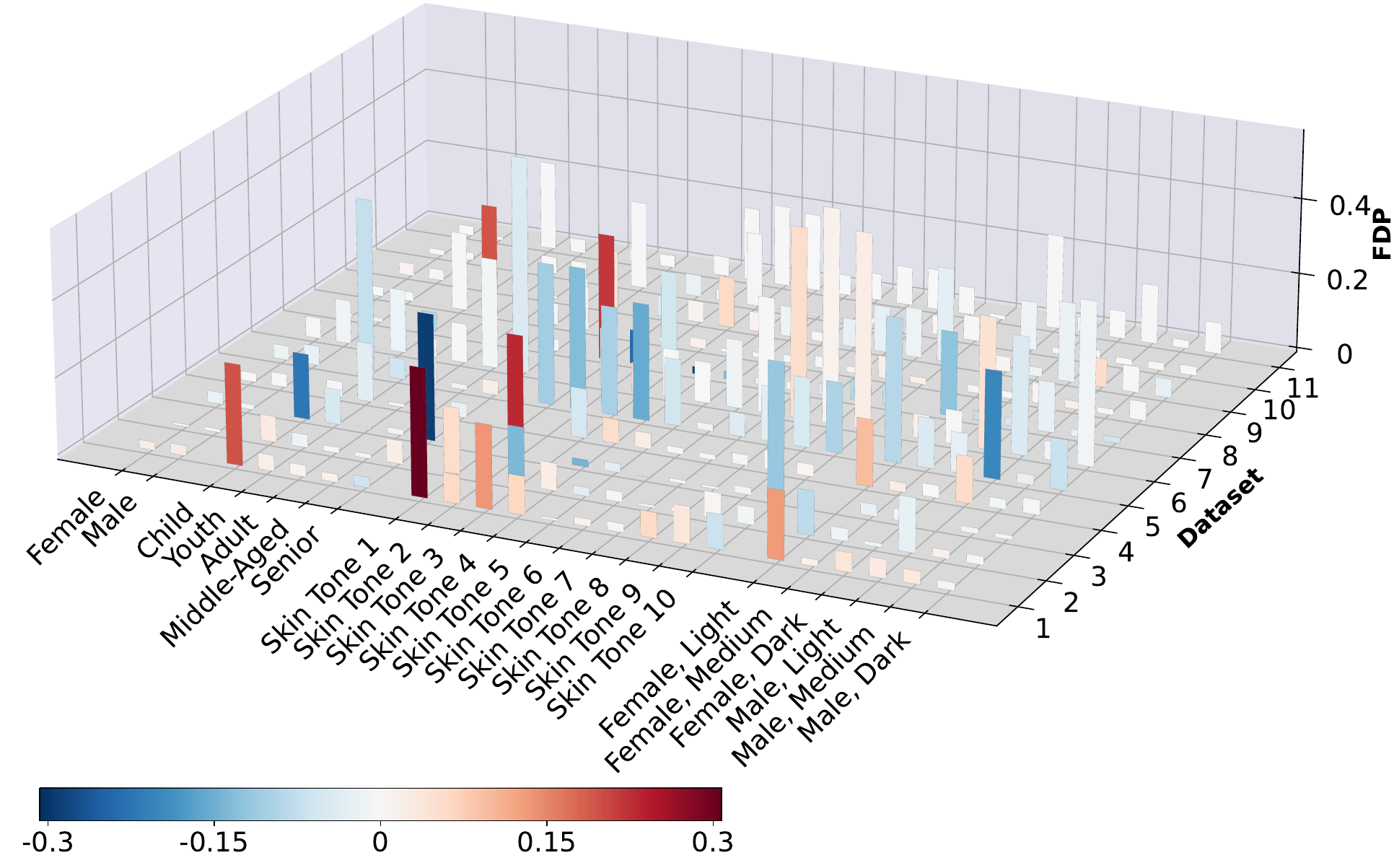}
        \label{fig:app_der_naive_fdp_app}%
    }
    \hspace{0.05\textwidth}
    \subfloat{%
        \includegraphics[width=0.42\textwidth]{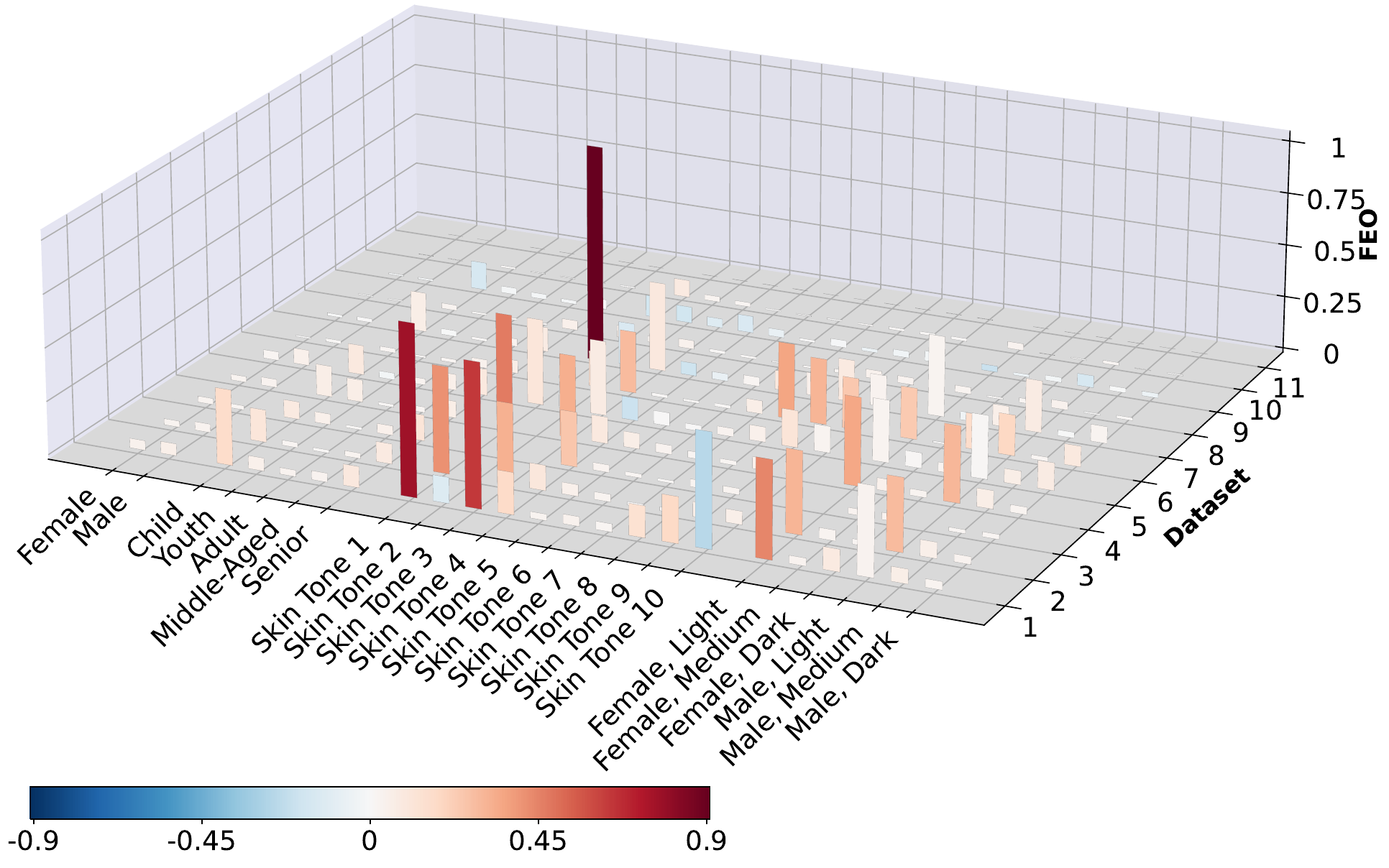}
        \label{fig:app_der_naive_feo_app}%
    }
    \\[0cm]
    \subfloat{%
        \includegraphics[width=0.42\textwidth]{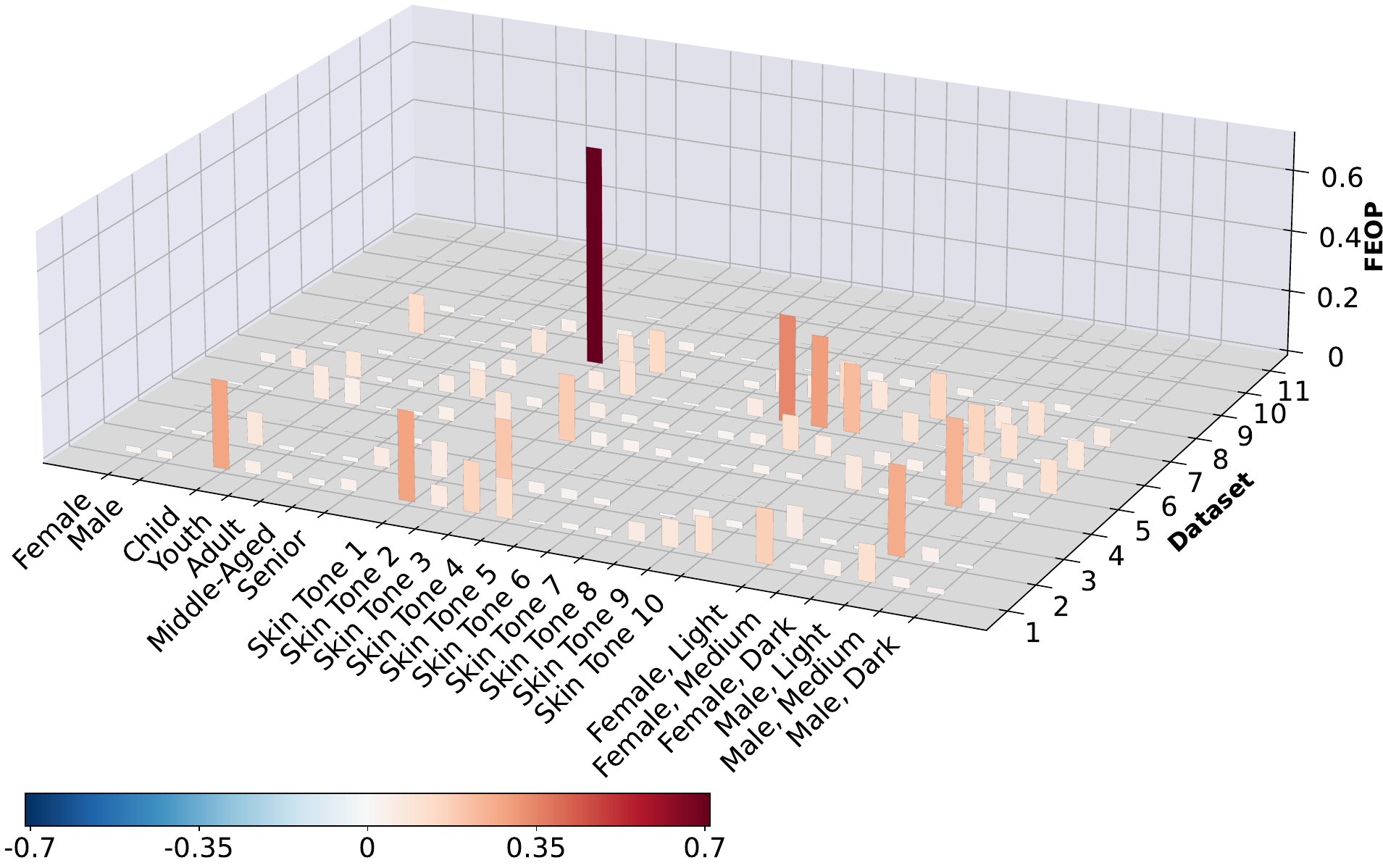}
        \label{fig:der_naive_feop_app}%
    }
    \hspace{0.05\textwidth}
    \subfloat{%
        \includegraphics[width=0.42\textwidth]{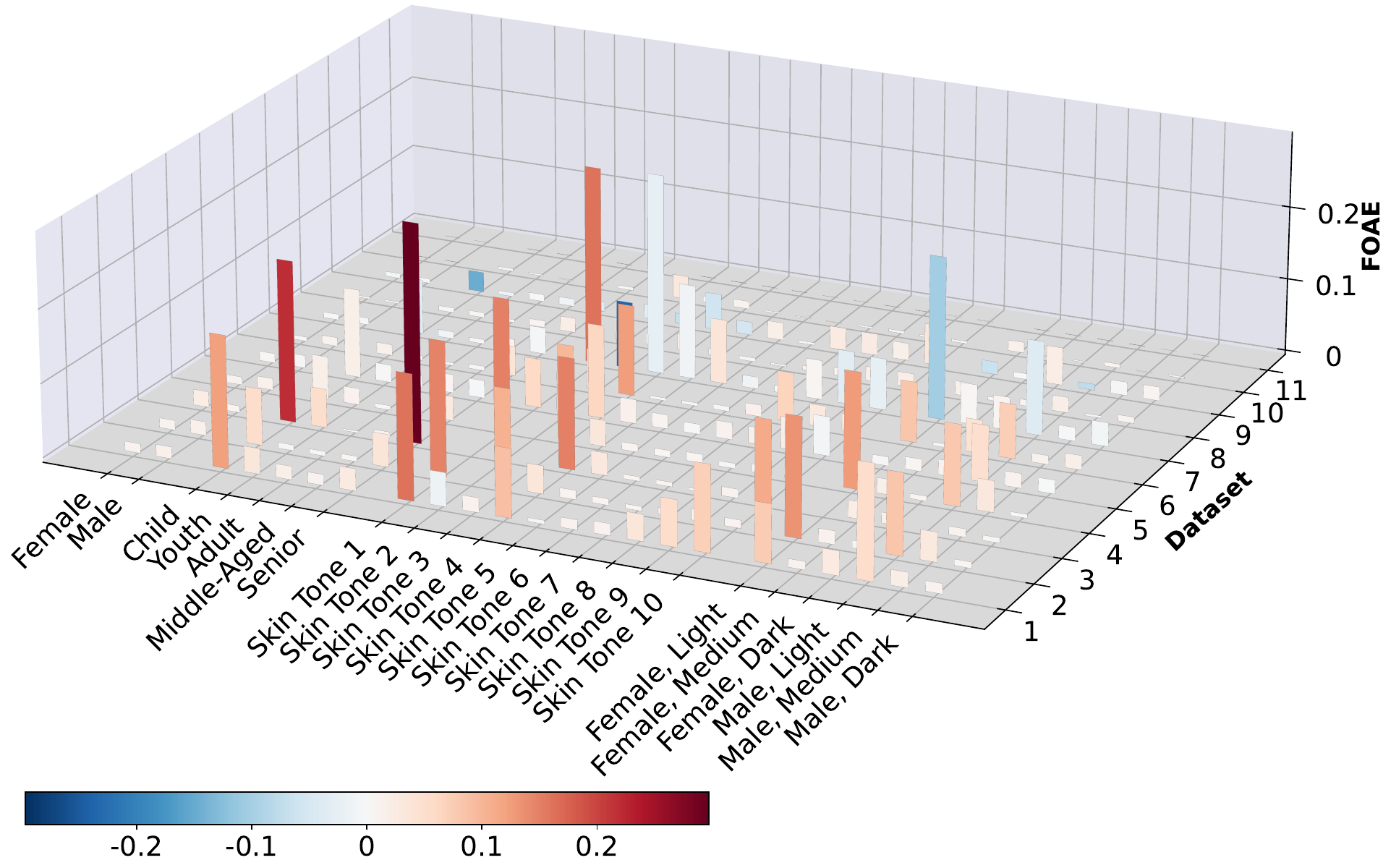}
        \label{fig:der_naive_foae_app}%
    }
    \caption{Evolution of fairness of DER++ across demographics (x-axes) for $F_{DP}$ (top left), $F_{EO}$ (top right), $F_{EOP}$ (bottom left) and $F_{OAE}$ (bottom right). Values are absolute fairness on each earlier dataset at the end of training (step 11), averaged over five seeds. As in earlier tables, higher means less fair. Color shows improvement (blue) or deterioration (red) relative to the naive sequential baseline, our reference to emphasize that fairness often deteriorates overall, and in parts even more than under naive fine-tuning.}
    \label{fig:derpp_fairness_evolution_from_naive_app}
\end{figure*}

In this section, we turn to which demographic subgroups are the most subject to the exacerbated unfairness of the continual learning (CL) algorithms.
In figure \ref{fig:derpp_fairness_evolution_from_naive_app}, we highlight the evolution of the fairness metrics, dissected across demographic attributes and for consistency shown for DER++. First, we see that overall fairness scores remain stable for binary gender, reflecting the balance in training data over time (recall figure \ref{fig:subgroups_percentage}).
In contrast, for age subgroups, fairness values exhibit substantially greater variability since typical vulnerable groups, like ``senior'' are  underrepresented. This leads to significant fluctuations and degradation, especially in $F_{DP}$ and $F_{EO}$. Importantly, the figure's color coding highlights where the values are even worse than in naive fine-tuning, which is the case in these dimensions.
For different skin tones, fairness scores display a similar uneven pattern for the lightest and darkest skin tones. These groups experience the largest increases in $F_{EO}$, e.g.~the highest increase for light skin tones at DS 7 or degradation of fairness for skin tone 10 early on. Although the distribution is fairly balanced for binary gender, fairness values collapse at the intersection of gender and the under represented lighter skin tones. Here, the $Female\times Light$ subgroup shows the most pronounced degradation in $F_{OAE}$, indicating disparate detection and false alarm rates. This bias signals that the minority attribute (lighter skin tone) exacerbates unfairness rather than being counterbalanced by the majority attribute (gender).

\begin{figure*}[p]
    \centering
    \subfloat{%
        \includegraphics[width=0.42\textwidth]{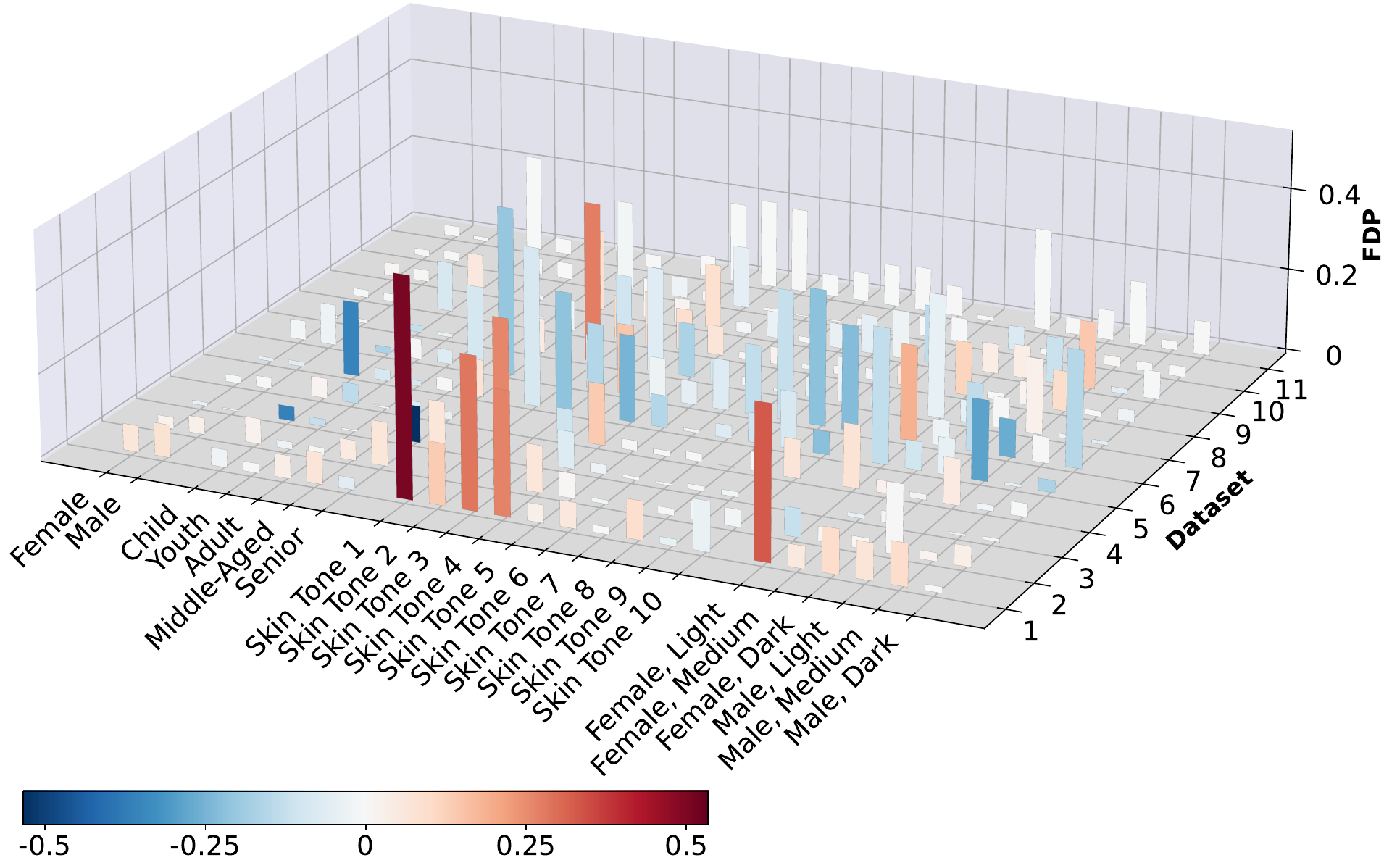}
        \label{fig:ewc_naive_fdp}%
    }
    \hspace{0.05\textwidth}
    \subfloat{%
        \includegraphics[width=0.42\textwidth]{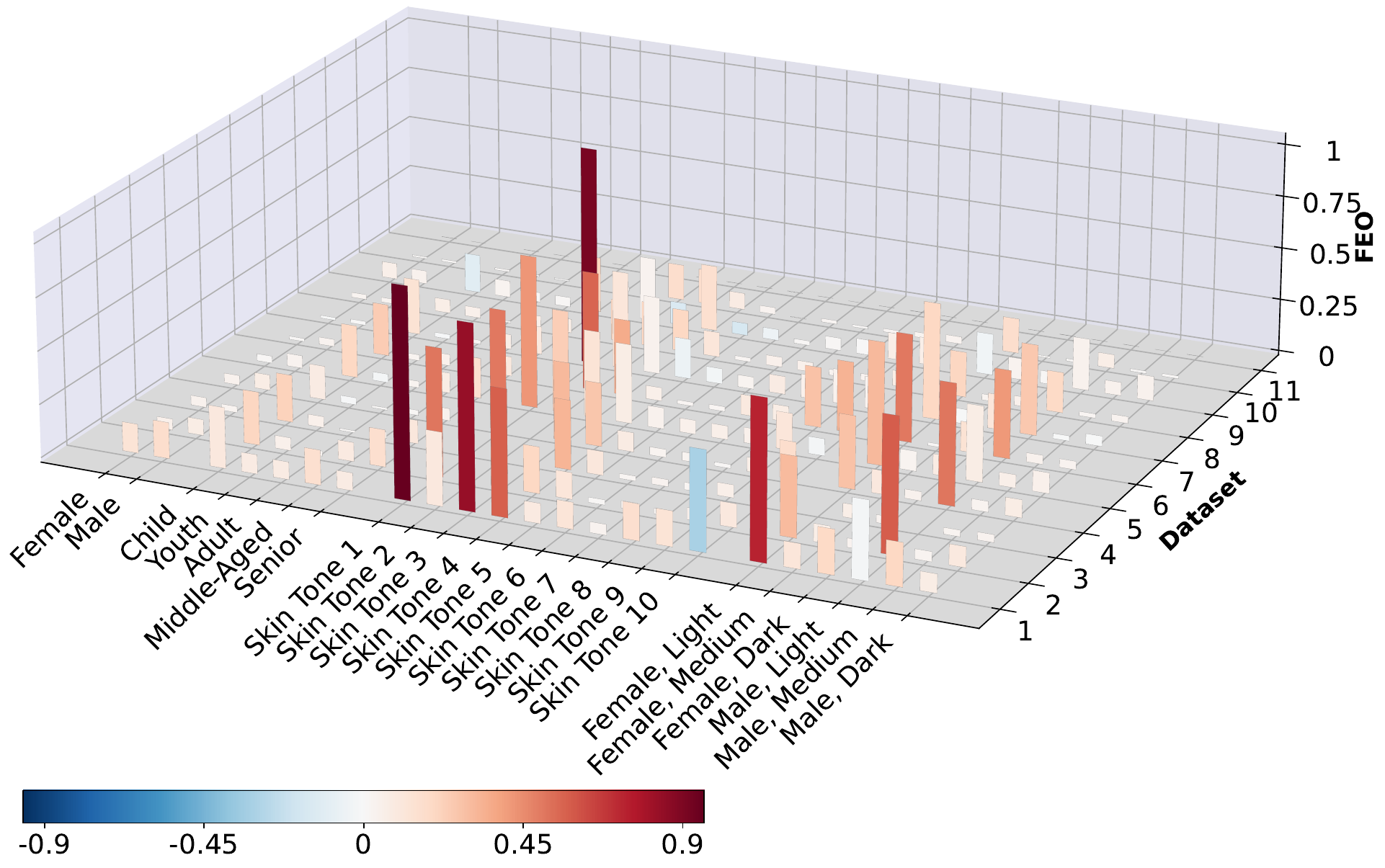}
        \label{fig:ewc_naive_feo}%
    }
    \\[0cm]
    \subfloat{%
        \includegraphics[width=0.42\textwidth]{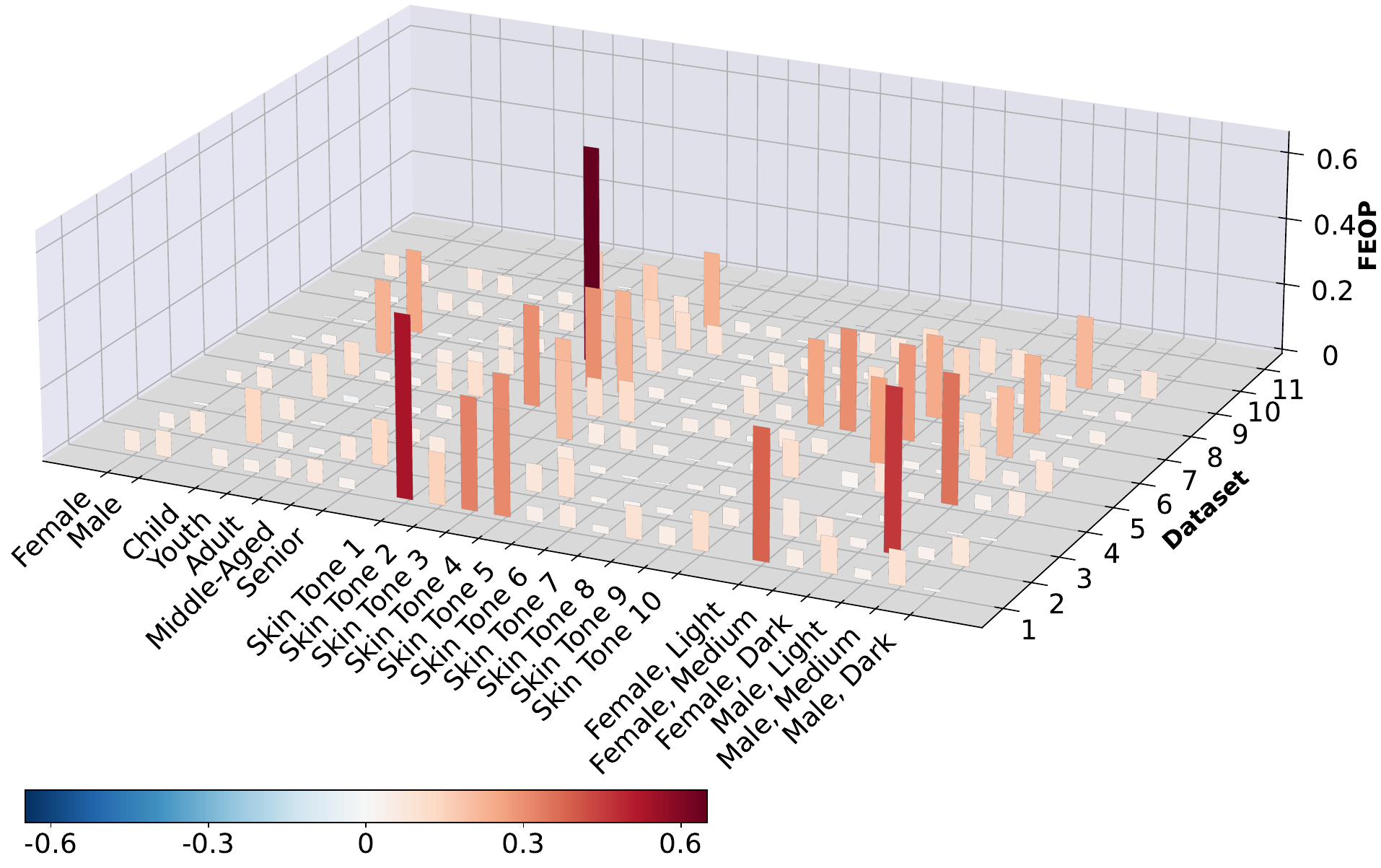}
        \label{fig:ewc_naive_feop}%
    }
    \hspace{0.05\textwidth}
    \subfloat{%
        \includegraphics[width=0.42\textwidth]{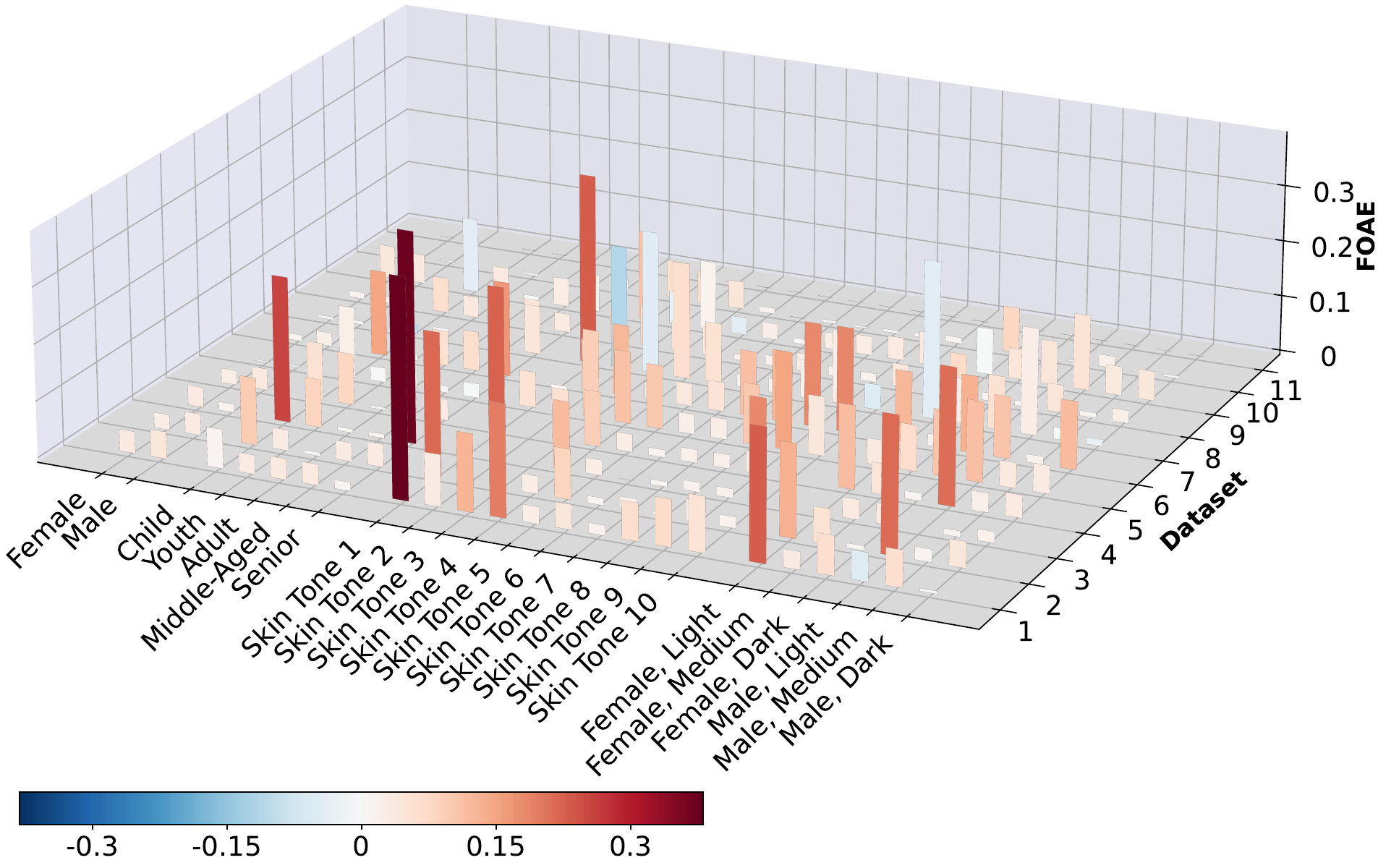}
        \label{fig:ewc_naive_foae}%
    }
    \caption{Evolution of fairness of EWC across demographics (x-axes) for $F_{DP}$ (top left), $F_{EO}$ (top right), $F_{EOP}$ (bottom left) and $F_{OAE}$ (bottom right). Values are absolute fairness on each earlier dataset at the end of training (step 11), averaged over five seeds. As in earlier tables, higher means less fair. Color shows improvement (blue) or deterioration (red) relative to the naive sequential baseline, our reference to emphasize that fairness often deteriorates overall, and in parts even more than under naive fine-tuning.}
    \label{fig:app_ewc_fairness_evolution_from_naive}
\end{figure*}

\begin{figure*}[t]
    \centering
    \subfloat{%
        \includegraphics[width=0.42\textwidth]{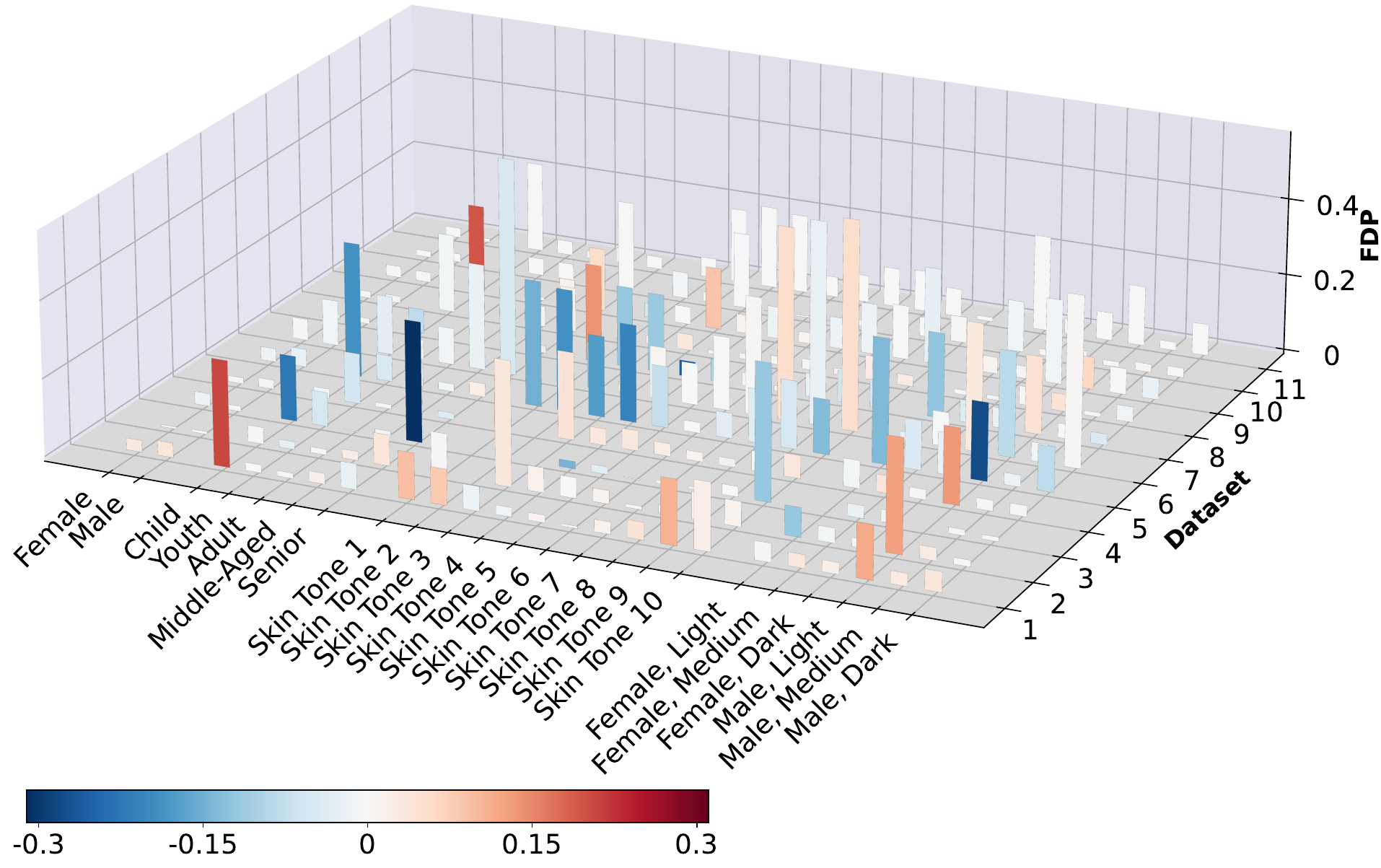}
        \label{fig:gm_naive_fdp}%
    }
    \hspace{0.05\textwidth}
    \subfloat{%
        \includegraphics[width=0.42\textwidth]{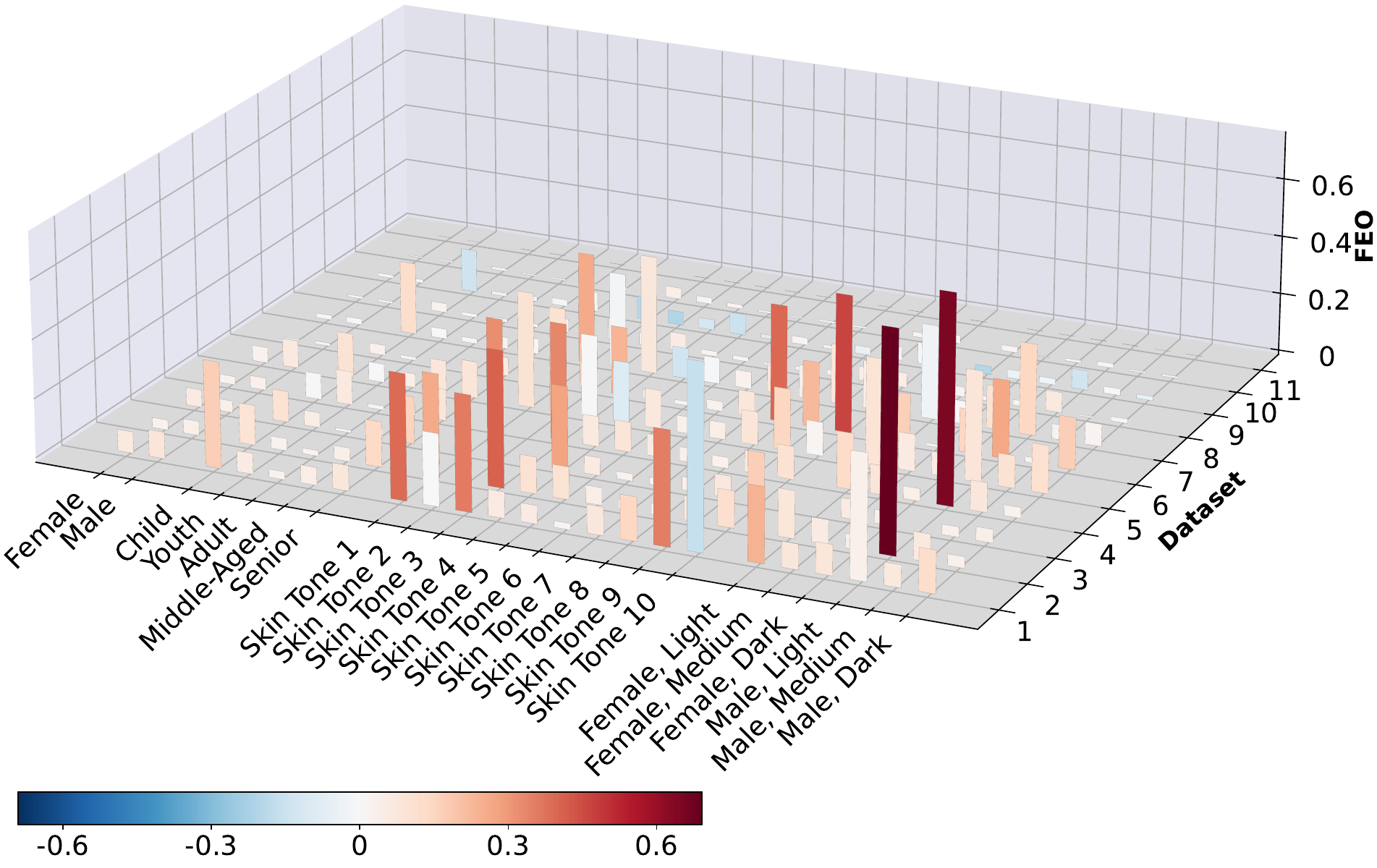}
        \label{fig:gm_naive_feo}%
    }
    \\[0cm]
    \subfloat{%
        \includegraphics[width=0.42\textwidth]{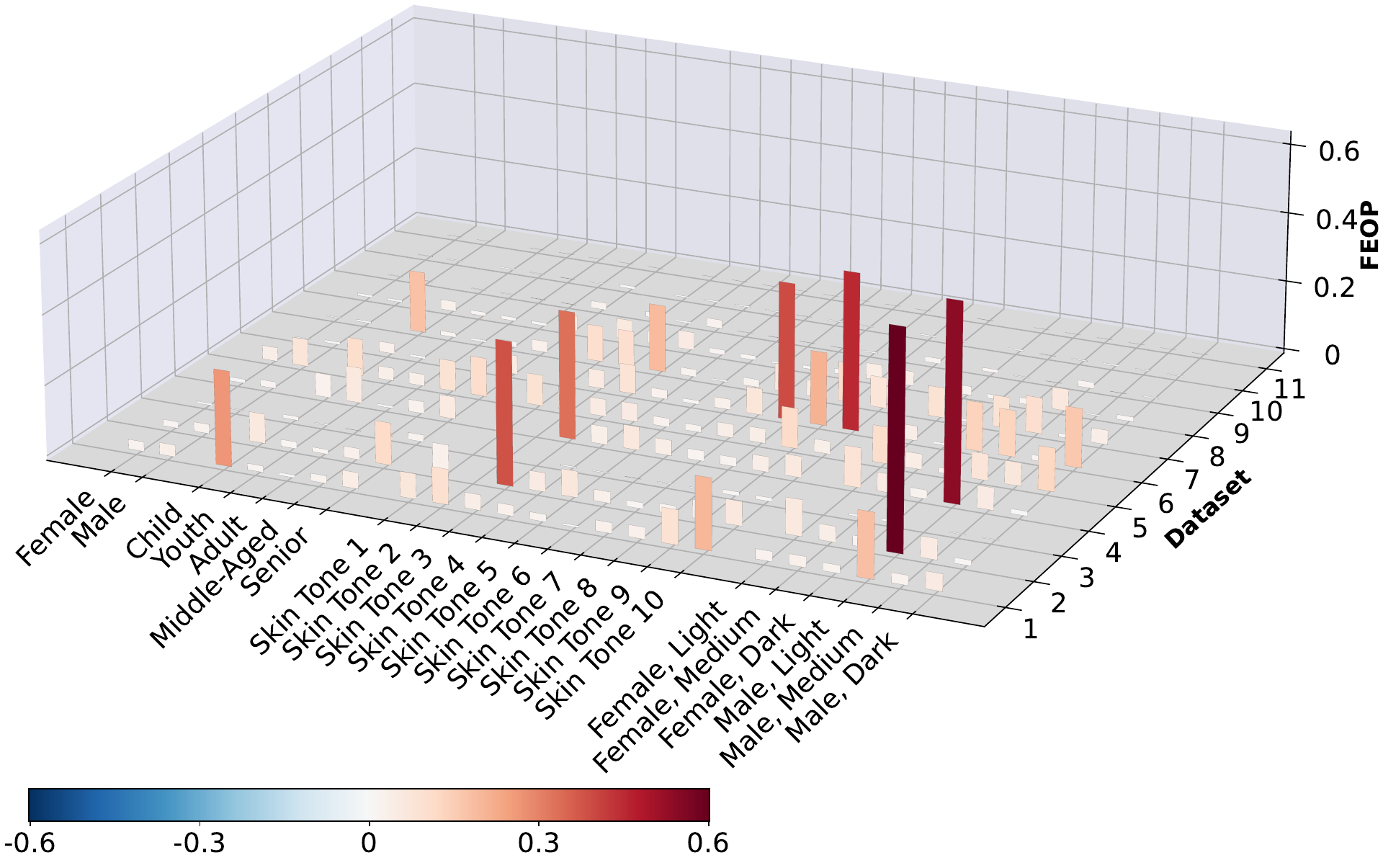}
        \label{fig:gm_naive_feop}%
    }
    \hspace{0.05\textwidth}
    \subfloat{%
        \includegraphics[width=0.42\textwidth]{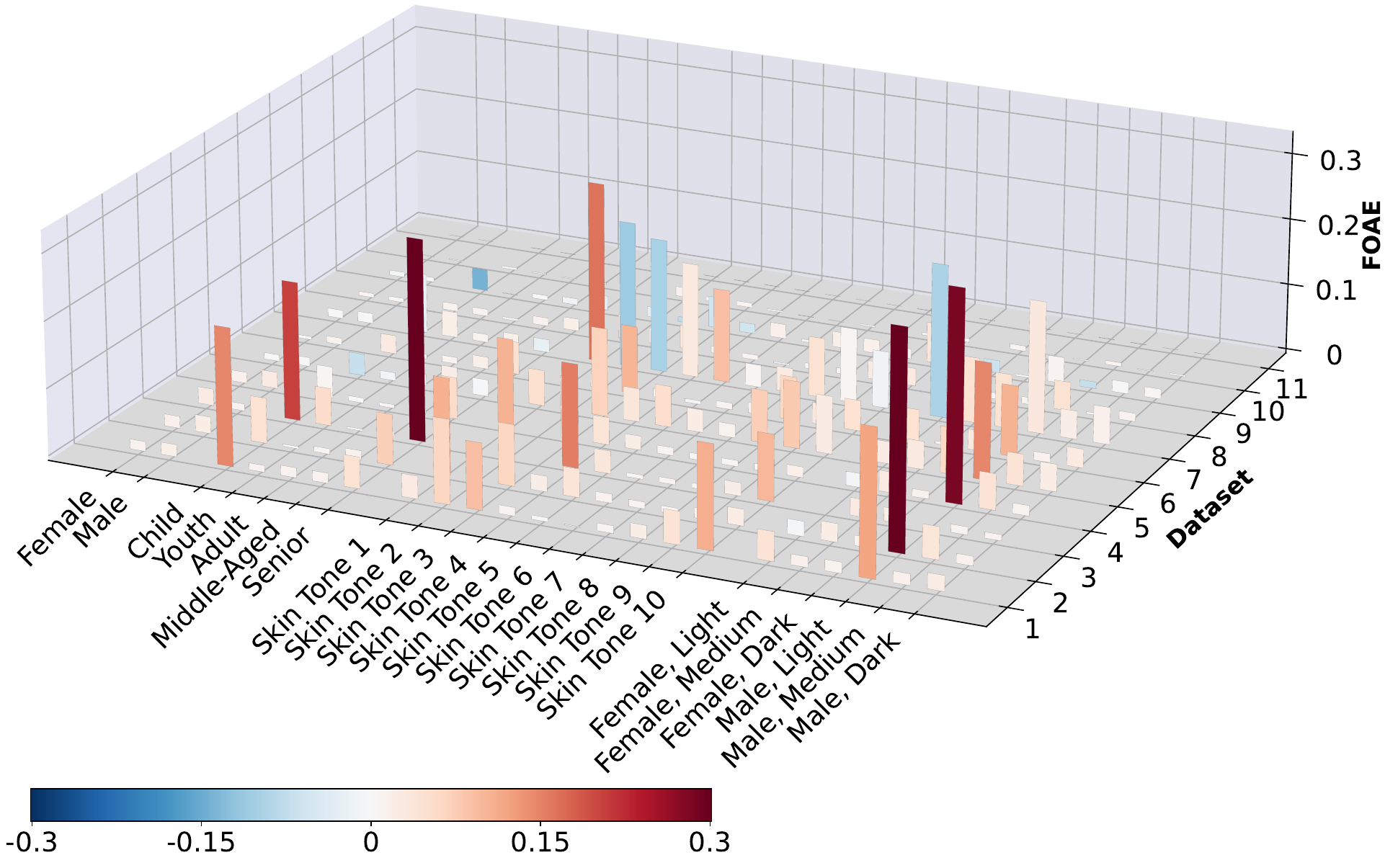}
        \label{fig:gm_naive_foae}%
    }
    \caption{Evolution of fairness of GradMatch across demographics (x-axes) for $F_{DP}$ (top left), $F_{EO}$ (top right), $F_{EOP}$ (bottom left) and $F_{OAE}$ (bottom right). Values are absolute fairness on each earlier dataset at the end of training (step 11), averaged over five seeds. As in earlier tables, higher means less fair. Color shows improvement (blue) or deterioration (red) relative to the naive sequential baseline, our reference to emphasize that fairness often deteriorates overall, and in parts even more than under naive fine-tuning.}
    \label{fig:app_gm_fairness_evolution_from_naive}
\end{figure*}

Compared to the naive lower bound, EWC shows highest degradation of fairness as marked in figure \ref{fig:app_ewc_fairness_evolution_from_naive}. We see more pronounced degradation in $F_{EO}$ and $F_{EOP}$ for lighter skin tones ($1-4$), $Male\times Light$ and $Female\times Light$, followed by age subgroups, child and senior revealing less protection against both attacks and false accusations. $F_{OAE}$ degrades significantly for child, senior, skin tone 1 and 10,  $Male\times Light$ and $Female\times Light$ between test set (DS) 1 and DS 6 indicating a further increase in unreliable predictions. As pointed out before, the figure's color coding highlights where the values are even worse than in naive fine-tuning, which is the case in these dimensions. Despite an improved $F_{DP}$ overall, GradMatch (figure \ref{fig:app_gm_fairness_evolution_from_naive}) shows noticeable degradation of $F_{EO}$ for $Male\times Light$ subgroup at DS 2, 4, 6 and 7 with the worst at DS 4. $F_{OAE}$ also deteriorates significantly for subgroups child, senior, several skin tones and $Male\times Light$ but asymmetrically across test sets.

As discussed in section \ref{app:fairness_vs_forgetting}, we see significant improvement in fairness disparities for SOYO across subgroups as well. However, at DS 7, we see degradation for $Male\times Light$ in $F_{EO}$ and $F_{EOP}$. At DS 1, skin tone 2, 3 and 10 show pronounced worsening of $F_{OAE}$ indicating unreliability of predictions, while accuracy of detection remains above $90\%$.
For S-Prompts, $F_{EO}$ and $F_{OAE}$ degrade almost consistently for under represented age groups, lighter and darkest skin tones. $Male\times Light$ and $Female\times Light$ also suffer but unevenly. Further, child shows highest degradation of $F_{OAE}$ at DS 3 while skin tone 8-9 shows highest degradation in $F_{EO}$ and $F_{EOP}$ at DS 6. However, accuracy of detection remains above $80\%$ at DS 1 and above $90\%$ for DS 3 and 6.
Similarly, for PINA, at DS 1, accuracy of detection remains above $80\%$. However, combined with an increase in all other unfairness metrics for the same subgroup reveal that fairness actually degrades: a higher $F_{DP}$ refers to higher systematic bias, higher $F_{OAE}$ less reliable detection rate and higher $F_{EO}$ and $F_{EOP}$ reveal that skin tone 10 now suffers from less protection against both attacks and false accusations. At later test sets, we see similar uneven patterns and under-represented subgroups getting subjected to exacerbated unfairness.

Based on these observations, we infer that the forgetting is not evenly distributed across demographic groups and that unfairness is unfortunately exacerbated for the often underrepresented or vulnerable population, including children, seniors, or very light or dark skin tones. These findings corroborate the inference drawn in the main paper and hence, we present the main insights in there and defer a more detailed discussion to the appendix due to space constraints.

\begin{figure*}[t]
    \centering
    \subfloat{%
        \includegraphics[width=0.42\textwidth]{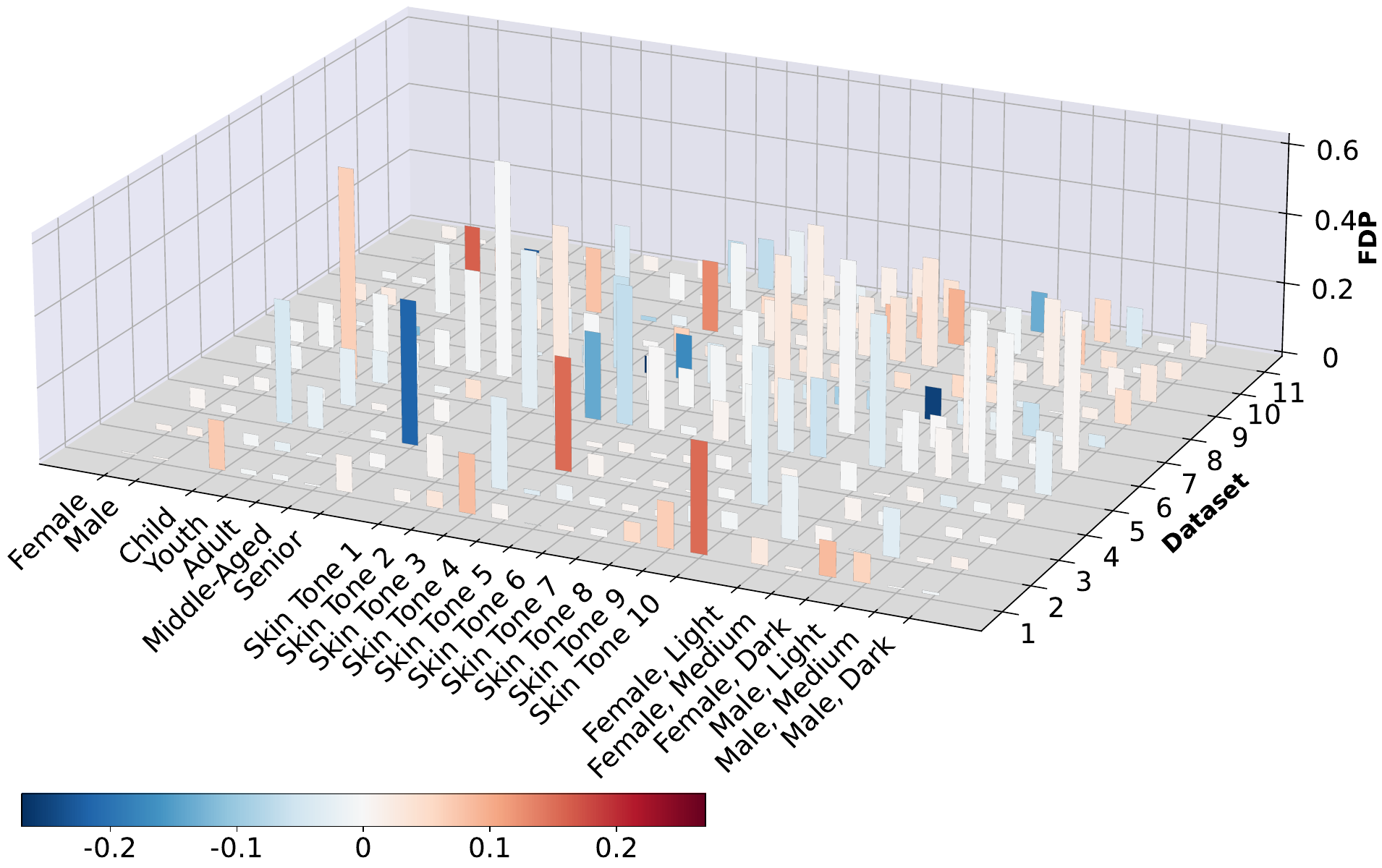}
        \label{fig:soyo_naive_fdp}%
    }
    \hspace{0.05\textwidth}
    \subfloat{%
        \includegraphics[width=0.42\textwidth]{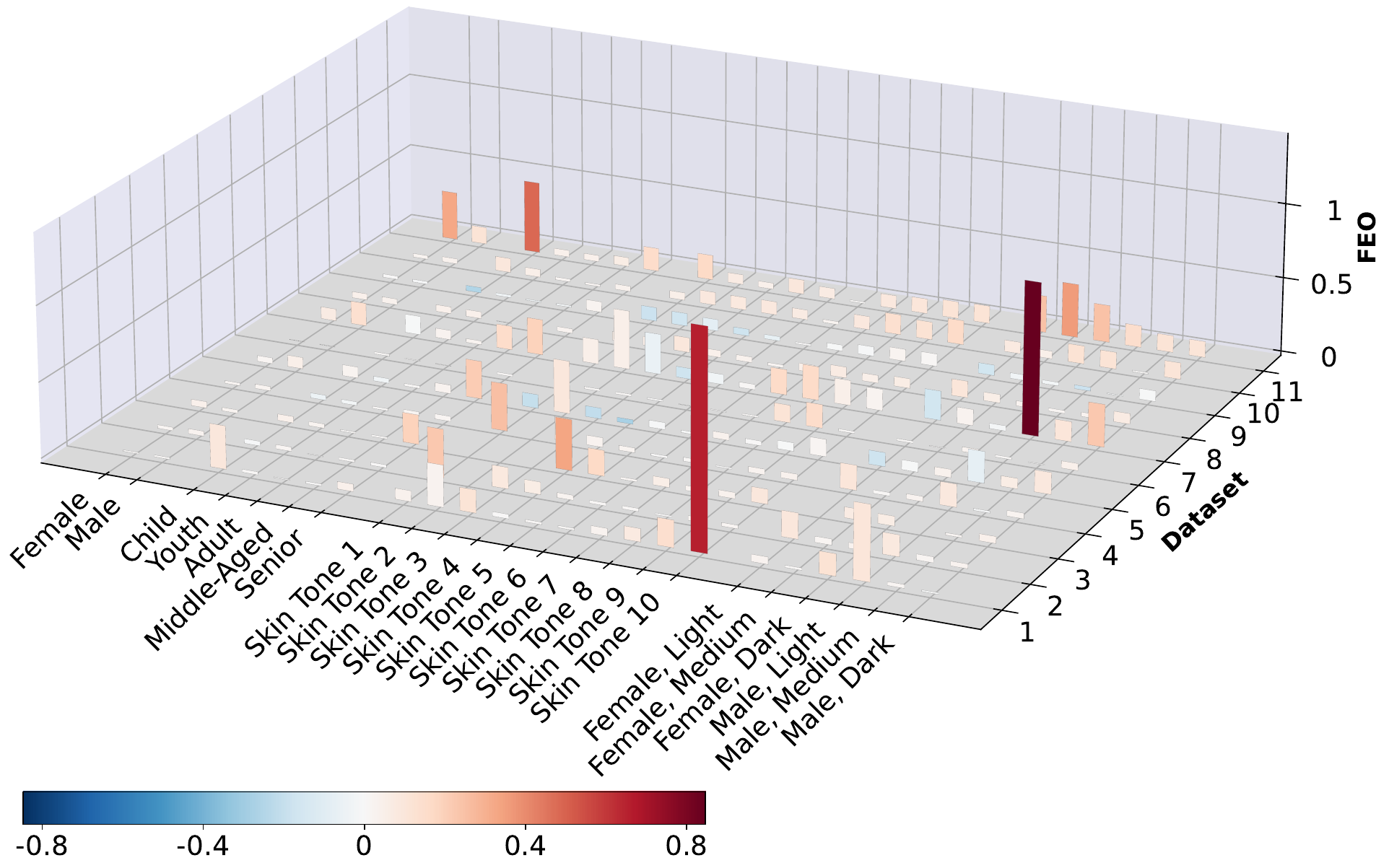}
        \label{fig:soyo_naive_feo}%
    }
    \\[0cm]
    \subfloat{%
        \includegraphics[width=0.42\textwidth]{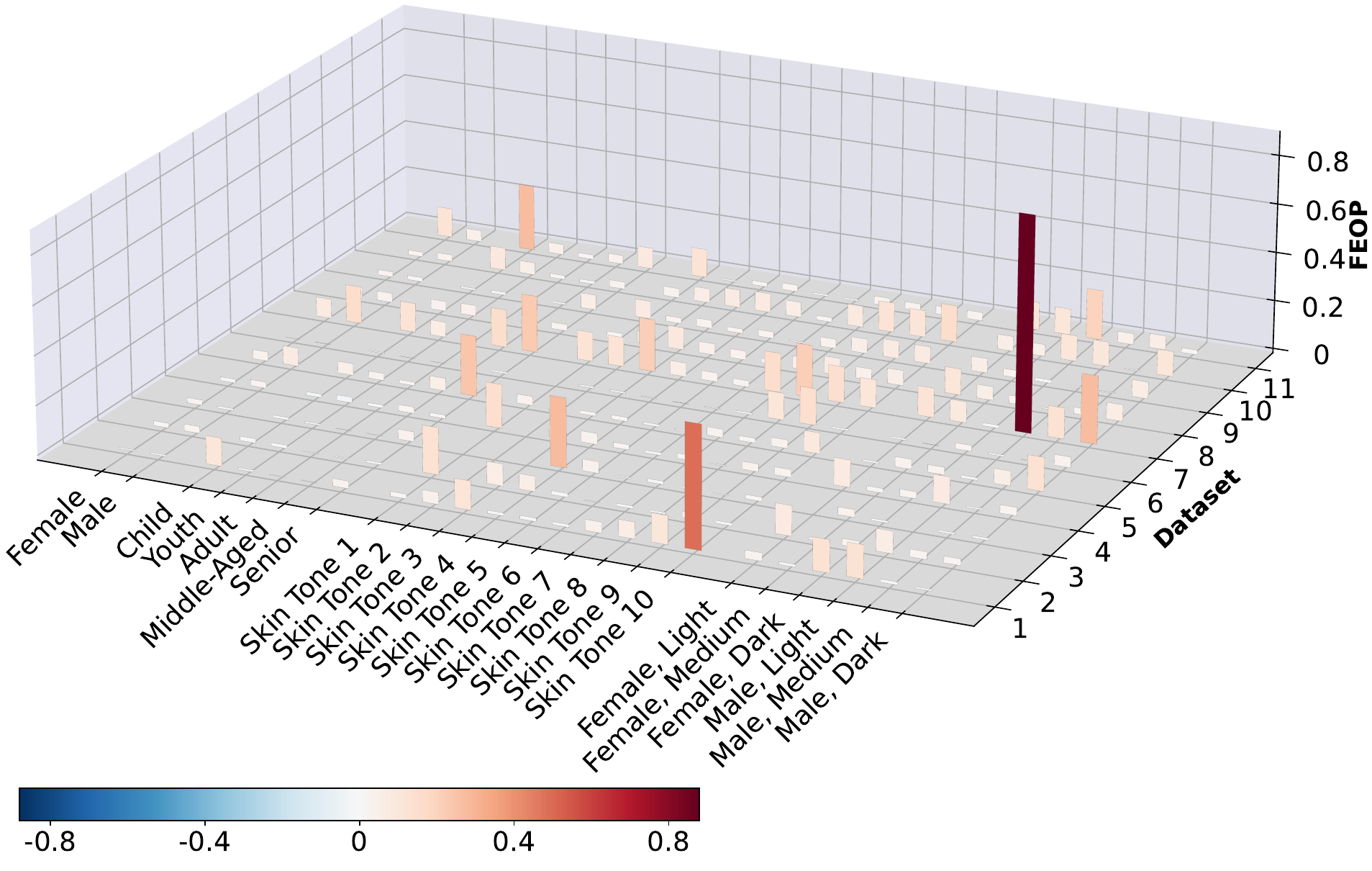}
        \label{fig:soyo_naive_feop}%
    }
    \hspace{0.05\textwidth}
    \subfloat{%
        \includegraphics[width=0.42\textwidth]{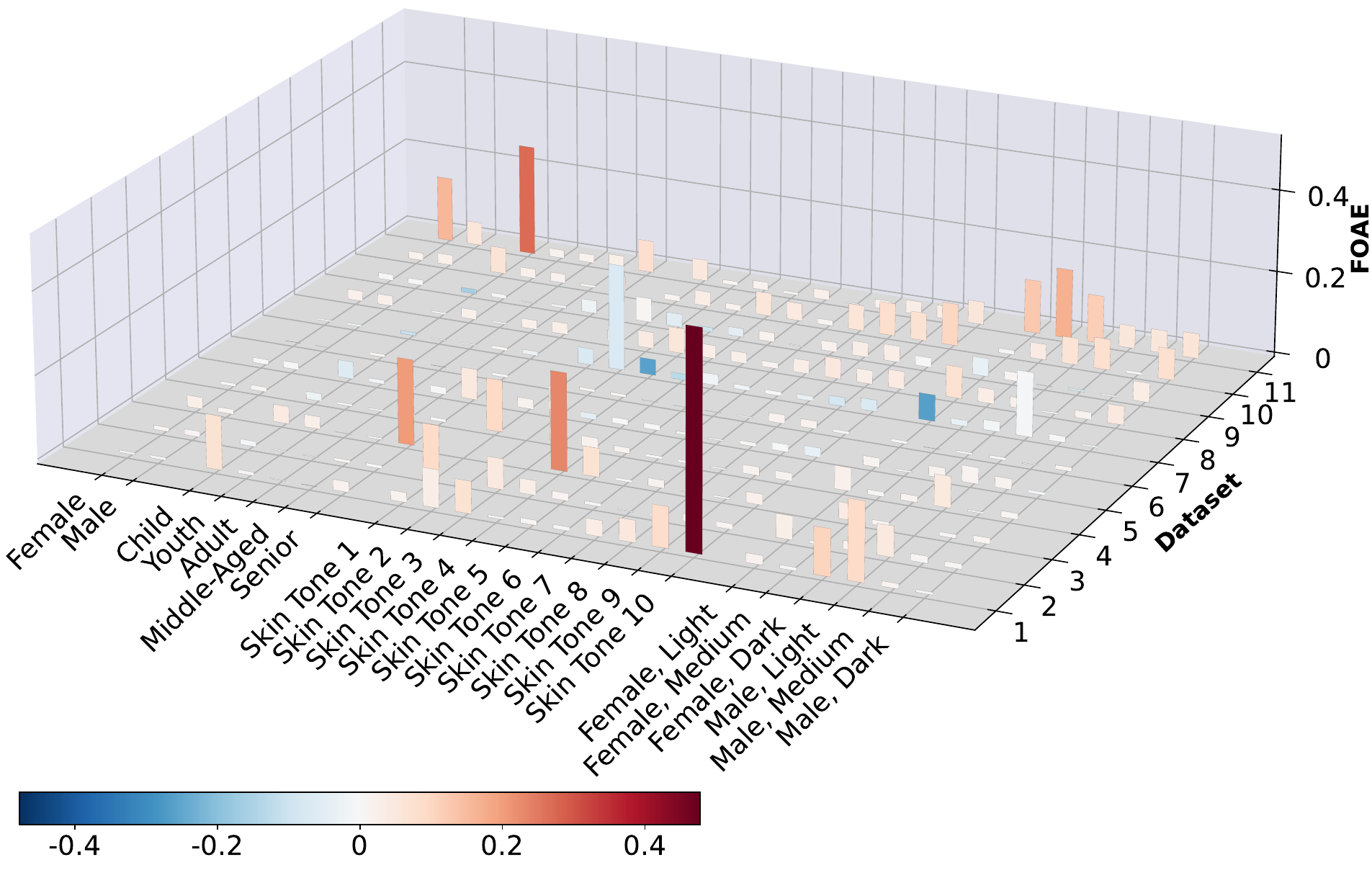}
        \label{fig:soyo_naive_foae}%
    }
    \caption{Evolution of fairness of SOYO across demographics (x-axes) for $F_{DP}$ (top left), $F_{EO}$ (top right), $F_{EOP}$ (bottom left) and $F_{OAE}$ (bottom right). Values are absolute fairness on each earlier dataset at the end of training (step 11), averaged over five seeds. As in earlier tables, higher means less fair. Color shows improvement (blue) or deterioration (red) relative to the naive sequential baseline, our reference to emphasize that fairness often deteriorates overall, and in parts even more than under naive fine-tuning.}
    \label{fig:app_soyo_fairness_evolution_from_naive}
\end{figure*}

\begin{figure*}[t]
    \centering
    \subfloat{%
        \includegraphics[width=0.42\textwidth]{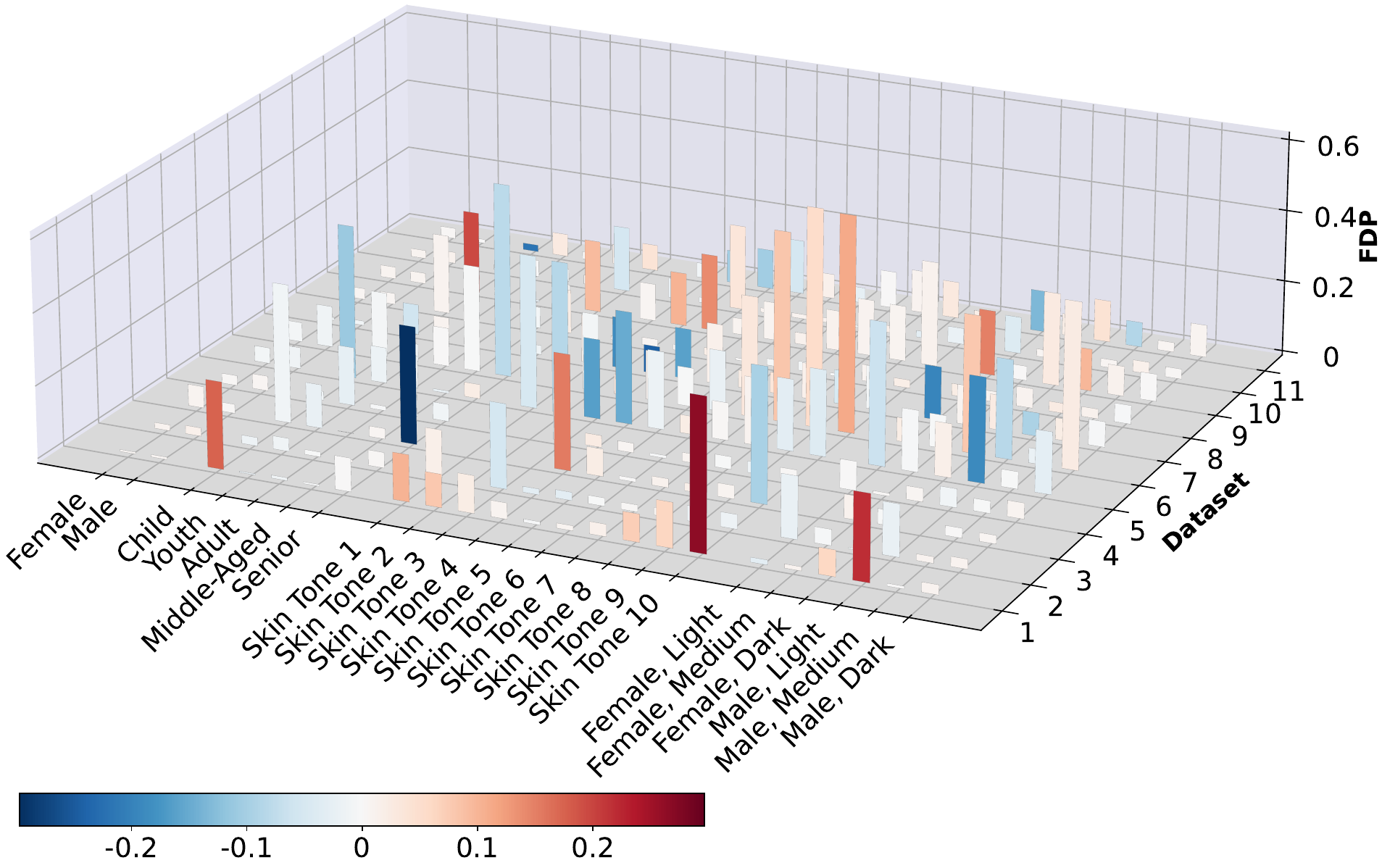}
        \label{fig:pina_naive_fdp}%
    }
    \hspace{0.05\textwidth}
    \subfloat{%
        \includegraphics[width=0.42\textwidth]{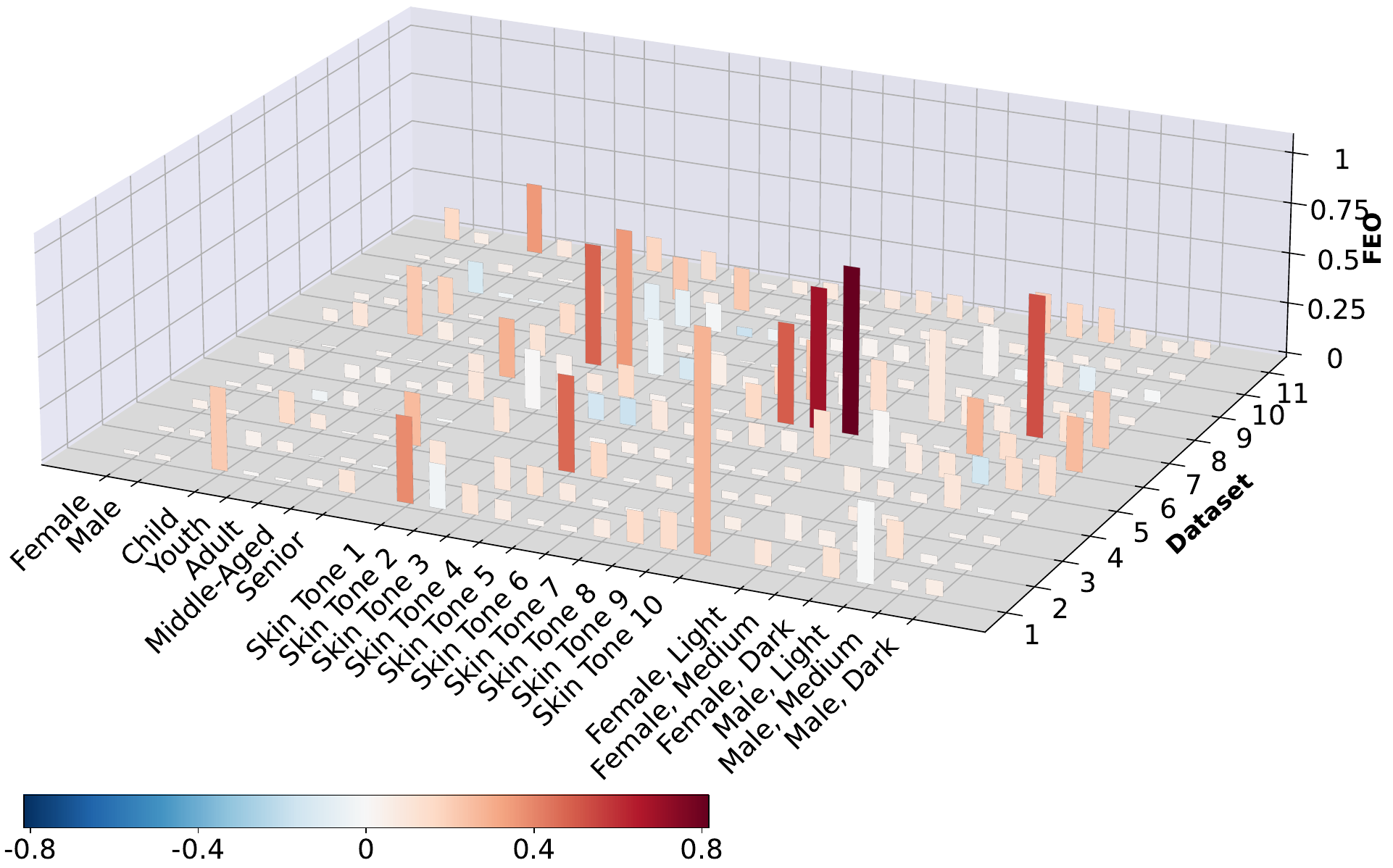}
        \label{fig:pina_naive_feo}%
    }
    \\[0cm]
    \subfloat{%
        \includegraphics[width=0.42\textwidth]{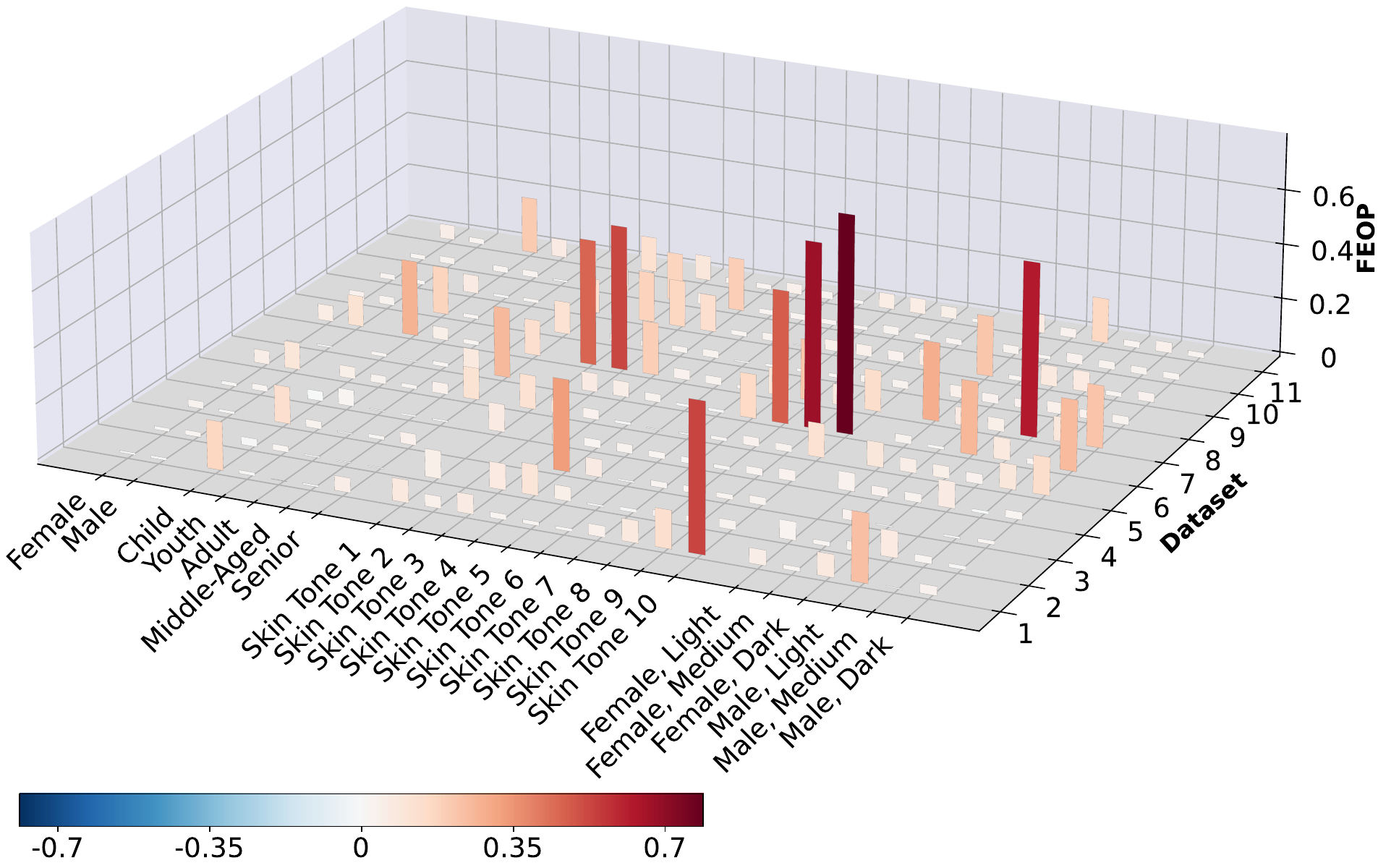}
        \label{fig:pina_naive_feop}%
    }
    \hspace{0.05\textwidth}
    \subfloat{%
        \includegraphics[width=0.42\textwidth]{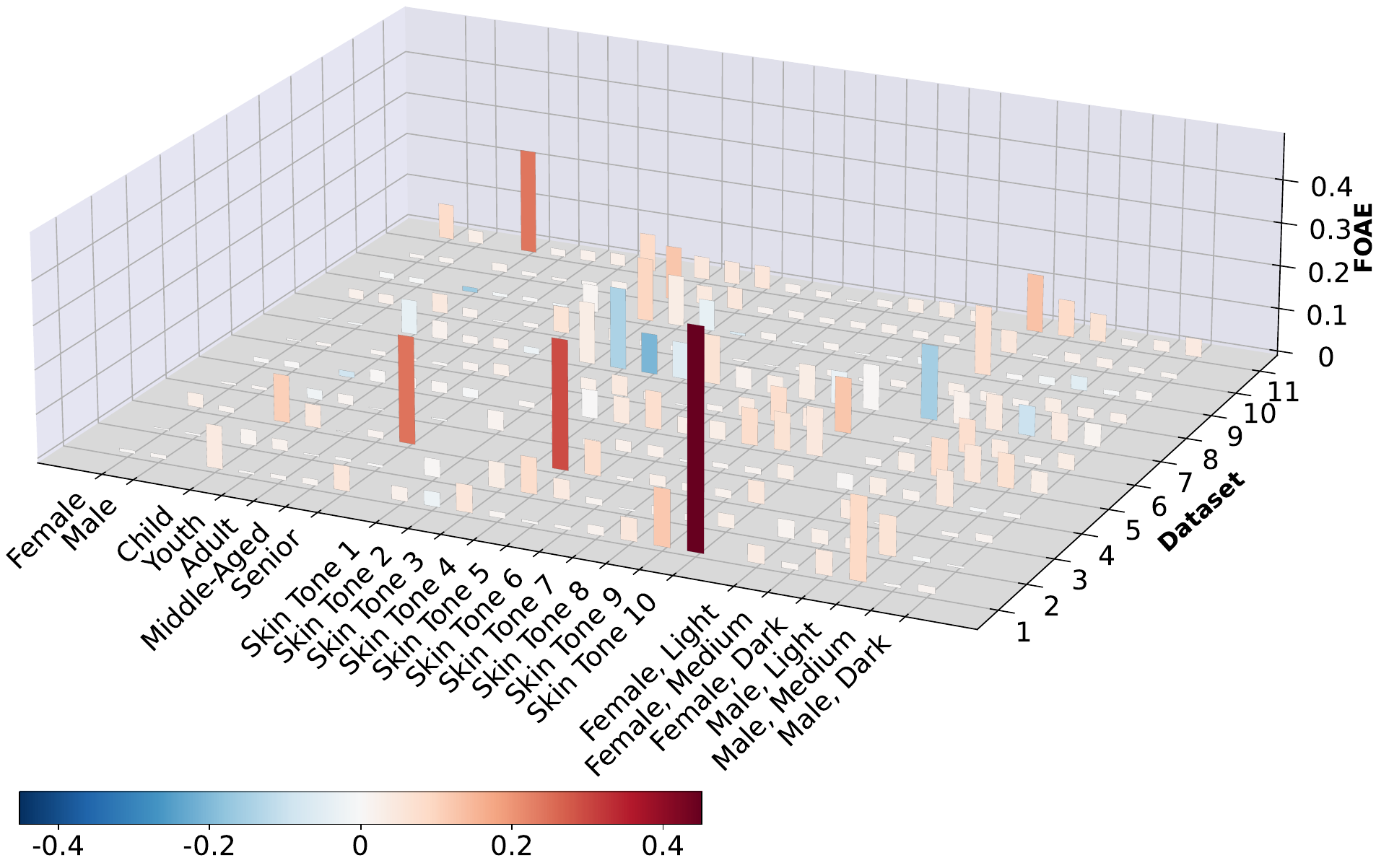}
        \label{fig:pina_naive_foae}%
    }
    \caption{Evolution of fairness of PINA across demographics (x-axes) for $F_{DP}$ (top left), $F_{EO}$ (top right), $F_{EOP}$ (bottom left) and $F_{OAE}$ (bottom right). Values are absolute fairness on each earlier dataset at the end of training (step 11), averaged over five seeds. As in earlier tables, higher means less fair. Color shows improvement (blue) or deterioration (red) relative to the naive sequential baseline, our reference to emphasize that fairness often deteriorates overall, and in parts even more than under naive fine-tuning.}
    \label{fig:app_pina_fairness_evolution_from_naive}
\end{figure*}

\begin{figure*}[t]
    \centering
    \subfloat{%
        \includegraphics[width=0.42\textwidth]{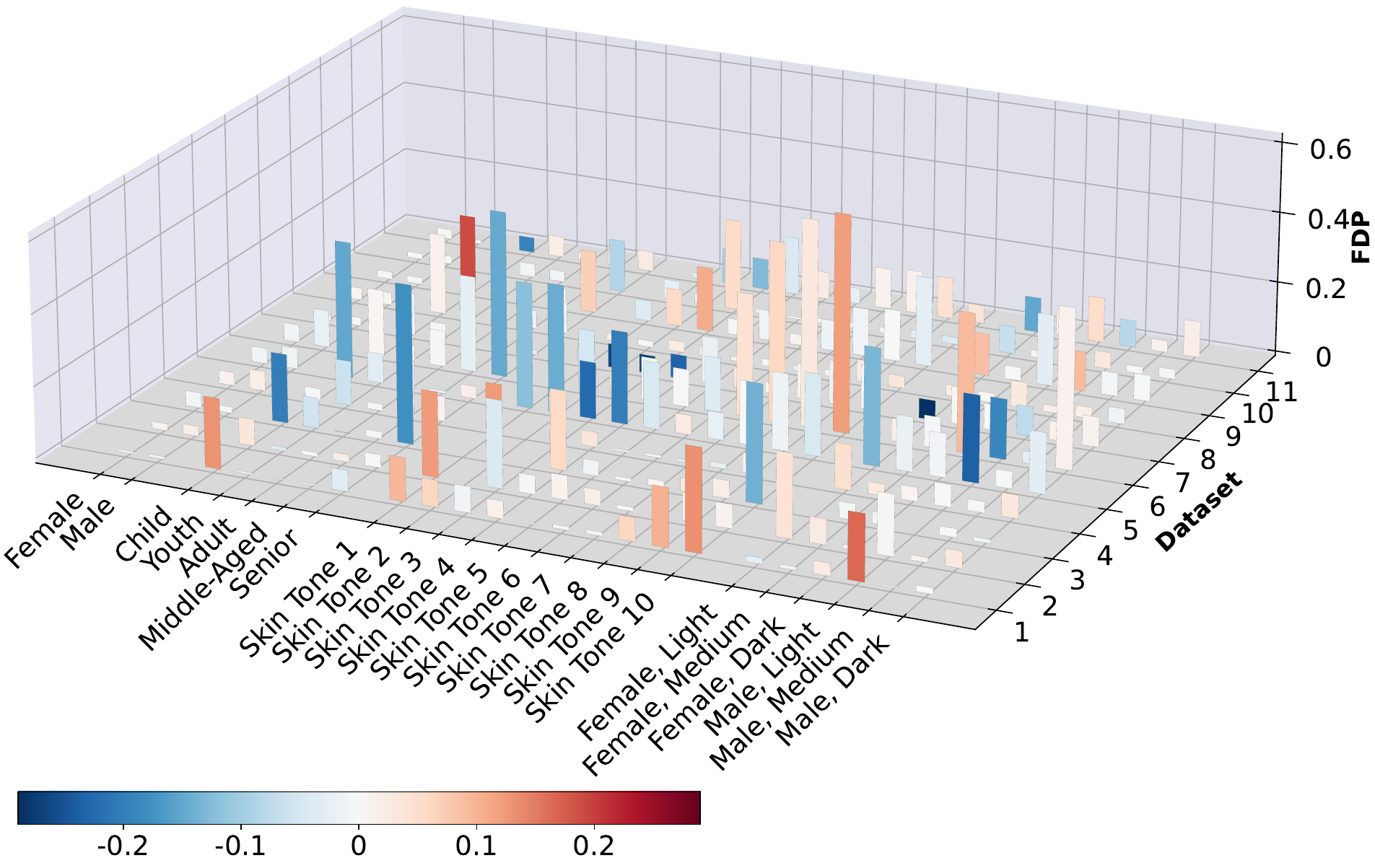}
        \label{fig:sprompts_naive_fdp}%
    }
    \hspace{0.05\textwidth}
    \subfloat{%
        \includegraphics[width=0.42\textwidth]{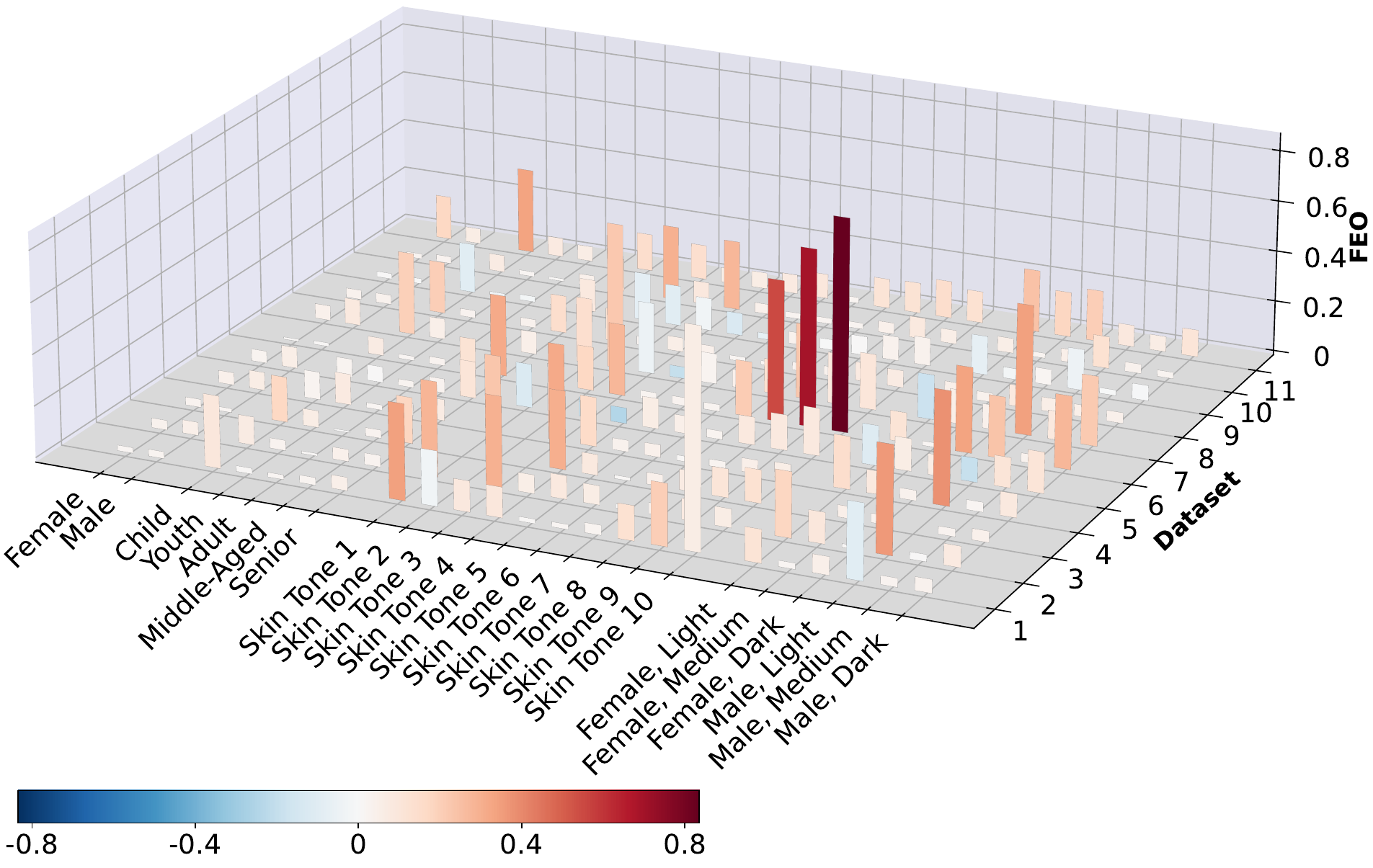}
        \label{fig:sprompts_naive_feo}%
    }
    \\[0cm]
    \subfloat{%
        \includegraphics[width=0.42\textwidth]{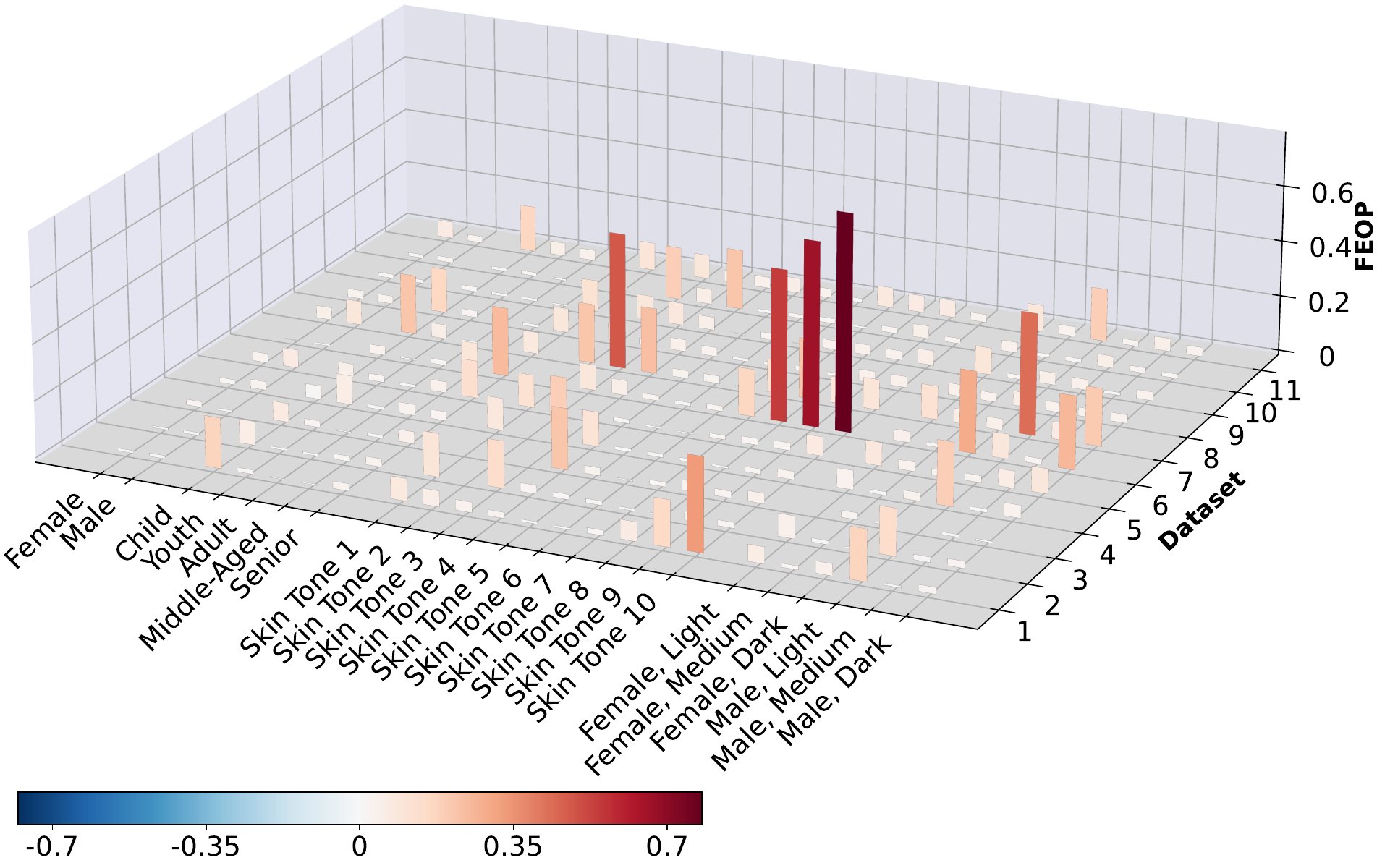}
        \label{fig:sprompts_naive_feop}%
    }
    \hspace{0.05\textwidth}
    \subfloat{%
        \includegraphics[width=0.42\textwidth]{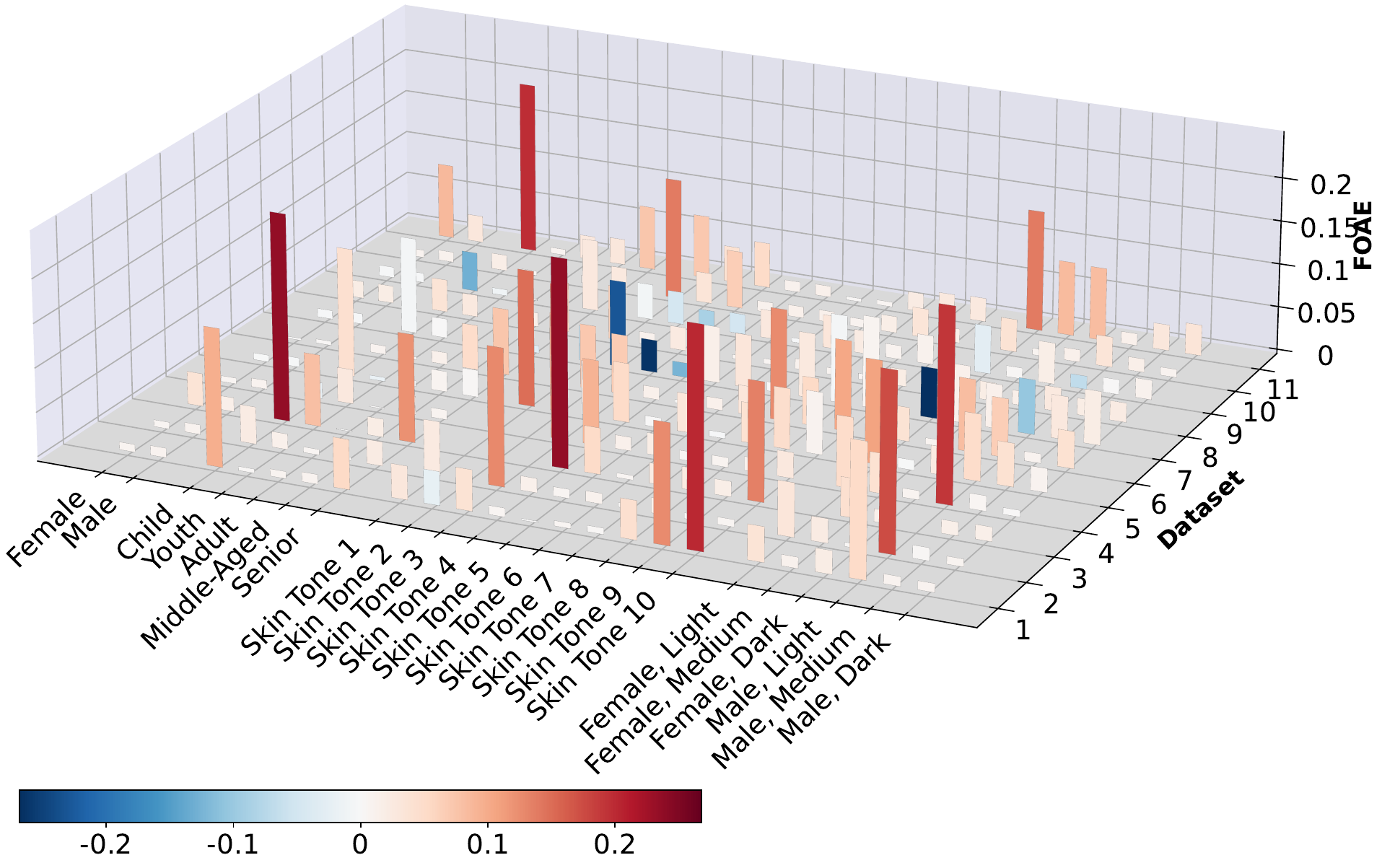}
        \label{fig:sprompts_naive_foae}%
    }
    \caption{Evolution of fairness of S-Prompts across demographics (x-axes) for $F_{DP}$ (top left), $F_{EO}$ (top right), $F_{EOP}$ (bottom left) and $F_{OAE}$ (bottom right). Values are absolute fairness on each earlier dataset at the end of training (step 11), averaged over five seeds. As in earlier tables, higher means less fair. Color shows improvement (blue) or deterioration (red) relative to the naive sequential baseline, our reference to emphasize that fairness often deteriorates overall, and in parts even more than under naive fine-tuning.}
    \label{fig:app_sprompts_fairness_evolution_from_naive}
\end{figure*}

\end{document}